\documentclass{article} 
\usepackage{iclr2027_conference,times}
\usepackage{microtype}
\usepackage[dvipsnames,table]{xcolor}
\usepackage{amsmath}
\usepackage{textcomp}

\usepackage{amsmath,amsfonts,bm}

\def\eqref#1{equation~\ref{#1}}

\def\1{\bm{1}}

\DeclareMathAlphabet{\mathsfit}{\encodingdefault}{\sfdefault}{m}{sl}
\SetMathAlphabet{\mathsfit}{bold}{\encodingdefault}{\sfdefault}{bx}{n}

\let\RelicOrigSection\section
\let\RelicOrigSubsection\subsection
\let\RelicOrigParagraph\paragraph
\newlength{\RelicOrigFloatsep}
\newlength{\RelicOrigDblfloatsep}
\newlength{\RelicOrigIntextsep}
\newlength{\RelicOrigAboveDisplaySkip}
\newlength{\RelicOrigBelowDisplaySkip}
\newlength{\RelicOrigAboveDisplayShortSkip}
\newlength{\RelicOrigBelowDisplayShortSkip}
\makeatletter
\def\section{\@startsection{section}{1}{\z@}{-1.45ex plus -.30ex minus -.15ex}{0.90ex plus .18ex minus .12ex}{\large\sc\raggedright}}
\def\subsection{\@startsection{subsection}{2}{\z@}{-1.30ex plus -.30ex minus -.15ex}{0.45ex plus .12ex}{\normalsize\sc\raggedright}}
\def\paragraph{\@startsection{paragraph}{4}{\z@}{0.85ex plus .20ex minus .12ex}{-1em}{\normalsize\bf}}
\makeatother

\usepackage{url}

\usepackage{graphicx}

\usepackage{booktabs}
\usepackage{longtable}
\usepackage{array}
\newcolumntype{P}[1]{>{\raggedright\arraybackslash}p{#1}}
\usepackage{placeins}
\usepackage{float}

\newfloat{relicalgorithm}{tbp}{loa}
\floatname{relicalgorithm}{Algorithm}

\usepackage{needspace}

\title{\textsc{Relic}: From Multi-Agent Collaboration to\\ Persistent Organizational Capability}

\usepackage{tabularx}
\usepackage{xltabular}
\usepackage{makecell}
\usepackage{colortbl}
\usepackage{threeparttable}
\usepackage{amssymb}

\definecolor{B0Dark}{HTML}{6F7D8F} 
\definecolor{B1Dark}{HTML}{4E6F91} 
\definecolor{B2Dark}{HTML}{6B5D8C} 
\definecolor{B3Dark}{HTML}{8F1237} 

\definecolor{ExecDark}{HTML}{8F1237}    
\definecolor{TextDark}{HTML}{6B5D8C}    
\definecolor{ScratchDark}{HTML}{78889A} 

\definecolor{PanelFill}{HTML}{E7DCE1}    
\definecolor{PanelSubFill}{HTML}{F1E9EC} 
\definecolor{BestFill}{HTML}{E4B6C6}     
\definecolor{EffectFill}{HTML}{F5E8ED}   
\definecolor{RuleGray}{HTML}{B9C6D3}     
\definecolor{TableMuted}{HTML}{748297}   

\newcolumntype{L}{>{\raggedright\arraybackslash}X}
\newcolumntype{R}{>{\raggedleft\arraybackslash}X}

\newcommand{\armhead}[2]{%
    \textcolor{#1}{\rule{1.1ex}{1.1ex}}%
    \hspace{0.35em}\textbf{#2}%
}

\newcommand{\bestcell}[1]{%
    \cellcolor{BestFill}\textbf{#1}%
}

\newcommand{\effectcell}[1]{%
    \cellcolor{EffectFill}\textbf{#1}%
}

\newcolumntype{C}[1]{>{\centering\arraybackslash}m{#1}}
\newcolumntype{Y}{>{\centering\arraybackslash}X}

\newcommand{\relictabletitle}[2]{%
    \rowcolor{PanelFill}%
    \multicolumn{#1}{@{}l@{}}{\normalsize\textbf{#2}}\\
    \addlinespace[1pt]%
}

\newcommand{\relictablepanel}[2]{%
    \rowcolor{PanelSubFill}%
    \multicolumn{#1}{@{}l@{}}{\textbf{#2}}\\
}

\newcommand{\relicestci}[2]{%
    \shortstack[c]{#1\\[0.05ex]{\normalfont\unboldmath\fontsize{6.5}{7.3}\selectfont
    \textcolor{TableMuted}{[#2]}}}%
}

\newcommand{\relicbestci}[2]{\bestcell{\relicestci{#1}{#2}}}
\newcommand{\reliceffectcell}[1]{\cellcolor{EffectFill}{\bfseries\boldmath #1}}
\newcommand{\reliceffectci}[2]{\reliceffectcell{\relicestci{#1}{#2}}}

\providecommand{\code}[1]{\texttt{\small #1}}
\providecommand{\NAcell}{\textsc{na}}

\providecommand{\Bzero}{B0}
\providecommand{\Bone}{B1}
\providecommand{\Btwo}{B2}
\providecommand{\Bthree}{B3}

\providecommand{\armhead}[2]{\textbf{#2}}
\providecommand{\bestcell}[1]{\textbf{#1}}
\providecommand{\effectcell}[1]{#1}

\newcommand{\coremark}{\textsuperscript{*}}
\newcommand{\leadmark}{\textsuperscript{\textdagger}}
\newcommand{\corrmark}{\textsuperscript{\textdaggerdbl}}

\author{
\bfseries
Hongyi Du\textsuperscript{1}\coremark\leadmark\corrmark
\quad
Tianyi Zhang\textsuperscript{2}\coremark
\quad
Weijia Zhang\textsuperscript{3}\coremark
\quad
Yi Yang\textsuperscript{1}
\quad
Haofei Yu\textsuperscript{1}
\quad
Kunlun Zhu\textsuperscript{1}
\\[0.35em]
\bfseries
Tianxiang Dai\textsuperscript{4}
\quad
Shang Jiang\textsuperscript{5}
\quad
Zhelun Gao\textsuperscript{6}
\quad
Jiaxin Pei\textsuperscript{7}
\quad
Shang Zhu\textsuperscript{8}\corrmark
\quad
Jiaxuan You\textsuperscript{1}\corrmark
\\[0.65em]
\normalfont\small
\textsuperscript{1}University of Illinois Urbana-Champaign
\quad
\textsuperscript{2}Harvey Mudd College
\quad
\textsuperscript{3}Yale University
\\[-0.05em]
\normalfont\small
\textsuperscript{4}Stanford University
\quad
\textsuperscript{5}Peking University
\quad
\textsuperscript{6}National University of Singapore
\\[-0.05em]
\normalfont\small
\textsuperscript{7}University of Texas at Austin
\quad
\textsuperscript{8}Together AI
}

\usepackage{seqsplit}
\ExplSyntaxOn
\cs_new_protected:Npn \relic_breakable_code:n #1
  {
    \group_begin:
    \ttfamily
    \tl_set:Nn \l_tmpa_tl {#1}
    \tl_replace_all:Nnn \l_tmpa_tl { ~ } { \ }
    \exp_args:NV \seqsplit \l_tmpa_tl
    \group_end:
  }
\NewDocumentCommand{\RelicCode}{m}{\relic_breakable_code:n{#1}}
\ExplSyntaxOff

\usepackage{titletoc}
\usepackage{multicol}

\usepackage{hyperref}
\usepackage{bookmark}
\hypersetup{
    pdftitle={Relic: From Multi-Agent Collaboration to Persistent Organizational Capability},
    bookmarksnumbered=true,
    bookmarksopen=true,
    bookmarksopenlevel=0,
    colorlinks=true,
    citecolor=blue,
    linkcolor=blue,
    urlcolor=blue
}

\newcommand{\RelicCooperSolved}{28}

\usepackage{etoolbox}

\iclrfinalcopy

\begin{document}

\maketitle

\begingroup
\renewcommand{\thefootnote}{\fnsymbol{footnote}}
\footnotetext[1]{%
\footnotesize
Joint first authors; author order reflects relative contribution.
\textdagger\ Team lead.
\textdaggerdbl\ Corresponding authors.
Correspondence:
Hongyi Du, University of Illinois Urbana-Champaign
(\href{mailto:hongyid4@illinois.edu}{\textcolor{black}{hongyid4@illinois.edu}});
Shang Zhu, Together AI
(\href{mailto:shang@together.ai}{\textcolor{black}{shang@together.ai}});
and Jiaxuan You, University of Illinois Urbana-Champaign
(\href{mailto:jiaxuan@illinois.edu}{\textcolor{black}{jiaxuan@illinois.edu}}).
}
\endgroup
\setcounter{footnote}{0}

\begin{abstract}
Multiple agents may often conflict in an organization: for example, one coding agent changes an interface in a repository, but another continues to develop on the old version where existing tests become stale. A conversation can resolve the episode, but when the participants change, what makes the lesson continue to govern the team?
We introduce \textsc{Relic}, which turns recurring collaboration failures into organization-owned, executable protocols. Members reflect on visible work, propose rules, and govern their adoption. Adopted protocols bind triggers, responsibilities, required evidence, and execution consequences to the runtime, while remaining open to revision and retirement. In one traced case, repeated integration friction produces an interface-review rule that governs later pull requests and is revised as work continues.
Across 360 controlled runs over ten software workloads and three models, Relic raises complete-contract delivery from 14.06\% to 19.76\% (+5.71 percentage points) over a matched structured team without the protocol lifecycle, improving all four verified production endpoints in every model stratum. Under fresh-member transfer, behavioral correctness is 25.4\% with no inherited protocol, 34.6\% with the same rules provided as readable text, and 41.2\% with executable bindings, a +6.5-point advantage over text alone. On the full CooperBench benchmark, after excluding 183 broken benchmark pairs, Relic achieves 371/469 (79.1\%), establishing the best reported result among peer-structured systems. On the 47-pair same-model subset, Relic also exceeds Solo (28/47 vs. 26/47), reversing the coordination loss exhibited by the official peer baseline. Together, these results show how collaboration experience can become persistent organizational state that remains useful beyond the members who created it.
\end{abstract}

\section{Introduction}
\label{sec:intro}
\begin{figure}[t]
    \centering
    \includegraphics[width=\textwidth]{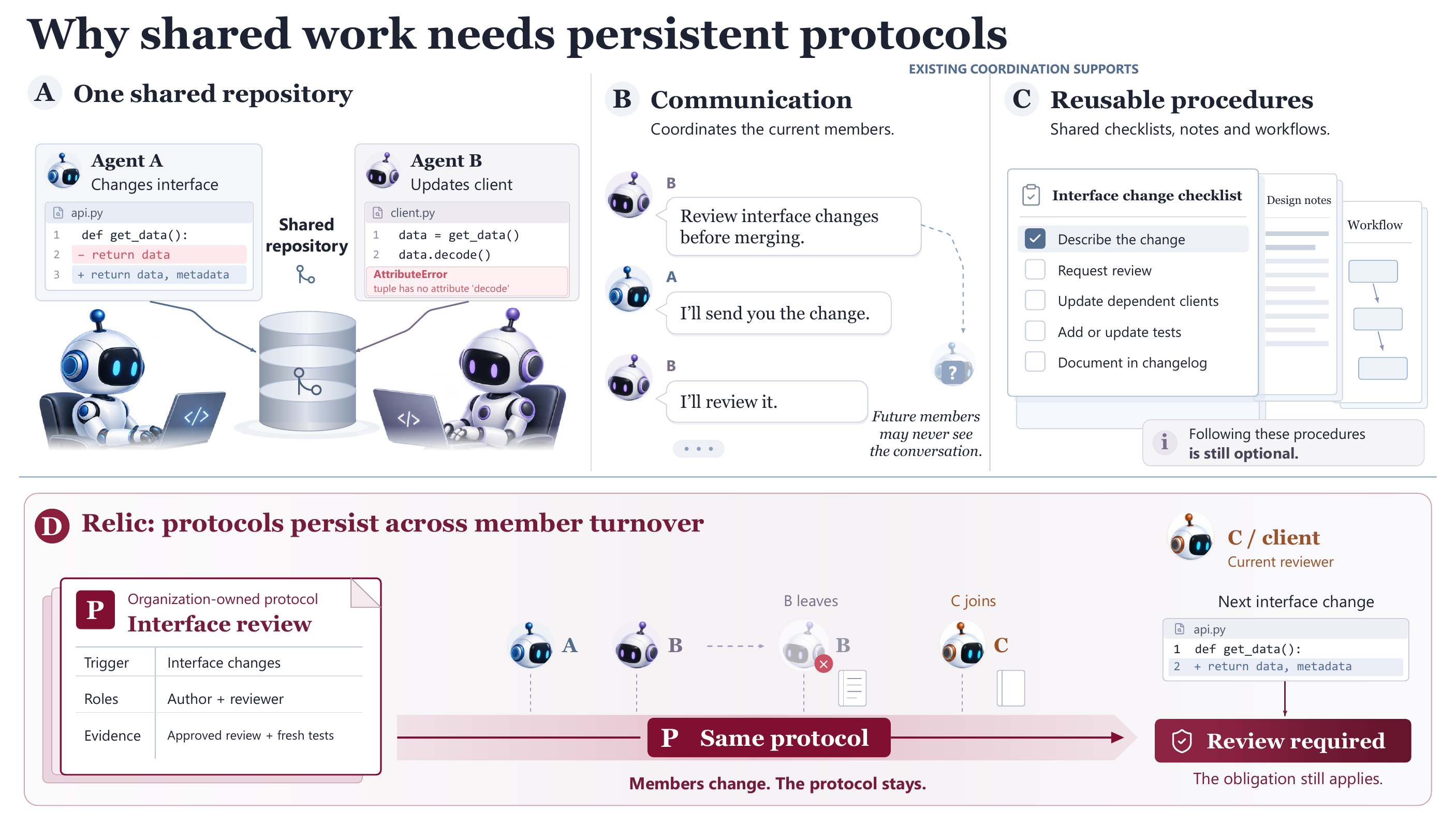}
    \caption{\textbf{Why shared work needs organizational protocols.}
    An illustrative interface change breaks a client developed against the old
    interface. Communication establishes agreements; memory and skills preserve
    knowledge and procedures. Relic retains an organization-owned, runtime-bound
    review protocol: agent B is replaced by agent C, while agent A and shared
    work continue and the review obligation still applies.}
    \label{fig:organizational_gap}
\end{figure}

Consider a developer assigning several coding agents to one repository.
Agent A changes a shared interface, agent B updates a client using the old
interface, and a third agent runs tests. Each makes progress, yet integration
can break the client. Branches may lag hundreds of commits behind the
integrated main branch (\emph{mainline}); overlapping edits can overwrite
another agent's work, and earlier tests may concern outdated code.
Prime Intellect reports a related incident: GPT-5.6 Sol observed unexpected
shared-code changes, suspected subagent interference, and gave research
advisors read-only instructions~\citep{bakouch2026automatedairesearch}.
Shared work needs rules specifying who may modify which artifacts, when to
synchronize, and what evidence integration requires
(Figure~\ref{fig:organizational_gap}).

\textbf{Can communication solve this problem?} ChatDev, MetaGPT, and AutoGen support
role-based dialogue and workflows~\citep{qian2023chatdev,hong2023metagpt,wu2023autogen}.
Suppose A and B agree to synchronize mainline, rerun tests, and obtain review
before merging. Even with perfect compliance, replacing B with C does not
automatically establish the agreement for C. Anthropic reports that fresh
coding-agent sessions lack prior memory; undocumented partial work leaves
successors guessing, while structured handoffs add orchestration
overhead~\citep{young2025longrunningagents,rajasekaran2026harnessdesign}.
Collaborative memory preserves experience~\citep{zhang2025gmemory}; the question
remains how an agreement becomes an obligation for whoever occupies the role.

\textbf{Could reusable skills preserve the agreement?} Shared skill libraries retain
procedures across sessions, including executable code, as in
Voyager~\citep{wang2023voyager}. A testing skill can run a suite; its availability
alone does not assign an independent reviewer or require current evidence
before integration. The additional requirement is to bind learned rules to
shared work: specify when they apply, who is responsible, and how compliance
and revisions are governed. Skills can help members fulfill these obligations.
Long-horizon collaboration thus produces both software and ways of working:
retaining the latter lets an organization apply its experience to future work.

We introduce \textsc{Relic}, which turns collaboration experience into
\textbf{organizational capability}: learned ways of working represented by
governed, executable protocols outside member-local state. Members propose
rules from visible work friction and submit them for validation and approval.
\textbf{Runtime binding} connects adopted rules to the code that selects actions,
assigns responsibilities, and checks shared work. The rules remain revisable
and can govern fresh members without their authors' private histories
(Section~\ref{sec:method}).

\paragraph{Contributions.}
First, we formulate organizational capability as learned, governed, executable
protocol state outside member-local memory, allowing organization-owned rules to
persist across member turnover. Second, across 360 controlled runs spanning
three models, Relic improves pooled complete contracts by 5.71 pp over the
matched structured team, with gains on all four verified endpoints in every
model stratum. Across the Terra+Opus 60-run Relic census, 280 autonomous
protocol lineages enter sustained use. Third, two controls isolate executability: a prose-only ablation
retains online rule proposal and governance but loses 7.18 pp on complete
contracts; content-matched transfer to fresh members gains 6.5 pp from executable
binding over prose alone. Finally, external tests show portability: ProgramBench improves mean behavioral correctness by 6.752 pp, while on the full CooperBench benchmark Relic achieves 371/469 (79.1\%) after excluding 183 broken benchmark pairs, establishing the best reported peer-structured result. On the 47-pair same-model subset, Relic also exceeds Solo (28/47 vs. 26/47), reversing the coordination loss observed for the official peer system.

\section{Related Work}
\label{sec:related}
\paragraph{Communication and shared workflows.}
CAMEL, ChatDev, and AutoGen coordinate specialized agents through dialogue
and delegation~\citep{li2023camel,qian2023chatdev,wu2023autogen}; MetaGPT
incorporates human-designed operating procedures into agent
workflows~\citep{hong2023metagpt}. MultiAgentBench evaluates collaboration,
ProtocolBench compares communication protocols, and CooperBench examines
coordination over interacting code
changes~\citep{zhu2025multiagentbench,du2026protocolbench,cooperbench2026}.
Relic asks how experience from collaboration becomes governed rules for
subsequent work, extending the focus from coordinating actions to learning
how collective work should proceed.

\paragraph{Memory and reusable skills.}
Reflexion and ExpeL retain feedback and extracted experience for later
reasoning~\citep{shinn2023reflexion,zhao2023expel}. G-Memory retrieves
collaboration trajectories and cross-trial
insights~\citep{zhang2025gmemory}, while Voyager accumulates executable
skills~\citep{wang2023voyager}. These mechanisms preserve knowledge and
procedures. Relic focuses on cross-member obligations and their runtime
binding: a skill supplies a testing procedure, while a protocol assigns responsibility
for producing and reviewing the evidence required for shared work.

\paragraph{Self-evolving agent organizations.}
Meta-Team learns improvements to agent behavior, coordination, and team
organization~\citep{metateam2026}. OneManCompany builds an organizational
layer around portable agent identities and dynamic
recruitment~\citep{yu2026onemancompany}; TheBotCompany adapts teams during
continuous software development~\citep{lyu2026botcompany}. Relic focuses on
the learned protocol itself: its evidence-grounded proposal, governed adoption,
execution, and revision. Rule content evolves while model parameters and the
decision and execution substrate remain fixed. Fresh-member transfer and text-only controls test
whether retained protocols remain useful and whether runtime binding adds
value beyond readable rules.

\paragraph{Organizational routines and governance.}
Routines and dynamic capabilities explain how collectives retain and
reconfigure ways of working~\citep{nelson1982evolutionary,feldman2003routines,teece1997dynamic}.
Computer-supported cooperative work (CSCW) examines coordination of
interdependent tasks (articulation work) and artifacts shared across groups
(boundary objects)~\citep{schmidt1992taking,ackerman2000intellectual,lee2007boundary}.
Relic makes learned working rules explicit, revisable organizational objects
outside member-local state. Its evaluation separately measures whether they
enter sustained practice, improve verified outcomes, and remain useful after
member replacement.

\section{\textsc{Relic}: From Experience to Organizational Protocols}
\label{sec:method}
Relic adds a governed protocol lifecycle to long-horizon collaboration.
Members propose rules from observed problems; validation and approval precede
runtime binding, revision, or retirement. Figure~\ref{fig:relic_overview}
follows one interface-review protocol from its origin to execution and
new-member reuse.

\subsection{What an Organizational Protocol Contains}
\label{sec:organizational_state}
Relic separates \textbf{member-local state} (private memory, private workspaces,
and experience), \textbf{shared work state} (code, documents, tasks, and messages),
and \textbf{protocol state} (learned rules and role mappings). Shared work can
outlast a member; protocols carry obligations for later occupants of a role.

A protocol specifies its trigger and scope, responsible roles, required steps
and evidence, affected actions or artifacts, execution consequences, and
revision or retirement state. The example in Figure~\ref{fig:relic_overview}
requires an author and reviewer to check interface changes against a written
interface specification and current test evidence. Its response combines
review priority and readiness checks. Readable text explains the obligations;
runtime bindings apply them.

Members have partial views: only accessible information they read or retrieve
enters their context. Private workspaces and explicit sharing determine how
information reaches each member. Evaluator-side audit records support measurement
and replay separately from member context
(Appendix~\ref{app:state_information}).

\subsection{From Work Friction to Governed Protocols}
\label{sec:capability_formation}
Repeated interface failures can prompt reflection and a proposal naming the
problem, visible evidence, affected work, and responsible roles. Validation
checks evidence grounding and compatibility with supported actions and checks;
approval determines whether the proposal becomes a shared rule. Adoption
compiles its structured specification into runtime bindings. Amendments repeat
this governance process; retirement disables the rule while preserving its
history. Rule content changes online; model parameters and validation/execution code remain fixed.

For measurement, \emph{formation} requires adoption followed by repeated use
across independent organizational use units and distinct times, sustained for
a minimum duration. All revisions of one protocol within a run count as one
\emph{lineage}. Stronger evidence links execution or enforcement by other
members to work-object changes and independently evaluated outcomes. These
records establish sustained practice; controlled outcome comparisons test its
benefit (Appendix~\ref{app:formation_criterion}; Algorithm~\ref{alg:a3}).

\subsection{How Protocols Govern Later Work}
\label{sec:structured_decision}
\begin{figure}[!htbp]
    \centering
    \includegraphics[width=\textwidth]{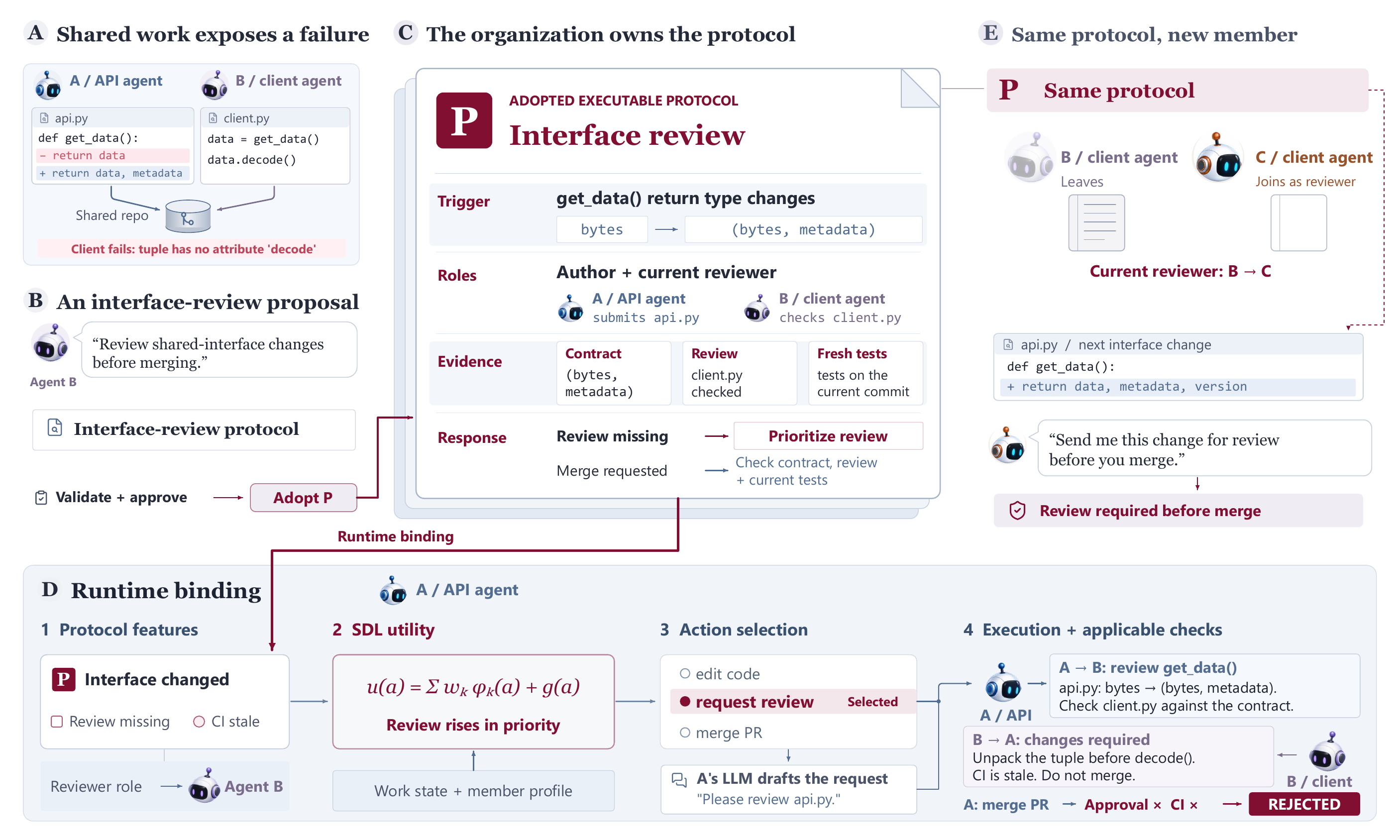}
    \caption{\textbf{How Relic forms, binds, and retains a protocol.}
    Visible work evidence becomes an organization-owned protocol through
    governed adoption. Protocol and work/member features feed SDL scoring
    (Equation~\ref{eq:sdl-decision}); role mappings and applicable checks connect
    selected actions and generated content to shared work. The same protocol P
    governs a replacement reviewer with empty private memory.}
    \label{fig:relic_overview}
\end{figure}

At scheduling step $t$, member $i$ has feasible actions $\mathcal{C}_{i,t}$
constructed from member-visible state. The \textbf{Structured Decision Layer (SDL)}
scores them from work, member, and applicable protocol features
(Figure~\ref{fig:relic_overview}). Its stochastic branch converts these
scores into selection probabilities:
\begin{equation}
\begin{aligned}
 u_{i,t}(a)
 &= \sum_{k=1}^{32} w_k(i,t)\,\phi_k(a,x_{i,t}) + g_{i,t}(a), \\
 \pi_{i,t}(a)
 &= \frac{\exp\!\bigl((u_{i,t}(a)+\epsilon_{i,t,a})/\tau\bigr)}
 {\sum_{a'\in\mathcal{C}_{i,t}}
  \exp\!\bigl((u_{i,t}(a')+\epsilon_{i,t,a'})/\tau\bigr)}.
\end{aligned}
\label{eq:sdl-decision}
\end{equation}
Here $x_{i,t}$ contains visible work and applicable protocols; $\phi_k$
measures progress, role fit, coordination, evidence, or protocol
obligations. Weights $w_k(i,t)$ combine fixed coefficients with member
profile/skill inputs. The adjustment $g_{i,t}$
combines authority, reputation, coding preference, and behavioral constraints.
Stochastic selection samples from $\pi_{i,t}$, with temperature $\tau$ and
reproducible random perturbations $\epsilon_{i,t,a}$; deterministic selection
chooses a highest-scoring action. Figure~\ref{fig:relic_overview} abbreviates
the first line of Equation~\ref{eq:sdl-decision} as $u=\sum w\phi+g$.

\paragraph{From protocol to decision.}
An adopted interface-review rule supplies responsibility and evidence
obligations to features $\phi_k$. These can change relative utilities $u_{i,t}$
and probabilities $\pi_{i,t}$ in Equation~\ref{eq:sdl-decision}, prioritizing
review or evidence work. The scoring map stays fixed; protocols change its
inputs. Skill-backed inputs also evolve through the shared member-learning path.

\paragraph{From decision to execution.}
Role mappings identify who acts. When an action needs code or messages, the
LLM supplies them to its handler---the code implementing that action. Routine
selection itself needs no LLM call. SDL scores set action priorities;
applicable execution checks update or reject work-object transitions, and
protocol events record subsequent use or enforcement. The action registry and handlers
remain fixed as protocols evolve
(Appendices~\ref{app:sdl_details} and~\ref{app:execution_operator}).

\subsection{Member-Independent Reuse}
\label{sec:member_independent_reuse}
Figure~\ref{fig:relic_overview} retains protocol P as reviewer B leaves and C
joins with empty private memory: the obligation follows the role. Replacement
resets private memory, workspaces, sandboxes, message-read state, commitments,
reflections, wishes, and accumulated experience. Shared work can remain
(Figure~\ref{fig:organizational_gap}). The reported transfer experiment goes
further: a fresh target product receives only the designated protocols and
role mappings, without source-product code, source-world state, or private
histories.

\textbf{Text} supplies the frozen rules as reusable prose to model-mediated
work, including code editing; \textbf{Exec} exposes the same readable content
and additionally compiles it into runtime bindings. Their comparison isolates
the added value of executable organizational binding on the shared SDL backbone
(Appendix~\ref{app:transfer}).

\paragraph{Controlled configurations.}
We use \textbf{B0} for a reflective single agent and \textbf{B1} for an
eight-member long-horizon team with shared work and peer review. \textbf{B2}
adds SDL selection, profile conditioning, and member-local updates to skills,
reputation, and authority; its protocol lifecycle is disabled. \textsc{Relic}
(\textbf{B3}) shares this backbone and enables the governed protocol lifecycle:
members propose rules from recurring friction; validated and approved rules
become persistent organizational protocols whose bindings affect later action
selection, responsibility routing, and evidence checks. The selector, model
parameters, and execution code remain fixed. B3--B2 tests this added lifecycle.

\section{Evaluation: Formation, Effectiveness, and Transfer}
\label{sec:experiments}
Persistent work, interacting tasks, and recurring coordination demands motivate
our evaluation of formation, verified delivery, executable binding, fresh-member
transfer, and external portability. The main study tests multi-member software
production with B0 as the single-agent reference; ProgramBench attaches a frozen
protocol layer to a single coding agent; CooperBench evaluates two-agent delivery
of interacting features.

\begin{figure}[!htbp]
\centering
\includegraphics[width=\linewidth,trim=20 10 20 10,clip]{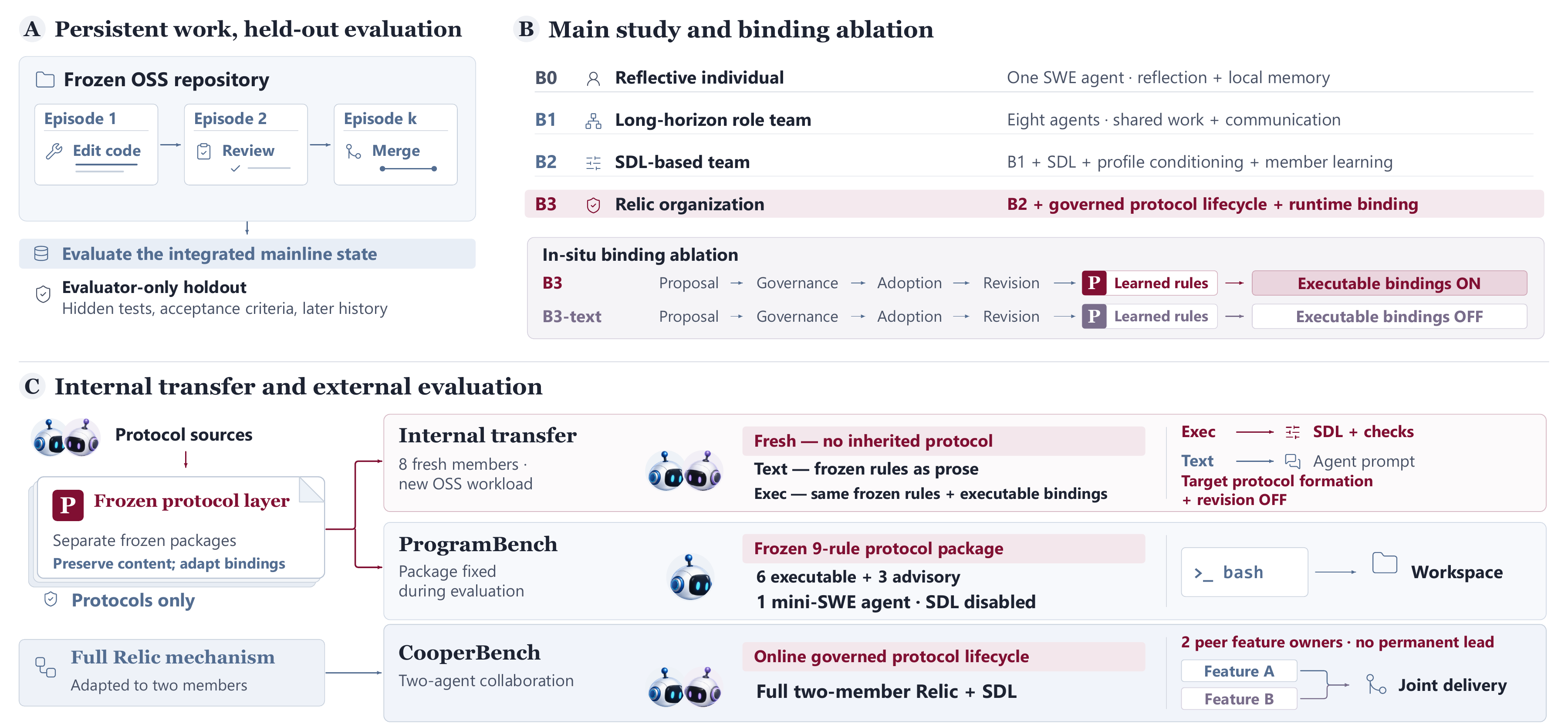}
\caption{\textbf{Evaluation design.} B0--B3 share frozen workloads and 336
steps. B3-text retains governed rule creation and readable rules but removes bindings;
Text/Exec hold rule content fixed. ProgramBench tests protocol portability
without SDL; CooperBench tests two-agent interacting-feature delivery.}
\label{fig:evaluation-design}
\end{figure}

\paragraph{Workloads and outcomes.}
Five tasks build specified programs from starter skeletons; five repair or
extend frozen repository versions toward later-version requirements. Their
interdependent implementation, testing, review, and integration create repeated
coordination demands. Related acceptance cases form a \emph{contract}, complete
only when every case passes on mainline. \emph{Exposed} and \emph{held-out}
cases test visible and withheld requirements. \emph{Evaluator-confirmed seeded
issues} are initial benchmark problems verified as resolved relative to failing
starter behavior; ``seeded'' concerns task construction, not repeated-run seeds.
Workspace scores separate local changes from integrated delivery. Private
acceptance tests and reference implementations remain hidden
(Appendix~\ref{app:benchmark-curation}).

\paragraph{Conditions and design.}
GPT-5.6 Terra, Claude Opus 4.6, and DeepSeek-V4-Flash each run ten workloads
with three random seeds under B0--B3, totaling 360 controlled runs. Relic (B3)
and the structured control (B2) share model, tools,
task-visible information, SDL parameters, member learning, and execution code;
their comparison tests the governed protocol lifecycle. All conditions use 336 scheduling steps, with token expenditure recorded for each run
(Appendices~\ref{app:conditions} and~\ref{app:time_resources}).

\paragraph{Binding, transfer, and external evidence.}
A separate 30-run Claude Opus 4.6 ablation compares B3 with \textbf{B3-text},
its prose-only variant, which retains proposal, governance, adoption, revision,
and readable rules but removes learned-protocol executable bindings. Text and Exec then provide a content-matched test:
each adds 30 GPT-5.6 Terra runs with fresh members and the same frozen protocol
package; only Exec receives executable bindings, while new target protocol proposal, adoption, and revision are disabled. Fresh reuses the 30 corresponding B2 runs.
ProgramBench compares mini-SWE-agent with and without executable protocols on
the same 25 tasks, without SDL. CooperBench evaluates the complete two-member
Relic architecture on the full benchmark against released peer and Team references,
while a 47-pair same-model subset provides the direct Solo--Peer coordination comparison.

\FloatBarrier
\section{Results}
\label{sec:results}
We first test verified delivery, then examine learned protocols, runtime
binding, and transfer.

\begin{table}[!htbp]
\centering
\begingroup
\arrayrulecolor{RuleGray}
\newcommand{\mainestci}[2]{\shortstack[c]{#1\\[-0.45ex]{\normalfont\tiny\textcolor{TableMuted}{[#2]}}}}
\newcommand{\maineffectci}[2]{\effectcell{\mainestci{#1}{#2}}}
    {
    \small
    \setlength{\tabcolsep}{2.0pt}
    \renewcommand{\arraystretch}{1.12}

    \begin{tabularx}{\textwidth}{
        @{}
        >{\raggedright\arraybackslash}X
        >{\centering\arraybackslash}m{0.105\textwidth}
        >{\centering\arraybackslash}m{0.105\textwidth}
        >{\centering\arraybackslash}m{0.105\textwidth}
        >{\centering\arraybackslash}m{0.105\textwidth}
        >{\centering\arraybackslash}m{0.145\textwidth}
        @{}
    }
        \toprule
        \rowcolor{PanelFill}
        \multicolumn{6}{@{}l@{}}{\normalsize\textbf{Verified software-production outcomes --- 360 runs}} \\
        \addlinespace[1pt]
        \textbf{Metric}
        & \armhead{B0Dark}{B0}
        & \armhead{B1Dark}{B1}
        & \armhead{B2Dark}{B2}
        & \armhead{B3Dark}{B3}
        & \textcolor{B3Dark}{\textbf{B3--B2}} \\
        \midrule

        Complete contracts on mainline $\uparrow$
        & \mainestci{8.22\%}{4.69, 12.32}
        & \mainestci{15.97\%}{11.01, 21.57}
        & \mainestci{14.06\%}{9.92, 21.09}
        & \bestcell{\mainestci{19.76\%}{14.23, 25.92}}
        & \maineffectci{+5.71 pp}{2.97, 8.55} \\

        Held-out cases on mainline $\uparrow$
        & \mainestci{6.86\%}{0, 16.79}
        & \mainestci{8.17\%}{0, 20.88}
        & \mainestci{9.57\%}{1.16, 26.48}
        & \bestcell{\mainestci{17.28\%}{4.46, 45.02}}
        & \maineffectci{+7.72 pp}{2.67, 17.63} \\

        Exposed cases on mainline $\uparrow$
        & \mainestci{11.97\%}{6.72, 18.16}
        & \mainestci{22.56\%}{14.95, 31.40}
        & \mainestci{19.31\%}{14.00, 31.94}
        & \bestcell{\mainestci{25.89\%}{17.47, 35.09}}
        & \maineffectci{+6.58 pp}{3.41, 10.14} \\

        Evaluator-confirmed seeded issues $\uparrow$
        & \mainestci{10.69\%}{6.02, 16.18}
        & \mainestci{20.46\%}{13.73, 28.25}
        & \mainestci{17.91\%}{12.16, 24.16}
        & \bestcell{\mainestci{24.87\%}{17.30, 33.14}}
        & \maineffectci{+6.96 pp}{4.15, 10.83} \\

        \addlinespace[2pt]
        Avg. tokens / run $\downarrow$
        & \bestcell{\mainestci{3.385M}{3.216, 3.565}}
        & \mainestci{45.200M}{42.993, 47.550}
        & \mainestci{4.274M}{3.861, 4.693}
        & \mainestci{7.259M}{6.278, 8.510}
        & \maineffectci{+2.985M}{1.870, 4.364} \\

        Tokens / confirmed issue $\downarrow$
        & \mainestci{4.757M}{3.404, 5.449}
        & \mainestci{40.069M}{32.443, 49.793}
        & \bestcell{\mainestci{4.010M}{3.179, 5.327}}
        & \mainestci{5.630M}{3.994, 7.508}
        & \maineffectci{$-$0.448M}{$-$1.573, +0.659} \\

        \bottomrule
    \end{tabularx}
    }

\caption{\textbf{Main study (three models; 360 runs).} Brackets: 95\% CIs. B3--B2 is paired; cost/issue means use arm-specific eligible blocks, with 20 common blocks for the contrast.}
\label{tab:primary-results}

\endgroup
\arrayrulecolor{black}
\end{table}

\subsection{Executable Protocols Improve Verified Delivery}
\label{sec:results_outcomes}
\begin{figure}[!htb]
\centering
\includegraphics[width=\textwidth]{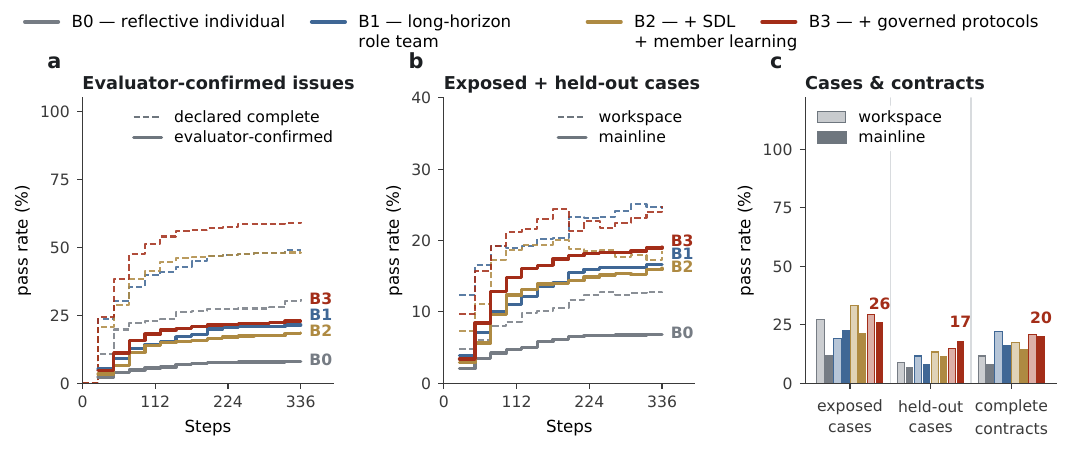}
\caption{\textbf{Verified production outcomes across three models.} Declared and verified completion, workspace and mainline trajectories, and endpoint pass rates.}
\label{fig:verified-production-outcomes}
\end{figure}
Relic \textbf{improves all four verified endpoints} over the structured control
(Table~\ref{tab:primary-results}). Across all 360 runs, complete contracts rise
from 14.06\% to 19.76\% (+5.71 pp, 95\% CI [2.97, 8.55]); held-out,
exposed, and confirmed-issue gains are +7.72 [2.67, 17.63], +6.58 [3.41, 10.14], and
+6.96 [4.15, 10.83] pp, respectively. Every endpoint also has a positive
B3--B2 point estimate in each of the three model strata
(Table~\ref{tab:model_stratified_effects}).
Appendix~\ref{app:robustness} reports the robustness analyses.

Figure~\ref{fig:verified-production-outcomes} tracks declarations, verified mainline delivery, and locally passing work across all three models.

\Needspace{5\baselineskip}
\subsection{What Protocols Does the Organization Learn?}
\label{sec:results_mechanisms}
Across the Terra+Opus 60-run Relic census, we record 497 autonomous
protocol proposals, counting one rule and all its revisions as a single within-run lineage. Of these,
393 remain adopted at the endpoint and 280 satisfy the sustained-use criterion
in Section~\ref{sec:capability_formation}; 32 additionally link execution by
other members to shared-state changes and independently evaluated outcomes.
Review/merge, release engineering, and evidence governance account for 87.1\%
of formed lineages. These rules organize the transition from local changes to
verified shared delivery. Appendix~\ref{app:capability_census} gives the lineage census.

\begin{figure}[!htb]
\centering
\includegraphics[width=\textwidth]{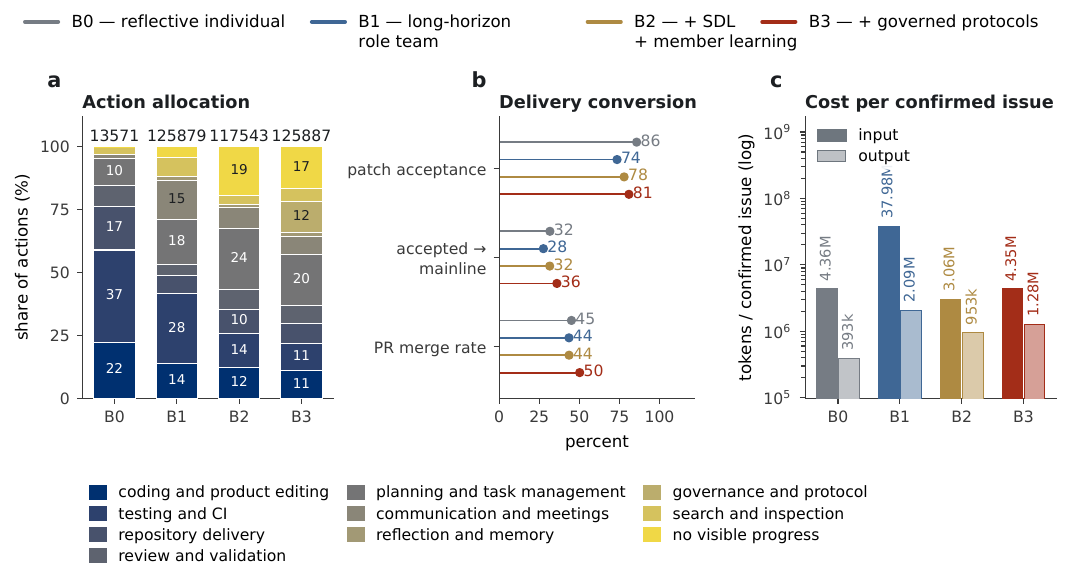}
\caption{\textbf{Work allocation, delivery, and resource use across three models.}
Action shares, delivery conversion, and input/output tokens per
evaluator-confirmed issue across B0--B3.}
\label{fig:operative-mechanisms}
\end{figure}

Figure~\ref{fig:operative-mechanisms} links these rules to work allocation
and delivery. Relic moves 36\% of accepted work to mainline versus 32\% for the
structured control; its pull-request (PR) merge rate is 50\% versus 44\%.
The single agent has the highest patch-acceptance rate (86\%, versus Relic's
81\%) but the lowest complete-contract score. Local acceptance alone does not
ensure delivery. Section~\ref{sec:case_commit_bound} traces one protocol's
content and execution.

Across the three-model pooled accounting, Relic averages 7.259M tokens/run
versus the control's 4.274M (+2.985M, 95\% CI [1.870M, 4.364M]). Tokens per evaluator-confirmed issue average 5.630M for B3 and 4.010M for B2 over each arm's eligible model--workload blocks; the paired B3--B2 estimate over 20 common model--workload blocks is $-$0.448M (95\% CI [$-$1.573M, +0.659M]; Appendix~\ref{app:efficiency_metrics}). The Terra+Opus matched-spend analysis additionally selects
the last saved Relic checkpoint within each matched control run's realized token
spend. Under this matched-spend sensitivity, all four verified endpoints remain
positive: +7.45 pp for complete contracts, +5.08 pp for held-out cases,
+11.72 pp for exposed cases, and +6.21 pp for evaluator-confirmed issues
(Appendix~\ref{app:budget_capped_sensitivity}).

\subsection{Executable Binding and Member-Independent Transfer}
\label{sec:results_transfer}
Table~\ref{tab:transfer-current} compares fresh members receiving no inherited
rules (Fresh), a frozen protocol package as prose (Text), or identical content
with runtime bindings (Exec). Source selection, runtime adaptation, and wording
are matched between Text and Exec; neither learns or revises target protocols.

\begin{table}[!htbp]
\centering
\begingroup
\small
\arrayrulecolor{RuleGray}
\setlength{\tabcolsep}{2.0pt}
\renewcommand{\arraystretch}{1.16}
\newcommand{\estci}[2]{\shortstack[c]{#1\\[-0.45ex]{\normalfont\tiny\textcolor{TableMuted}{[#2]}}}}
\newcommand{\bestestci}[2]{\bestcell{\estci{#1}{#2}}}
\newcommand{\effectci}[2]{\effectcell{\estci{#1}{#2}}}
\begin{tabularx}{\textwidth}{
        @{}
        >{\raggedright\arraybackslash}X
        *{3}{>{\centering\arraybackslash}m{0.105\textwidth}}
        *{2}{>{\centering\arraybackslash}m{0.185\textwidth}}
        @{}
    }
        \toprule

        \rowcolor{PanelFill}
        \multicolumn{6}{@{}l@{}}{
            \normalsize\textbf{Internal protocol transfer}
        } \\

        \addlinespace[1pt]

        \textbf{Metric}
        & \armhead{ScratchDark}{Fresh}
        & \armhead{TextDark}{Text}
        & \armhead{ExecDark}{Exec}
        & \textcolor{ExecDark}{\textbf{Exec--Fresh}}
        & \textcolor{ExecDark}{\textbf{Exec--Text}} \\

        \midrule

        Behavioral-case pass rate $\uparrow$
        & \estci{25.4\%}{13.2\%, 41.1\%}
        & \estci{34.6\%}{24.1\%, 46.4\%}
        & \bestestci{41.2\%}{23.4\%, 54.5\%}
        & \effectci{+15.8 pp}{9.2, 31.8}
        & \effectci{+6.5 pp}{0.7, 15.6} \\

        Avg. tokens / run (M) $\downarrow$
        & \bestestci{2.381}{1.747, 3.189}
        & \estci{2.615}{2.213, 3.033}
        & \estci{2.448}{2.199, 2.720}
        & \effectci{+0.067}{$-$0.569, +0.532}
        & \effectci{$-$0.167}{$-$0.467, +0.073} \\

        Tokens / evaluated case $\downarrow$
        & \estci{176k}{106k, 271k}
        & \estci{154k}{96k, 265k}
        & \bestestci{144k}{94k, 243k}
        & \effectci{$-$32k}{$-$112k, 31k}
        & \effectci{$-$10k}{$-$33k, 4k} \\

        Tokens / verified pass $\downarrow$
        & \estci{691k}{488k, 1,419k}
        & \estci{444k}{271k, 935k}
        & \bestestci{350k}{196k, 958k}
        & \effectci{$-$342k}{$-$818k, $-$14k}
        & \effectci{$-$94k}{$-$152k, 54k} \\

        \bottomrule
    \end{tabularx}
\caption{\textbf{Internal protocol transfer.} Brackets give 95\% confidence intervals; Exec--Fresh and Exec--Text report Exec minus Fresh and Exec minus Text, respectively.}
\label{tab:transfer-current}
\endgroup
\arrayrulecolor{black}
\end{table}

Exec passes 41.2\% of behavioral cases, exceeding Text's
34.6\% by 6.5 pp (95\% confidence interval [0.7, 15.6]) and Fresh's
25.4\% by 15.8 pp.
\textbf{Runtime binding adds value to the same readable content}; Table~\ref{tab:transfer-current}
reports costs, with complete outcomes in Appendix~\ref{app:all_transfer_results}.

\paragraph{Binding ablation during rule creation.}
A separate 30-run Claude Opus 4.6 experiment compares Relic with its prose-only
variant (B3-text) on matched workloads and random seeds. The variant retains
proposal, governance, adoption, revision, and readable rules but removes
learned-protocol bindings. Complete contracts fall from 22.20\% to 15.03\%,
a +7.18 pp paired advantage for Relic (95\% confidence interval [+3.54, +10.91]).
Held-out, exposed, and confirmed-issue outcomes have the same direction.
Every B3-text run ends with readable rules; learned-protocol bindings, automatic
use logging, and protocol-specific enforcement are disabled. Both conditions
independently develop their rules from $t=0$, while Text/Exec transfer holds
realized content fixed. Appendix~\ref{app:insitu_binding_ablation} reports all
four endpoint intervals and the binding manipulation check.

\subsection{External Software Benchmark Extensions}
\label{sec:external-benchmarks}
\label{sec:programbench}
\label{sec:cooperbench}

\begin{table}[H]
\centering
\begingroup
\small
\arrayrulecolor{RuleGray}
\setlength{\tabcolsep}{3.4pt}
\renewcommand{\arraystretch}{1.00}
\begin{tabularx}{\linewidth}{@{}L >{\centering\arraybackslash}m{0.19\linewidth} >{\centering\arraybackslash}m{0.24\linewidth}@{}}
\toprule

\rowcolor{PanelSubFill}
\multicolumn{3}{@{}l@{}}{\textbf{A. CooperBench --- full benchmark}}\\
\textbf{System} & \textbf{Successful pairs} & \textbf{Role in comparison}\\
\midrule
\armhead{ExecDark}{Relic, 2-member} &
\bestcell{\textbf{371/469 (79.1\%)}} &
\bestcell{peer organization; no fixed lead}\\
Released Peer / coop+git &
277/469 (59.1\%) &
strongest released peer reference\\
Team no-proto &
349/469 (74.4\%) &
strongest released lead--member reference\\

\addlinespace[1pt]
\rowcolor{PanelSubFill}
\multicolumn{3}{@{}l@{}}{\textbf{B. CooperBench --- same-model coordination check (47 final-valid pairs)}}\\
\textbf{System} & \textbf{Successful pairs} & \textbf{Role in comparison}\\
\midrule
\armhead{ExecDark}{Relic, 2-member (Claude Opus 4.6)} &
\bestcell{28/47} &
\bestcell{two peer feature owners; no fixed lead}\\
Official Solo (Claude Opus 4.6) &
26/47 &
one agent; both features\\
Official Peer (Claude Opus 4.6) &
13/47 &
two peers; split features\\

\addlinespace[1pt]
\rowcolor{PanelSubFill}
\multicolumn{3}{@{}l@{}}{\textbf{C. ProgramBench --- same 25 tasks}}\\
\textbf{System} & \multicolumn{2}{c}{\textbf{Mean behavioral pass}}\\
\midrule
\armhead{ScratchDark}{Official mini-SWE-agent} &
\multicolumn{2}{c}{64.164\%}\\
\armhead{ExecDark}{+ executable protocols} &
\multicolumn{2}{c}{\bestcell{70.916\%}}\\

\bottomrule
\end{tabularx}
\caption{\textbf{External benchmark extensions.}
CooperBench full-benchmark results exclude broken benchmark pairs under the
same exclusion set for every compared system; audit details and raw results
are reported in Appendix~\ref{app:cooperbench}. ProgramBench settings are
reported in Appendix~\ref{app:programbench_external}.}
\label{tab:external-benchmarks}
\label{tab:cooperbench-main}
\label{tab:programbench}
\endgroup
\arrayrulecolor{black}
\end{table}

On \textbf{CooperBench}, Relic evaluates the full benchmark with two peer
feature owners and no permanent lead. After excluding broken benchmark pairs
under the same exclusion set for every compared system, Relic reaches
\textbf{371/469 (79.1\%)}, compared with 277/469 (59.1\%) for the strongest
released peer reference and 349/469 (74.4\%) for the strongest released
hierarchical Team reference. Relic therefore exceeds the released peer
reference by 20.0 pp and the hierarchical Team reference by 4.7 pp. On the
47-pair same-model subset, Official Peer
falls from Solo's 26/47 to 13/47, whereas Relic reaches 28/47, reversing the \emph{coordination curse}
on this controlled comparison.

On \textbf{ProgramBench}, six executable and three advisory rules cover
originality, evidence freshness, verification, and submission readiness.
This frozen package raises mini-SWE-agent's mean behavioral pass rate on 25
tasks from 64.164\% to 70.916\% (+6.752 pp; +10.5\% relative;
Appendix~\ref{app:programbench_external}).

\section{Case Study: From Repeated Friction to a Governing Rule}
\label{sec:case_commit_bound}
\begingroup
In one matched main-study case, we trace an integration problem into a
learned protocol, holding model, random seed, tools, and horizon fixed.

\begin{figure}[!htbp]
    \centering
    \includegraphics[
        width=\textwidth,
        trim=0 6mm 0 0,
        clip
    ]{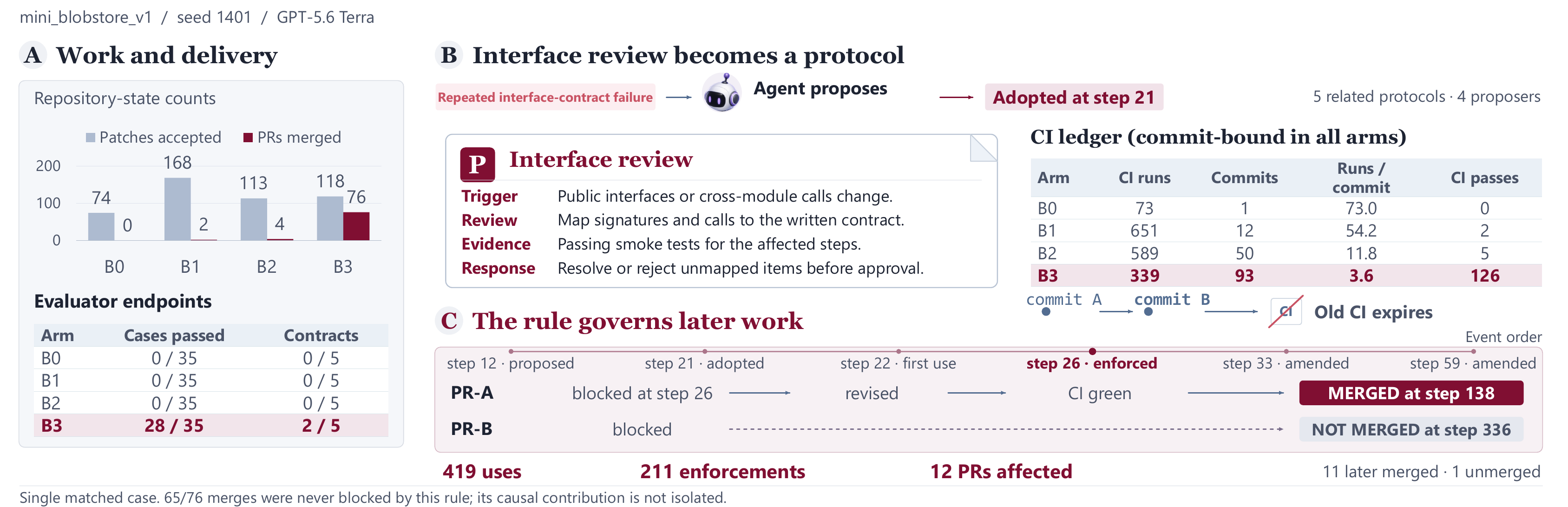}
    \caption{\textbf{What the organization learned.} A matched unit shows local delivery, convergent verification rules, and one rule's enforcement and revision. $t12$ denotes scheduling step 12.}
    \label{fig:commit-bound-verification-case}
\end{figure}

\paragraph{Before: activity without sustained delivery.}
The role-based team (B1) accepted 168 patches but merged only two PRs; Relic
accepted 118 and merged 76 (Figure~\ref{fig:commit-bound-verification-case}a).
Local activity therefore did not reliably become shared delivery.

\paragraph{Formation: repeated friction becomes a rule.}
Repeated interface and verification failures produced an interface-review
protocol requiring changed public interfaces to be mapped to the written
contract and backed by current test evidence before approval
(Appendix~\ref{app:five_rules}).

\paragraph{After: the rule governs later work.}
The protocol was proposed at step~12, adopted at step~21, first enforced at
step~26, and amended at steps~33 and~59. It records 211 enforcements across
12 PRs and 419 uses through step~336; 11 affected PRs later merged, including
PR-A after its initial block (Figure~\ref{fig:commit-bound-verification-case}c;
Appendix~\ref{app:case_commit_bound}).

\endgroup
\FloatBarrier

\section{Discussion, Future Work, and Conclusion}

\paragraph{Discussion.}
Relic represents organizational capability as shared rules that are formed from
experience, executed, revised, and reused by later members. B3--B2 tests the
governed lifecycle, B3-text tests its executable bindings, and content-matched
Text/Exec tests reuse by fresh members. B3 and Transfer Exec also instantiate
learned and fixed-rule regimes. Verified gains recur across model strata,
persist in the two-model workload-omission and matched-spend analyses, and
extend to external software benchmarks. Rule quality, revision, and retirement
shape how this organizational layer develops over time
(Appendix~\ref{app:broader_discussion}).

\paragraph{Future work.}
Our evaluation focuses on software-production settings; broader domains, selective protocol transfer and forgetting, rule-quality diagnosis, alternative decision layers, and human--organization interaction remain important directions (Appendix~\ref{app:research_agenda}).

\paragraph{Conclusion.}
Relic turns collaboration experience into governed, executable protocols held
outside member-local histories. Across the main study, binding ablation,
content-matched transfer, fresh-member reuse, and external benchmarks, the
results support persistent executable organizational state as a useful layer
for long-horizon, persistent multi-agent collaboration. The organizational state remains explicit and revisable.
\let\section\RelicOrigSection
\let\subsection\RelicOrigSubsection
\let\paragraph\RelicOrigParagraph
\setlength{\floatsep}{\RelicOrigFloatsep}
\setlength{\dblfloatsep}{\RelicOrigDblfloatsep}
\setlength{\intextsep}{\RelicOrigIntextsep}
\setlength{\abovedisplayskip}{\RelicOrigAboveDisplaySkip}
\setlength{\belowdisplayskip}{\RelicOrigBelowDisplaySkip}
\setlength{\abovedisplayshortskip}{\RelicOrigAboveDisplayShortSkip}
\setlength{\belowdisplayshortskip}{\RelicOrigBelowDisplayShortSkip}
\clearpage
\pdfbookmark[1]{Ethics Statement}{relic.ethics-statement}
\subsection*{Ethics Statement}
The human-centered component of this work consists of formative paired walkthroughs of the P2 and P3 interfaces by four members of the author team who did not participate in the design or implementation of Relic or the two interfaces. Participants were informed in advance that their interaction experience and feedback would be used for research purposes and reported as part of this study. These walkthroughs are reported as formative internal observations of the implemented interfaces.

\pdfbookmark[1]{Reproducibility Statement}{relic.reproducibility-statement}
\subsection*{Reproducibility Statement}
The appendices document the experimental conditions, frozen workload construction and qualification, model and decoding configurations, random seeds, run protocol, prompts, protocol representations, metric definitions, and statistical procedures used in the reported experiments. They additionally specify evaluator isolation, artifact and configuration identities, checkpointing and replay, retry and recovery policies, and condition-conformance checks. The CooperBench and ProgramBench extensions are documented separately with their fixed evaluation sets, system configurations, and scoring procedures. These materials are intended to support reproduction and inspection of the reported experimental settings, comparisons, and analyses.

\pdfbookmark[1]{AI Use Statement}{relic.ai-use-statement}
\subsection*{AI Use Statement}
Generative AI tools, including ChatGPT, Codex, Claude, and Claude Code, assisted with benchmark construction, methodological and experimental design, implementation and debugging, analysis of experimental outputs, and manuscript drafting and editing. For the constructed workloads, AI assistance included producing technical specifications, reference implementations, tests, and verifiers from human-selected functional briefs; retained benchmark cases were subsequently qualified against frozen starter and reference implementations before evaluation. The authors reviewed AI-assisted artifacts and analyses, verified the reported results, and made all final research decisions. The authors take responsibility for the manuscript and associated research artifacts.

\pdfbookmark[1]{References}{relic.references}
\bibliography{references}

@article{ackerman2000intellectual,
  author  = {Mark S. Ackerman},
  title   = {The Intellectual Challenge of {CSCW}: The Gap Between Social Requirements and Technical Feasibility},
  journal = {Human--Computer Interaction},
  volume  = {15},
  number  = {2--3},
  pages   = {179--203},
  year    = {2000},
  doi     = {10.1207/S15327051HCI1523_5}
}

@misc{bakouch2026automatedairesearch,
  author       = {Elie Bakouch and {Prime Intellect}},
  title        = {Measuring Autonomous {AI} Research},
  howpublished = {Prime Intellect Blog},
  year         = {2026},
  month        = aug,
  url          = {https://www.primeintellect.ai/blog/measuring-autonomous-research},
  note         = {Published August 14, 2026}
}

@misc{cooperbench_code_2026,
  author       = {{CooperBench}},
  title        = {{CooperBench}: Official Implementation},
  howpublished = {GitHub repository},
  year         = {2026},
  url          = {https://github.com/cooperbench/CooperBench},
  note         = {Accessed 2026-09-14}
}

@misc{cooperbench_team_trajectories_2026,
  author       = {{CooperBench}},
  title        = {{CooperBench} Coordination Study: Agent Trajectories},
  howpublished = {Hugging Face Dataset},
  year         = {2026},
  url          = {https://huggingface.co/datasets/CooperBench/team-trajectories},
  note         = {Accessed 2026-09-14}
}

@inproceedings{du2026protocolbench,
  author    = {Hongyi Du and Jiaqi Su and Jisen Li and Lijie Ding and Yingxuan Yang and Peixuan Han and Xiangru Tang and Kunlun Zhu and Jiaxuan You},
  title     = {{ProtocolBench}: Which {LLM} {MultiAgent} Protocol to Choose?},
  booktitle = {Proceedings of the 43rd International Conference on Machine Learning},
  year      = {2026},
  url       = {https://arxiv.org/abs/2510.17149}
}

@article{feldman2003routines,
  author  = {Martha S. Feldman and Brian T. Pentland},
  title   = {Reconceptualizing Organizational Routines as a Source of Flexibility and Change},
  journal = {Administrative Science Quarterly},
  volume  = {48},
  number  = {1},
  pages   = {94--118},
  year    = {2003},
  doi     = {10.2307/3556620}
}

@misc{metateam2026,
  author        = {Zhezheng Hao and Tianfu Wang and Huanshuo Dong and Ziyan Liu and Hong Wang and Xiankun Lin and Qiang Lin and Can Wang and Hande Dong and Jiawei Chen},
  title         = {Evolve as a Team: Collaborative Self-Evolution for {LLM}-based Multi-Agent Systems},
  year          = {2026},
  eprint        = {2605.29790},
  archivePrefix = {arXiv},
  url           = {https://arxiv.org/abs/2605.29790}
}

@article{hong2023metagpt,
  author  = {Sirui Hong and Mingchen Zhuge and Jiaqi Chen and Xiawu Zheng and Yuheng Cheng and Ceyao Zhang and Jinlin Wang and Zili Wang and Steven Ka Shing Yau and Zijuan Lin and Liyang Zhou and Chenyu Ran and Lingfeng Xiao and Chenglin Wu and J{\"u}rgen Schmidhuber},
  title   = {{MetaGPT}: Meta Programming for a Multi-Agent Collaborative Framework},
  journal = {arXiv preprint arXiv:2308.00352},
  year    = {2023},
  url     = {https://arxiv.org/abs/2308.00352}
}

@misc{cooperbench2026,
  author        = {Arpandeep Khatua and Hao Zhu and Peter Tran and Arya Prabhudesai and Frederic Sadrieh and Johann K. Lieberwirth and Xinkai Yu and Yicheng Fu and Michael J. Ryan and Jiaxin Pei and Diyi Yang},
  title         = {{CooperBench}: Why Coding Agents Cannot Be Your Teammates Yet},
  year          = {2026},
  eprint        = {2601.13295},
  archivePrefix = {arXiv},
  url           = {https://arxiv.org/abs/2601.13295}
}

@article{lee2007boundary,
  author  = {Charlotte P. Lee},
  title   = {Boundary Negotiating Artifacts: Unbinding the Routine of Boundary Objects and Embracing Chaos in Collaborative Work},
  journal = {Computer Supported Cooperative Work},
  volume  = {16},
  number  = {3},
  pages   = {307--339},
  year    = {2007},
  doi     = {10.1007/s10606-007-9044-5}
}

@article{li2023camel,
  author  = {Guohao Li and Hasan Abed Al Kader Hammoud and Hani Itani and Dmitrii Khizbullin and Bernard Ghanem},
  title   = {{CAMEL}: Communicative Agents for ``Mind'' Exploration of Large Language Model Society},
  journal = {arXiv preprint arXiv:2303.17760},
  year    = {2023},
  url     = {https://arxiv.org/abs/2303.17760}
}

@misc{lyu2026botcompany,
  author        = {Wenhan Lyu and Yue Xiao and Yixuan Zhang and Yifan Sun},
  title         = {Self-Organizing Multi-Agent Systems for Continuous Software Development},
  year          = {2026},
  eprint        = {2603.25928v1},
  archivePrefix = {arXiv},
  url           = {https://arxiv.org/abs/2603.25928v1}
}

@book{nelson1982evolutionary,
  author    = {Richard R. Nelson and Sidney G. Winter},
  title     = {An Evolutionary Theory of Economic Change},
  publisher = {Harvard University Press},
  address   = {Cambridge, MA},
  year      = {1982}
}

@article{qian2023chatdev,
  author  = {Chen Qian and Wei Liu and Hongzhang Liu and Nuo Chen and Yufan Dang and Jiahao Li and Cheng Yang and Weize Chen and Yusheng Su and Xin Cong and Juyuan Xu and Dahai Li and Zhiyuan Liu and Maosong Sun},
  title   = {{ChatDev}: Communicative Agents for Software Development},
  journal = {arXiv preprint arXiv:2307.07924},
  year    = {2023},
  url     = {https://arxiv.org/abs/2307.07924}
}

@misc{rajasekaran2026harnessdesign,
  author       = {Prithvi Rajasekaran},
  title        = {Harness Design for Long-Running Application Development},
  howpublished = {Anthropic Engineering},
  year         = {2026},
  month        = mar,
  url          = {https://www.anthropic.com/engineering/harness-design-long-running-apps},
  note         = {Published March 24, 2026}
}

@article{schmidt1992taking,
  author  = {Kjeld Schmidt and Liam Bannon},
  title   = {Taking {CSCW} Seriously: Supporting Articulation Work},
  journal = {Computer Supported Cooperative Work},
  volume  = {1},
  number  = {1--2},
  pages   = {7--40},
  year    = {1992},
  doi     = {10.1007/BF00752449}
}

@inproceedings{shinn2023reflexion,
  author    = {Noah Shinn and Federico Cassano and Ashwin Gopinath and Karthik Narasimhan and Shunyu Yao},
  title     = {{Reflexion}: Language Agents with Verbal Reinforcement Learning},
  booktitle = {Advances in Neural Information Processing Systems},
  volume    = {36},
  year      = {2023},
  url       = {https://papers.neurips.cc/paper_files/paper/2023/hash/1b44b878bb782e6954cd888628510e90-Abstract-Conference.html}
}

@article{teece1997dynamic,
  author  = {David J. Teece and Gary Pisano and Amy Shuen},
  title   = {Dynamic Capabilities and Strategic Management},
  journal = {Strategic Management Journal},
  volume  = {18},
  number  = {7},
  pages   = {509--533},
  year    = {1997},
  doi     = {10.1002/(SICI)1097-0266(199708)18:7<509::AID-SMJ882>3.0.CO;2-Z}
}

@misc{wang2023voyager,
  author        = {Guanzhi Wang and Yuqi Xie and Yunfan Jiang and Ajay Mandlekar and Chaowei Xiao and Yuke Zhu and Linxi Fan and Anima Anandkumar},
  title         = {{Voyager}: An Open-Ended Embodied Agent with Large Language Models},
  year          = {2023},
  eprint        = {2305.16291},
  archivePrefix = {arXiv},
  url           = {https://arxiv.org/abs/2305.16291}
}

@misc{wu2023autogen,
  author        = {Qingyun Wu and Gagan Bansal and Jieyu Zhang and Yiran Wu and Beibin Li and Erkang Zhu and Li Jiang and Xiaoyun Zhang and Shaokun Zhang and Jiale Liu and Ahmed Hassan Awadallah and Ryen W. White and Doug Burger and Chi Wang},
  title         = {{AutoGen}: Enabling Next-Gen {LLM} Applications via Multi-Agent Conversation},
  year          = {2023},
  eprint        = {2308.08155},
  archivePrefix = {arXiv},
  url           = {https://arxiv.org/abs/2308.08155}
}

@misc{young2025longrunningagents,
  author       = {Justin Young},
  title        = {Effective Harnesses for Long-Running Agents},
  howpublished = {Anthropic Engineering},
  year         = {2025},
  month        = nov,
  url          = {https://www.anthropic.com/engineering/effective-harnesses-for-long-running-agents},
  note         = {Published November 26, 2025}
}

@misc{yu2026onemancompany,
  author        = {Zhengxu Yu and Yu Fu and Zhiyuan He and Yuxuan Huang and Ka Yiu Lee and Meng Fang and Weilin Luo and Jun Wang},
  title         = {From Skills to Talent: Organising Heterogeneous Agents as a Real-World Company},
  year          = {2026},
  eprint        = {2604.22446},
  archivePrefix = {arXiv},
  url           = {https://arxiv.org/abs/2604.22446}
}

@misc{zhang2025gmemory,
  author        = {Guibin Zhang and Muxin Fu and Guancheng Wan and Miao Yu and Kun Wang and Shuicheng Yan},
  title         = {{G-Memory}: Tracing Hierarchical Memory for Multi-Agent Systems},
  year          = {2025},
  eprint        = {2506.07398},
  archivePrefix = {arXiv},
  url           = {https://arxiv.org/abs/2506.07398}
}

@misc{zhao2023expel,
  author        = {Andrew Zhao and Daniel Huang and Quentin Xu and Matthieu Lin and Yong-Jin Liu and Gao Huang},
  title         = {{ExpeL}: {LLM} Agents Are Experiential Learners},
  year          = {2023},
  eprint        = {2308.10144},
  archivePrefix = {arXiv},
  url           = {https://arxiv.org/abs/2308.10144}
}

@article{zhu2025multiagentbench,
  author  = {Kunlun Zhu and Hongyi Du and Zhaochen Hong and Xiaocheng Yang and Shuyi Guo and Zhe Wang and Zhenhailong Wang and Cheng Qian and Xiangru Tang and Heng Ji and Jiaxuan You},
  title   = {{MultiAgentBench}: Evaluating the Collaboration and Competition of {LLM} Agents},
  journal = {arXiv preprint arXiv:2503.01935},
  year    = {2025},
  url     = {https://arxiv.org/abs/2503.01935}
}

@misc{yang2026programbench,
  author        = {John Yang and Kilian Lieret and Jeffrey Ma and Parth Thakkar and Dmitrii Pedchenko and Sten Sootla and Emily McMilin and Pengcheng Yin and Rui Hou and Gabriel Synnaeve and Diyi Yang and Ofir Press},
  title         = {{ProgramBench}: Can Language Models Rebuild Programs From Scratch?},
  year          = {2026},
  eprint        = {2605.03546},
  archivePrefix = {arXiv},
  primaryClass  = {cs.SE},
  url           = {https://arxiv.org/abs/2605.03546}
}

@inproceedings{yang2024sweagent,
  author    = {John Yang and Carlos E. Jimenez and Alexander Wettig and Kilian Lieret and Shunyu Yao and Karthik R. Narasimhan and Ofir Press},
  title     = {{SWE-agent}: Agent-Computer Interfaces Enable Automated Software Engineering},
  booktitle = {The Thirty-eighth Annual Conference on Neural Information Processing Systems},
  year      = {2024},
  url       = {https://arxiv.org/abs/2405.15793}
}
\bibliographystyle{iclr2027_conference}

\appendix

\clearpage
\startcontents[appendices]
\pdfbookmark[1]{Appendix Contents}{relic.appendix.contents}
\section*{Appendix Contents}
\begingroup
\setcounter{tocdepth}{2}
\setlength{\columnsep}{1.25em}
\begin{multicols}{2}
\footnotesize
\printcontents[appendices]{}{1}{}
\end{multicols}
\endgroup
\clearpage

\raggedbottom
\setlength{\LTpre}{6pt plus 2pt minus 2pt}
\setlength{\LTpost}{4pt plus 1pt minus 1pt}
\setlength{\LTleft}{\fill}
\setlength{\LTright}{\fill}
\makeatletter
\newcommand{\RelicCurrentTable}{}
\newcommand{\RelicTableNotes}[1]{%
  \par\begingroup\footnotesize
  \leftskip=0pt\rightskip=0pt\parfillskip=0pt plus 1fil
  \noindent\textit{Notes.} #1\par\endgroup
}
\newcommand{\RelicMeasureTableStart}{%
  \ifdefempty{\RelicCurrentTable}{}{%
    \begingroup
    \dimen@=\ht\ifvoid\LT@firsthead\LT@head\else\LT@firsthead\fi
    \advance\dimen@ by \dp\ifvoid\LT@firsthead\LT@head\else\LT@firsthead\fi
    \advance\dimen@ by \ht\LT@foot
    \advance\dimen@ by \LTpre
    \vfuzz=\maxdimen\vbadness=10000
    \setbox\tw@=\copy\z@
    \setbox\tw@=\vsplit\tw@ to \ht\@arstrutbox
    \setbox\tw@=\vbox{\unvbox\tw@}
    \advance\dimen@ by \ht\ifdim\ht\@arstrutbox>\ht\tw@\@arstrutbox\else\tw@\fi
    \advance\dimen@ by \dp\ifdim\dp\@arstrutbox>\dp\tw@\@arstrutbox\else\tw@\fi
    \protected@write\@auxout{}{%
      \string\expandafter\string\gdef\string\csname\space
      relic@start@\RelicCurrentTable\string\endcsname{\the\dimen@}}
    \endgroup
  }
}
\pretocmd{\LT@start}{\RelicMeasureTableStart}{}{}
\newcommand{\RelicNeedTable}[2]{%
  \par\begingroup
  \@ifundefined{relic@start@#1}{\dimen@=80pt}{%
    \dimen@=\csname relic@start@#1\endcsname\relax}
  \advance\dimen@ by #2\baselineskip
  \Needspace{\dimen@}\endgroup
}
\newcommand{\RelicHeadingSpace}[1]{%
  \if@nobreak\else\Needspace{#1}\fi
}
\makeatother
\renewcommand{\relictablepanel}[2]{%
  \rowcolor{PanelSubFill}%
  \multicolumn{#1}{@{}l@{}}{\textbf{#2}}\\*
}
\AddToHook{cmd/section/before}{\RelicHeadingSpace{5\baselineskip}}
\AddToHook{cmd/subsection/before}{\RelicHeadingSpace{4\baselineskip}}
\AddToHook{cmd/subsubsection/before}{\RelicHeadingSpace{3\baselineskip}}

\renewcommand{\code}[1]{\RelicCode{#1}}

\section{Broader Discussion: Governance, Rule Quality, and Organizational Trade-offs}
\label{app:broader_discussion}

\subsection{Evidence Map: What Each Experiment Establishes}
\label{app:evidence_map}
\begin{table}[H]
\centering
\scriptsize
\setlength{\tabcolsep}{3.0pt}
\renewcommand{\arraystretch}{1.17}
\begin{tabularx}{\linewidth}{@{}P{0.255\linewidth}P{0.275\linewidth}L@{}}
\toprule
\rowcolor{PanelFill}
\textbf{Claim} & \textbf{Evidence line} & \textbf{Key result} \\
\midrule
Autonomous organizational state forms and persists & Terra+Opus 60-run B3 formation census & 497 autonomous proposals; 405 ever adopted; 280 formed lineages, including 32 with strong object/evaluator-linked evidence. \\
Institutionalization improves verified delivery & B3--B2, 360 controlled runs & Complete +5.71 pp [2.97, 8.55]; held-out +7.72 pp [2.67, 17.63]; exposed +6.58 [3.41, 10.14]; confirmed +6.96 [4.15, 10.83]. \\
Executable binding matters during online formation & B3 vs. B3-text, 30 matched runs & +7.18 pp complete contracts [3.54, 10.91] while proposal, governance, adoption, revision, and readable rules remain active. \\
Executable binding adds value to fixed readable knowledge & Exec vs. Text, 30 matched targets & Behavioral correctness +6.5 pp [0.7, 15.6] with identical frozen readable rule content. \\
Organizational state remains useful to fresh members & Exec vs. Fresh & Behavioral correctness +15.8 pp [9.2, 31.8] after member-local state is reinitialized. \\
Verified gains persist at matched realized token spend & B3 checkpoint under each matched B2 token cap & Complete +7.45 pp; held-out +5.08 pp; exposed +11.72 pp; confirmed issues +6.21 pp; all four 95\% intervals remain above zero. \\
Peer coordination improves externally & CooperBench full benchmark + 47-pair same-model control & Full benchmark: Relic 371/469 (79.1\%) vs. Peer 277/469 (59.1\%) and Team no-proto 349/469 (74.4\%) after excluding 183 broken benchmark pairs; same-model control: Official Peer 13/47, Solo 26/47, Relic 28/47. \\
Protocol behavior transfers to another harness & ProgramBench, fixed 25 tasks & Mean behavioral pass 64.164\% $\rightarrow$ 70.916\% (+6.752 pp; +10.5\% relative). \\
\bottomrule
\end{tabularx}
\caption{\textbf{Experimental results across formation, binding, transfer, cost, and external benchmarks.}}
\label{tab:evidence_map}
\end{table}

\subsection{Organizational Capability as a Systems Layer}
Relic separates model or member capability from the persistent state that governs how
multiple members work together. A stronger member can improve local reasoning,
implementation, or tool use, while an organization additionally determines how evidence
is routed, responsibility is assigned, work is reviewed, and local progress becomes
shared delivery. The capability census gives this distinction empirical content: formed
lineages are strongly concentrated in review/merge, release engineering, and evidence
governance. In this software-production setting, the observed institutional responses
concentrate near the boundary between local work and trustworthy, integrated
shared delivery. Persistent organizational state therefore acts as an additional
systems layer through which recurring coordination problems are represented and
acted on.

\subsection{Autonomous Formation or Fixed Protocol Packages?}
B3 develops its operating rules through proposal, governance, execution, revision, and retirement. Transfer Exec deploys a source-derived frozen rule package to fresh members. These configurations support two deployment modes: adapting organizational rules during work and reusing an established rule package at initialization.

\subsection{The Organizational Maturity Gap}
Current agent systems exhibit an asymmetry between rapidly improving individual
capability and comparatively thin organizational structure. Agents can already write
code, use tools, search, plan, and execute long tasks, yet collective work is often
coordinated through temporary conversations, manually assigned roles, shared
scratchpads, or procedures reconstructed anew in each run. We call this gap an \emph{organizational stone age}: individual agent capability is advancing faster than the persistent structures that organize collective work.

Natural language alone does not close this gap. Agents already possess a language for
communicating content; what remains underdeveloped is a durable \emph{language of work}
that specifies responsibility, evidence, review, integration, exceptions, and revision
in forms that continue to govern future members. Relic studies executable protocols as
one such representation. Other organizational objects---routing policies, role
structures, escalation paths, shared abstractions, or resource-allocation rules---may
support different forms of persistent collective capability.

\subsection{Governance Trades Cost for Reliability}
Governance routes responsibility, review, and verification through shared procedures. B3 incurs additional tokens per run, while its matched-spend checkpoints retain gains on all four verified production endpoints (Appendix~\ref{app:budget_capped_sensitivity}).

\subsection{Strong Execution Raises the Stakes of Rule Quality}
Rule triggers, evidence requirements, responsibilities, and revision paths determine how organizational experience governs later work. Relic makes these components explicit, supporting rule inspection, amendment, retirement, and reuse across member turnover.

\subsection{From Rule Formation to Evidence-based Meta-governance}
As organizations grow, forming rules is only the first governance problem. A larger
system must also decide who may propose, approve, override, revise, or retire rules;
whether a rule applies to one team or the whole organization; how conflicting rules are
resolved; and when exceptions are legitimate. Persistent agent organizations therefore
eventually require governance of governance.

A natural extension is evidence-based rule deployment. A proposed mechanism could begin
in advisory or shadow mode, run within a limited scope, be evaluated against explicit
outcomes, expand only after successful trials, and carry revision or sunset conditions.
Many familiar human governance ideas--independent review, staged adoption, appeals,
exceptions, delegated authority, and sunset clauses--provide useful design hypotheses.
Agent organizations also make forms of governance experimentation unusually tractable:
organizational state can be logged exactly, policies can be versioned, and in suitable
settings a rule can be snapshotted, rolled back, or compared under controlled replay.

\section{Research Agenda for Persistent AI Organizations}
\label{app:research_agenda}
These results motivate a research program on persistent AI organizations, with organizational capability as a common unit of analysis.

\subsection{Structured Decision Layers}
Future work can compare fixed and learned decision layers, alternative action-space abstractions, and responsibility-, risk-, and cost-aware routing across agent harnesses, domains, and model strengths.

\subsection{Organizations as a First-class Evaluation Unit}
Persistent AI systems can be evaluated at the level of the organization rather than only
the individual model or temporary team. Future benchmarks could vary topology, hierarchy,
authority allocation, role specialization, membership turnover, and organizational
memory while holding the underlying work environment fixed. Such studies could reveal
organization-level failure modes that task-bounded agent benchmarks cannot express.

\subsection{Capability and Capacity Taxonomies}
A broader evaluation framework should distinguish where capability resides. Model
capability concerns what the underlying model can do; agent capability additionally
includes tools, memory, and an execution loop; collaborative capability can arise within
a temporary team; organizational capability is retained in shared structure that can
affect future work independently of the private histories of the members who created it.
These layers can be characterized by formation, scope, executability, persistence,
transferability, compositionality, and reversibility. A complementary capacity taxonomy
can ask how much work, coordination load, member turnover, and rule complexity each layer
can sustain before performance degrades. Such distinctions would make it clearer whether
a system improves because of a stronger model, a better agent scaffold, a transient team
strategy, or a persistent organizational capability.

\subsection{Transfer and Selective Forgetting of Organizational State}
Organizational-transfer research can examine partial and gradual
turnover, cross-model and cross-domain transfer, interference among transferred
mechanisms, and the relative value of transferring protocols, workflows, roles,
responsibility maps, or other shared state. Equally important is learning when
organizational state should be adapted or deliberately forgotten rather than preserved.

\subsection{Capability Ecology and Failures of Organizational Learning}
Future work can study how capability families compose, compete, become obsolete, or
depend on one another; how these distributions change with model strength and
environment; and where recurring friction fails to become institutionalized. A formation
funnel from friction exposure through recognition, proposal, adoption, formation, and
strong effect evidence could make failures of organizational learning as measurable as
successful capability formation.

\subsection{Human--Agent Organization Interaction}
As agent organizations become more persistent and internally complex, the human
interaction target may shift from an individual agent to an organization. Future studies
can compare direct human participation, secretary-assisted interaction, and a secretary
acting as an organizational interface that summarizes state, surfaces pending decisions,
and translates human intent into organizational actions. This raises questions about
abstraction, explainability, delegation, approval and veto rights, responsibility
attribution, and accountability in multi-human--multi-agent organizations.

\subsection{Toward Agentic Sociology}
Persistent organizations also create a scale of analysis beyond one team. Multiple agent
organizations may specialize, exchange work, depend on shared infrastructure, negotiate
interfaces, and develop rules for interaction across organizational boundaries. We use \emph{agentic sociology} as a research lens for studying persistent roles,
organizations, norms, institutions, and relations around populations of artificial
agents, filling the level between isolated multi-agent interaction and broader
multi-organization systems.

At this scale, questions of interoperability and governance separate. A communication
protocol may let organizations exchange messages without establishing who owns a task,
what evidence satisfies an obligation, which authority may revise a commitment, or how a
dispute is resolved. Studying those structures requires benchmarks that span persistent
organizations, turnover, cross-organizational dependencies, and human oversight rather
than only task-bounded conversations.

Together, these directions suggest a research agenda on how AI organizations make
decisions, accumulate and transfer capabilities, govern their own rules, adapt across
environments, and remain legible and accountable to humans.

\section{Extended Method and Formal Specification}
\label{app:extended_method}

This section specifies organizational state, action selection, governance, execution, and capability formation.

\subsection{State and Member Information}
\label{app:state_information}
\label{app:complete_history}

Let $H_t$ denote the complete recorded history of a run. For member $i$, let
$I^{\mathrm{acc}}_{i,t}\subseteq H_t$ be the information the member is allowed
to access, and let $v_{i,t}\subseteq I^{\mathrm{acc}}_{i,t}$ be the bounded
context actually read or retrieved at decision time. The runtime maintains
\[
    v_{i,t}\subseteq I^{\mathrm{acc}}_{i,t}\subseteq H_t.
\]
Private branches, unread messages, private workspaces, and evaluator-only assets
therefore need not appear in a member's context even though they exist in the
recorded run.

\subsubsection{Member-Local State}
\label{app:member_local_state}

Member-local state contains the information and learned quantities carried by
one member: functional role, decision prior, skills, current work, private
workspace, private memory, local commitments, and bounded workload state.
These quantities can affect that member's later decisions without becoming
organization-owned state.

\begin{table}[H]
\centering
\small
\begin{tabularx}{\linewidth}{@{}P{0.25\linewidth}L@{}}
\toprule
\rowcolor{PanelFill}
\multicolumn{2}{@{}l}{\textbf{Member-local state used in the experiments}}\\
\textbf{Category} & \textbf{Role in the system}\\
\midrule
Identity and function & Role, specialty, and stable functional decision prior.\\
Skill and standing & Skill, reputation, and authority variables used by the structured selector where enabled.\\
Current work & Active tasks, work status, and local scheduling state.\\
Private workspace & Member-owned notes, files, drafts, branches, and sandbox outputs.\\
Private memory & Salient previously observed events retained for later context.\\
Commitments & Member-specific promises, requests, and unresolved obligations.\\
\bottomrule
\end{tabularx}
\caption{Member-local state.}
\label{tab:member_local_state}
\end{table}

\subsubsection{Organizational State}
\label{app:organizational_state_schema}

We write organizational state as
\[
    O_t=(W_t,Q_t,R_t),
\]
where $W_t$ contains shared work objects, $Q_t$ contains responsibility and
decision-right mappings, and $R_t$ contains adopted organizational protocols.
The defining distinction is operational: a shared document belongs to $W_t$;
a rule belongs to $R_t$ when it has an active runtime binding that can affect
later work.

\begin{table}[H]
\centering
\small
\begin{tabularx}{\linewidth}{@{}P{0.12\linewidth}P{0.25\linewidth}L@{}}
\toprule
\rowcolor{PanelFill}
\multicolumn{3}{@{}l}{\textbf{Organizational state}}\\
\textbf{Part} & \textbf{Contents} & \textbf{Operational role}\\
\midrule
$W_t$ & Tasks, issues, shared documents, repository artifacts, boards, messages & Persistent shared work and evidence.\\
$Q_t$ & Task ownership, reviewer/approver responsibility, authority relations & Routes work and assigns organizational responsibility.\\
$R_t$ & Adopted protocol specifications, lifecycle state, use/violation/enforcement records & Stores learned ways of working that can govern later members.\\
\bottomrule
\end{tabularx}
\caption{Organizational state.}
\label{tab:org_state_schema}
\end{table}

\subsubsection{Execution Records}
\label{app:audit_provenance}

Evaluator-side provenance records actions, work-object transitions, governance
events, and decision traces for measurement and replay. These records preserve
the sequence connecting proposal, adoption, later use, and execution consequences.
Member decisions continue to use the bounded perception and retrieval interface
in Appendix~\ref{app:state_information}; the complete recorded history supports
researcher-side reconstruction of the same trajectory.

\subsection{Persistent Organizations and Organizational Learning}
\label{app:organization_definition}
\label{app:operational_organization}
\label{app:organization_vs_individual}

A persistent organization is a multi-member system in which learned rules and
cross-member responsibilities are stored outside member-local histories and
can remain operative when members are replaced. We represent the organization
at time $t$ as
\[
  \mathcal{A}_t = \bigl(N_t,\{m_{i,t}\}_{i\in N_t},O_t,(\mathcal{C},D_\theta,E),\Gamma\bigr),
\]
where $N_t$ is the roster, $m_{i,t}$ is member-local state, $O_t$ is
organizational state, $(\mathcal{C},D_\theta,E)$ is the fixed candidate,
decision, and execution substrate, and $\Gamma$ is the governance process that
introduces or revises persistent mechanisms.

\subsubsection{Member Learning and Organizational Learning}
\label{app:member_vs_org_learning}

Let $I_t=(Q_t,R_t)$ denote the institutional component of $O_t$, containing
responsibility and decision-right mappings and adopted protocols. The experiment
distinguishes three update paths:
\begin{equation}
\begin{aligned}
 W_{t+1} &= F_{\mathrm{work}}(W_t,a_t),\\
 I_{t+1} &= F_{\mathrm{institution}}(I_t,W_t,a_t,g_t),\\
 m_{i,t+1} &= F_{\mathrm{agent}}(m_{i,t},o_{i,t},a_t,r_t).
\end{aligned}
\label{eq:state-update-paths}
\end{equation}
Ordinary work changes shared artifacts; member learning changes a member's
local state; organizational learning introduces or revises governed mechanisms
whose effects can extend to members who did not create them.

\subsubsection{Persistence Across Member Replacement}
\label{app:organizational_persistence}
\label{app:fresh_member_semantics}

Member replacement reinitializes private memory, private workspace and sandbox
state, message read/acknowledgement state, commitments, reflections, wishes,
and accumulated member-local experience. The functional role slots are
reinstantiated, and shared work can persist across the replacement. In the
reported transfer experiment, a freshly initialized target receives the
specified protocol package and responsibility mappings. The target
initialization and transferred contents are specified in
Appendix~\ref{app:fresh_roster}.

\subsection{Structured Decision Layer and Execution}
\label{app:decision_execution_details}
\label{app:candidate_construction}

At each decision, the system constructs a feasible candidate set
$\mathcal{C}_{i,t}$ from member-visible work state, role constraints, target
validity, and the common action surface. Candidate construction is shared by
the compared conditions; the action-selection mechanism differs by condition.

\subsubsection{Structured Decision Layer (SDL)}
\label{app:sdl_details}

B2 and B3 score candidates with the same 32-dimensional, rule-computed feature
representation. The feature groups are shown in Table~\ref{tab:sdl_feature_groups}.

\begin{table}[H]
\centering
\small
\begin{tabularx}{0.72\linewidth}{@{}L r@{}}
\toprule
\rowcolor{PanelFill}
\multicolumn{2}{@{}l}{\textbf{SDL candidate-feature dimensions}}\\
\textbf{Feature group} & \textbf{Dimensions}\\
\midrule
Task progress and priority & 7\\
Skill and role fit & 4\\
Social coordination and trust & 8\\
Repository health and evidence quality & 6\\
Protocols and institutional memory & 5\\
Governance-proposal signals & 2\\
\midrule
\textbf{Total} & \textbf{32}\\
\bottomrule
\end{tabularx}
\caption{SDL feature groups used by B2 and B3.}
\label{tab:sdl_feature_groups}
\end{table}

For candidate $a$,
\[
 w_k(i,t)=b_k+\sum_j p_{ij}(t)\beta_{jk},\qquad
 s_{\mathrm{base}}(a)=\sum_{k=1}^{32}w_k(i,t)\phi_k(a,x_{i,t}),
\]
and
\[
 u(a)=s_{\mathrm{base}}(a)+A(a)+R(a)+C(a)+\sum_gG_g(a).
\]
The stochastic branch samples from
\[
 P(a)=\frac{\exp((u(a)+\epsilon_a)/0.6)}
 {\sum_{a'}\exp((u(a')+\epsilon_{a'})/0.6)},
 \qquad \epsilon_a\sim\mathrm{Uniform}(-0.05,0.05),
\]
using namespaced seeded randomness. The scoring parameterization is fixed
before the reported runs and shared by B2 and B3.

\begin{table}[H]
\centering
\scriptsize
\begin{tabularx}{\linewidth}{@{}P{0.23\linewidth}L r@{}}
\toprule
\rowcolor{PanelFill}
\multicolumn{3}{@{}l}{\textbf{SDL base-weight coefficients}}\\
\textbf{Group} & \textbf{Feature} & \textbf{Base weight}\\
\midrule
Task / progress & progress gain; deadline urgency; task priority; blocker resolution; dependency unlock; demo relevance; customer relevance & $+0.55,+0.30,+0.30,+0.35,+0.20,+0.30,+0.30$\\
Skill fit & skill match; role affinity; learning gain; low-skill failure risk & $+0.40,+0.30,+0.10,-0.30$\\
Social / coordination & coordination; clarity; trust gain/risk; conflict risk; visibility; reputation gain/risk & $+0.30,+0.20,+0.25,-0.25,-0.25,+0.20,+0.20,-0.20$\\
Repository / evidence & repository health; technical debt; review quality; reproducibility; untracked-result risk; claim evidence & $+0.30,-0.30,+0.25,+0.25,-0.20,+0.25$\\
Protocol & creation potential; use potential; violation risk; enforcement gain; institutional memory & $+0.20,+0.20,-0.40,+0.20,+0.20$\\
Governance & proposal endorsement; proposal skepticism & $+0.45,+0.45$\\
\bottomrule
\end{tabularx}
\caption{Complete base-weight vector, grouped for readability.}
\label{tab:sdl_base_weights}
\end{table}

\begin{table}[H]
\centering
\small
\begin{tabularx}{0.84\linewidth}{@{}L L@{}}
\toprule
\rowcolor{PanelFill}
\multicolumn{2}{@{}l}{\textbf{SDL post-score terms}}\\
Softmax temperature & $0.6$\\
Seeded jitter & $\mathrm{Uniform}(-0.05,+0.05)$\\
Authority bonus & $0.07\times$ domain authority\\
Reputation bonus & clipped at $0.10$\\
Coding affinity & bounded role--action bonus\\
\bottomrule
\end{tabularx}
\caption{Post-score terms used by the SDL.}
\label{tab:sdl_postscore_terms}
\end{table}

\begin{table}[H]
\centering
\small
\begin{tabularx}{0.86\linewidth}{@{}L r@{}}
\toprule
\rowcolor{PanelFill}
\multicolumn{2}{@{}l}{\textbf{Principal soft-guard penalties}}\\
Repetition & $-0.18$\\
Artifact churn & $-0.12$\\
Unresolved review & $-0.15$\\
Missing new information & $-0.20$\\
Opportunity cost & $-0.12$\\
Shipping/off-task & $-0.50$\\
Unshipped release idle & $-1.20$\\
\bottomrule
\end{tabularx}
\caption{Principal soft-guard terms outside the core dot product.}
\label{tab:sdl_guard_terms}
\end{table}

\begin{relicalgorithm}[H]
\caption{SDL-based collective decision (one member, one step)}
\label{alg:a1}
\begin{minipage}{0.96\linewidth}
\small
Construct visible feasible candidates $\mathcal{C}_{i,t}$. Apply shared
feasibility, cooldown, and pool-size guards. For B0/B1, the model chooses from
the candidate set. For B2/B3, extract the 32 features, compute $u(a)$ for each
candidate, and select by deterministic maximum or seeded softmax. Record the
choice trace for evaluation, then execute the selected action through the
common execution substrate.
\end{minipage}
\end{relicalgorithm}

\subsubsection{Execution Operator}
\label{app:execution_operator}
\label{app:fixed_architecture}

The common execution layer validates the selected action, applies its
work-state transition, accounts for costs, and records the resulting events.
In standard B3, adopted protocols supply decision features and responsibility
signals to the selector, followed by protocol-linked checks and records after
the action handler. SDL utilities determine action priorities and selection
probabilities; transition validation and execution checks determine the
resulting work-object disposition. Shared repository requirements, including
review and current-CI evidence, are enforced by the common environment.
The action registry and execution handlers remain fixed while adopted
protocol specifications and role mappings evolve.

\subsection{Capability Formation and Governance}
\label{app:capability_governance}
\label{app:friction_evidence}

Relic converts visible recurring work friction into proposals. A proposal names
the problem, supporting evidence, affected work, required actions, and the
intended organizational response. Validation checks grounding, supported
actions, scope, duplication, and governance requirements before a proposal can
be adopted.

\subsubsection{Proposal and Protocol Representation}
\label{app:proposal_validation}

\begin{table}[H]
\centering
\small
\begin{tabularx}{\linewidth}{@{}P{0.22\linewidth}L@{}}
\toprule
\rowcolor{PanelFill}
\multicolumn{2}{@{}l}{\textbf{Proposal and executable-protocol representation}}\\
\textbf{Group} & \textbf{Fields}\\
\midrule
Identity & proposal type, title, proposer\\
Evidence & source observations/wishes, target problem, proposed response\\
Feasibility & required actions and capabilities\\
Governance & review state, required approvals, support/opposition, adoption score\\
Lifecycle & amendment target, revision state, retirement state\\
Protocol content & trigger, scope, responsible roles, required steps/evidence, affected actions/artifacts, violation/exception rules, enforcement response, success metric\\
Accounting & use, violation, enforcement, revision, and last-use records\\
\bottomrule
\end{tabularx}
\caption{Proposal and adopted-protocol fields.}
\label{tab:proposal_schema}
\end{table}

\subsubsection{Governance, Adoption, and Runtime Binding}
\label{app:governance_compilation}

The reported main-study configuration uses semi-automatic governance: members
provide approvals first, while eligible proposals that remain pending for at
least 24 steps can receive bounded deadlock recovery. Adoption additionally
requires at least two supporters, at least two distinct approvers, a minimum
review interval of three steps, and an adoption score of at least $0.6$.

\begin{table}[H]
\centering
\small
\begin{tabularx}{0.88\linewidth}{@{}P{0.18\linewidth}L@{}}
\toprule
\rowcolor{PanelFill}
\multicolumn{2}{@{}l}{\textbf{Approval-mode semantics}}\\
Auto & Eligible proposals can receive required approvals automatically.\\
Agent & Required approvals must come from member actions.\\
Semi-auto & Member approval is primary; bounded deadlock recovery can complete missing approvals after the waiting period.\\
\bottomrule
\end{tabularx}
\caption{Approval modes supported by the governance layer.}
\label{tab:approval_modes}
\end{table}

After adoption, a structured protocol specification is registered as
organization-owned state. Its trigger, responsibilities, evidence obligations,
and affected actions become available to the decision and execution layers.
Amendments follow the same governed path; retirement disables future use while
retaining lifecycle history.

\begin{relicalgorithm}[H]
\caption{Governed organizational-state update}
\label{alg:a2}
\begin{minipage}{0.96\linewidth}
\small
From visible repeated friction, create a proposal. Validate its grounding and
runtime compatibility. Route it for governance. When support, approval,
review-time, and adoption-score requirements are satisfied, compile the
proposal into a structured protocol and register it as organizational state.
Later matching actions can generate use, violation, or enforcement records.
Amendments repeat the same governance path; retired protocols cease governing
future work while their history remains auditable.
\end{minipage}
\end{relicalgorithm}

\subsubsection{Operational Capability-Formation Criterion}
\label{app:formation_criterion}
\label{app:lifecycle_diagnostics}
\label{app:three_claims}

A registered autonomous protocol is counted as formed when it has a valid
proposal and adoption record, at least two qualifying post-adoption uses,
those uses span at least two recorded organizational units and two distinct
steps, and the span from adoption to the last qualifying use is at least 48
steps. Strong evidence additionally links later third-party execution or
enforcement to a governed-object state change and independent evaluator
evidence. 

\begin{relicalgorithm}[H]
\caption{Offline capability detection (read-only)}
\label{alg:a3}
\begin{minipage}{0.96\linewidth}
\small
For each adopted autonomous rule, collect qualifying post-adoption uses. Mark
weak formation when the use-count, organizational-unit, distinct-time, and
48-step persistence conditions all hold. Mark strong evidence when weak
formation is additionally linked to later third-party execution or enforcement,
a governed-object state change, and independent evaluator evidence. The
detector reads the recorded ledger and does not modify the run.
\end{minipage}
\end{relicalgorithm}

\subsection{Core Experimental Invariants}
\label{app:algorithms_invariants}
\label{app:runtime_algorithm}
\label{app:update_algorithm}
\label{app:detector_algorithm}
\label{app:core_invariants}

Table~\ref{tab:invariants} summarizes the implementation invariants linking
member-visible execution, governed updates, controlled comparisons, and
measurement. Configuration and access checks are detailed in
Appendix~\ref{app:condition_conformance}.

\begin{table}[H]
\centering
\small
\begin{tabularx}{\linewidth}{@{}P{0.34\linewidth}L@{}}
\toprule
\rowcolor{PanelFill}
\multicolumn{2}{@{}l}{\textbf{Experimental invariants}}\\
Member information interface & Member context is built from permitted information that has been read or rendered; evaluator-private assets are held in the evaluator interface.\\
Cross-member privacy & Sharing explicitly controls the visibility of private branches, workspaces, messages, and sandbox outputs to other members.\\
Governed institutional updates & Protocol adoption follows proposal validation and the configured review and approval path.\\
Registered action surface & State mutations occur through registered actions and validated transitions.\\
Matched B2/B3 substrate & B2 and B3 share the structured selector, candidate construction, member-learning path, task information, and execution substrate; B3 enables institutionalization.\\
Fresh-member reset & Transfer reinitializes member-local state and applies the designated organizational package and role mappings.\\
Read-only measurement & Formation detection reads recorded ledgers; evaluators score exported candidate state outside the organizational rollout.\\
\bottomrule
\end{tabularx}
\caption{Core invariants enforced by the experimental design.}
\label{tab:invariants}
\end{table}

\section{Execution-Grounded Organizational Environment}
\label{app:environment}

This section summarizes the common software-production environment used by all
main-study conditions. The environment provides persistent shared work,
partial observability, executable work transitions, and a frozen evaluator.
Condition-specific organizational learning is layered on top of this common
substrate.

\subsection{Environment and Work State}
\label{app:environment_boundary}
\label{app:execution_grounded_work}
\label{app:current_environment_boundary}

Members act on persistent software-production objects: tasks and issues,
working-tree artifacts, branches, commits, pull requests, CI records, reviews,
shared documents, messages, and release state. Product progress is measured from concrete state transitions.

\subsubsection{Main-Study Components}
\label{app:environment_status}

\begin{table}[H]
\centering
\small
\begin{tabularx}{\linewidth}{@{}P{0.34\linewidth}P{0.20\linewidth}L@{}}
\toprule
\rowcolor{PanelFill}
\multicolumn{3}{@{}l}{\textbf{Common environment and condition-specific activation}}\\
\textbf{Component} & \textbf{Available} & \textbf{Main-study role}\\
\midrule
Repository, CI, review, merge, release & yes & Common execution-grounded work substrate.\\
Tasks, issues, shared board & yes & Common ownership and responsibility substrate.\\
Channels and direct messages & yes & Common information-friction substrate.\\
Private workspaces and sandboxes & yes & Common partial-observability substrate.\\
Structured decision layer & yes & Used for action selection in B2/B3.\\
Member learning & yes & Enabled in B2/B3.\\
Protocol proposal, governance, and execution & yes & Institutionalization enabled in B3.\\
Frozen evaluator & yes & Scores candidate mainline state outside member context.\\
Token accounting & yes & Measures realized model usage.\\
\bottomrule
\end{tabularx}
\caption{Environment components relevant to the reported main study.}
\label{tab:module_status}
\end{table}

\subsection{Shared Work, Visibility, and Communication}
\label{app:object_model}
\label{app:member_workspace_objects}
\label{app:software_objects}
\label{app:communication_objects}
\label{app:governance_objects}
\label{app:evaluator_objects}
\label{app:visibility_propagation}
\label{app:private_shared_transition}

The environment separates private, shared, and evaluator-only state. Local
edits remain private until committed and exposed through the repository
workflow; sandbox outputs remain private until explicitly shared; a shared
document can be visible before its full content is opened; and messages become
part of a member's context only after the member reads them. Evaluator-only
tests, held-out requirements, and references never enter member-visible state.

\subsubsection{Read and Retrieval Semantics}
\label{app:read_semantics}
\label{app:search_retrieval}

Visibility, delivery, reading, acknowledgement, and memory are distinct. Inbox
triage admits at most 25 unread perceivable messages per step. Search operates
over frozen internal corpora, repository state, member-local sandbox state,
and fixed external-snapshot corpora; no live network retrieval is used in the
reported environment.

\subsection{Action Surface and Execution Validation}
\label{app:action_registry}
\label{app:task_actions}
\label{app:repository_actions}
\label{app:communication_actions}
\label{app:governance_actions}

The common action surface covers task work, repository operations,
verification, communication, meetings, search, documents, artifacts,
governance, and release operations. Feasible actions are generated from the
member's visible state and role. Selection is followed by transition-specific
validation, so an offered action can still fail when its target or precondition
has changed.

\subsubsection{Invalid Actions and Failures}
\label{app:invalid_actions}

\begin{table}[H]
\centering
\small
\begin{tabularx}{\linewidth}{@{}P{0.28\linewidth}P{0.28\linewidth}L@{}}
\toprule
\rowcolor{PanelFill}
\multicolumn{3}{@{}l}{\textbf{Execution outcomes}}\\
\textbf{Failure class} & \textbf{Where resolved} & \textbf{Recorded effect}\\
\midrule
Invisible or invalid target & candidate/transition validation & No product mutation; decision is rejected.\\
Invalid work transition & action-specific validation & Failed execution with recorded reason.\\
Missing role or approval & action/governance validation & Transition refused.\\
Malformed model output & structured-output validation & No action mutation; model usage remains counted.\\
Unavailable action & candidate construction & Action is absent from the feasible set.\\
\bottomrule
\end{tabularx}
\caption{High-level invalid-action handling used in the experiments.}
\label{tab:invalid_actions}
\end{table}

\subsection{Software-Production Lifecycle and Verification}
\label{app:production_lifecycles}
\label{app:repair_extension_lifecycle}
\label{app:zero_to_one_lifecycle}
\label{app:workspace_mainline}
\label{app:ci_review_merge}
\label{app:verification_actions}

Repository-based workloads begin from a frozen initial repository and specify
repairs or feature extensions. Construction workloads begin from an
unimplemented starter skeleton plus a frozen specification and public checks.
In both cases, the delivery chain is edit $\rightarrow$ commit $\rightarrow$
pull request $\rightarrow$ review/CI $\rightarrow$ merge, and reported product
metrics are evaluated on mainline.

Verification is commit-bound. A CI result records the tested pull-request head
and the mainline state against which it was evaluated. Pushing a new commit
invalidates earlier evidence for that head, and mainline movement can make a
previous result stale before merge. These checks are common environment
properties shared by all conditions; Relic can learn organizational rules that
route responsibility and evidence around them.

\subsection{Time, Scheduling, and Resource Accounting}
\label{app:time_resources}
\label{app:tick_semantics}
\label{app:action_duration}
\label{app:token_accounting}
\label{app:stopping_resume}

The main-study horizon is 336 scheduling steps with checkpoints every 24
steps. A scheduling step denotes one simulation update; elapsed execution time
is recorded separately in seconds. Members take at most one primary action
per step, while background work can complete asynchronously. Runtime records
retain the field name \code{tick} for the scheduling index.

Realized model usage is accumulated from provider receipts, including retry
calls. Tool and container compute is unmetered and remains separate from the
provider-token totals. Checkpoint resume loads the saved organization and
continues under the same run identity, preserving the relation between the
trajectory, its scheduling position, and its accumulated resource use.

\subsection{Organizational Runtime and Protocol Activation}
\label{app:org_runtime}
\label{app:workflow_schema}
\label{app:rights_schema}
\label{app:rule_schema}
\label{app:rule_activation}
\label{app:rule_revision}

The shared work layer stores operational artifacts; responsibility mappings
identify owners, reviewers, and approvers; adopted protocols add triggers,
required evidence, affected actions, responsible functions, exception rules,
and lifecycle state. In standard B3, learned rules can reprioritize actions,
route responsibility, and participate in protocol-linked execution checks.
Amendments follow the same governance path as initial adoption, and retirement
stops future application while retaining the recorded history.

\subsection{Logging, Checkpointing, and Replay}
\label{app:logging_replay}
\label{app:event_log_schema}
\label{app:checkpoint_contents}
\label{app:replay_semantics}
\label{app:inspector_boundary}

Each executed action records its actor, action type, success or failure,
affected work objects, costs, and resulting events. Governance transitions
add proposal, adoption, amendment, and retirement records; protocol execution
adds the corresponding use and enforcement events. Checkpoints serialize
organizational state, member state, and scheduling state for run recovery.

Case-study timelines are reconstructed from these ledgers and checkpoints.
Researcher-facing inspection joins the complete recorded state and event
history; member-facing decisions use the bounded perception interface.
This separation preserves both the local information used to choose an action
and the shared-state consequences used to analyze it.

\section{Benchmark Curation}
\label{app:benchmark-curation}

This section documents the ten frozen software-production workloads used in the
360-run headline endpoint study: three models, three seeds, and four conditions per workload.
W01--W10 are stable identifiers in the frozen analysis \code{PACK\_ORDER}.
Table~\ref{tab:pack_inventory} gives their task names, version transitions,
and original pack IDs; the same mapping applies to all workload-level results,
transfer references, and the selected case. Construction and pre-run
qualification are specified below.

\paragraph{Development tasks.}
Relic was developed and debugged on repositories outside W01--W10 before
the reported evaluation. Development records and the 360-run evaluation
matrix have separate run identities. Within B3 evaluation runs, online
protocol formation and revision remain active components of the method.

\subsection{Evaluation Scope and Workload Taxonomy}
\label{app:workload_taxonomy}

\subsubsection{Frozen-Version Repair and Feature Development}
\label{app:repair_tasks}

W06--W10 use existing upstream projects with both an initial version and a
final reference version frozen during benchmark construction. The complete
initial repository is the starter; the final repository is the reference
implementation. The exact version pairs appear in
Table~\ref{tab:pack_inventory}. Feature changes and issue requirements between
these versions define the target behaviors, covering both bug fixes and new
functionality. Behavioral test cases and verification logic are written for
those requirements and executed against both versions. Every retained scored
case must fail on the untouched initial version and pass on the frozen final
version. The resulting requirements, cases, verifier, and version pair are
fixed before organizational evaluation.

Members receive the starter, designated public requirements, and public checks.
The final implementation, hidden acceptance tests, and held-out requirements
remain evaluator-side during rollout. A verified repair or feature completion
requires a retained baseline-failing check to pass on the candidate, using
\code{baseline\_status == failed and candidate\_status == passed}.
The frozen analysis retains the label \emph{repair} for the W06--W10 family;
this repository-based family includes both defect repair and feature development.

\subsubsection{Cross-Cutting Feature Extension}
\label{app:extension_tasks}

Cross-cutting requirements modify public interfaces and dependencies across
modules. The interface-repair affordance inserts an edit candidate for each
importer when an earlier edit removes a public symbol. The parameter-threading
affordance similarly supplies edit candidates when a signature gains a
parameter that is not forwarded. These candidates expose the dependent
implementation work through the common action surface.

The frozen manifests distinguish complete repository snapshots from
construction skeletons, while component maps identify the module boundaries
touched by each requirement. Cross-cutting extension demand is recorded
within the repository-based and construction workload families.

\subsubsection{Zero-to-One Construction}
\label{app:construction_tasks}

W01--W05 are zero-to-one construction workloads grounded in the functionality
of existing software. For each workload, the author manually selected an
existing project as a functional reference and specified the capabilities of
a new, analogous program. The authoring LLM received this functional brief
without the original project's source code or existing implementation. It
first wrote technical documentation and an explicit feature specification,
then implemented a new reference program, and subsequently wrote feature
test cases and a verifier from the documented features and that completed
implementation. 

An unimplemented starter skeleton supplies the new program's package structure
and signatures. Every retained scored case is checked against both this
starter and the completed reference, requiring failure on the former and
success on the latter. Requirements, starter, reference, cases, and verifier
are frozen before any evaluated organization begins work. The evaluated
members receive the specification, starter, and designated public checks;
they do not receive the original project's implementation, the newly authored
reference implementation, or evaluator-only test content. 

Contracts group related behavioral cases; a complete contract requires every
constituent case to pass. Mini Blobstore (W01) has five progressive contracts
with seven cases each. Its qualified starter passes 0/35 cases and 0/5
contracts; its reference passes 35/35 and 5/5. All ten workloads satisfy the
per-case qualification in Appendix~\ref{app:evaluator_bounds}.

Dependencies between contracts create recurring implementation, verification,
and integration demands. The workload supplies these dependencies without
prescribing a particular division of labor.

\RelicNeedTable{tab:pack_inventory}{5}
\subsection{Workload Sourcing and Eligibility}
\label{app:workload_sourcing}

\subsubsection{Frozen Sources and Specifications}
\label{app:source_materials}
\label{app:source_repos}

\begingroup
\centering
\begingroup
\footnotesize
\arrayrulecolor{RuleGray}
\setlength{\tabcolsep}{3pt}
\renewcommand{\arraystretch}{1.18}
\def\RelicCurrentTable{tab:pack_inventory}
\setlength{\LTcapwidth}{\linewidth}
\begin{xltabular}{\linewidth}{@{}P{0.055\linewidth} P{0.405\linewidth} L@{}}
\caption{\textbf{Frozen main-study workload mapping.}}\label{tab:pack_inventory}\\
\toprule
\relictabletitle{3}{Frozen main-study workload mapping}
\textbf{ID} & \textbf{Benchmark workload} & \textbf{Frozen pack ID} \\
\midrule
\endfirsthead
\multicolumn{3}{@{}l@{}}{\normalfont\footnotesize \textbf{Table~\thetable{} (continued)}}\\[3pt]
\toprule
\relictabletitle{3}{Frozen main-study workload mapping}
\textbf{ID} & \textbf{Benchmark workload} & \textbf{Frozen pack ID} \\
\midrule
\endhead
\midrule
\multicolumn{3}{@{}r@{}}{\normalfont\scriptsize Continued on next page.}\\
\endfoot
\bottomrule
\endlastfoot
\relictablepanel{3}{W01--W05: function-grounded zero-to-one construction}
W01 & Mini Blobstore & \code{mini\_blobstore\_v1} \\
W02 & Traffic Watch & \code{traffic\_watch\_v1} \\
W03 & TG Automation & \code{tg\_automation\_v1} \\
W04 & PDF Reformatter & \code{pdf\_reformatter\_v1} \\
W05 & FastAPI Dashboard & \code{fastapi\_dashboard\_v1} \\
\relictablepanel{3}{W06--W10: upstream version-transition tasks}
W06 & Boltons 24.0.0 $\rightarrow$ 26.1.0 & \code{boltons\_v2400\_to\_v2610} \\
W07 & Celery 5.6.0 $\rightarrow$ 5.6.3 & \code{celery\_v560\_to\_v563} \\
W08 & Soup Sieve 2.6 $\rightarrow$ 2.9.1 & \code{soupsieve\_v26\_to\_v291} \\*
W09 & cattrs 25.1.0 $\rightarrow$ 26.1.0 & \code{cattrs\_v2510\_to\_v2610} \\*
W10 & Tenacity 8.2.3 $\rightarrow$ 9.1.4 & \code{tenacity\_v823\_to\_v914} \\
\end{xltabular}

\endgroup
\arrayrulecolor{black}
\endgroup

W01--W10 identify the same workloads in task manifests, result tables,
transfer records, and case-study references. Each frozen manifest links the
starter, requirements, evaluator assets, reference, qualification result, and
freeze record. Original pack IDs and upstream version pairs are retained in
Table~\ref{tab:pack_inventory}, providing the join from paper-level workload
identifiers to the frozen source artifacts.

\subsubsection{Inclusion Criteria}
\label{app:inclusion_criteria}

A pack is usable only when three conditions hold and
\code{tools/preflight\_pack\_environment.py --score} reports all three:

\begin{enumerate}
\item \textbf{the environment works} -- every seeded issue names a file an agent
can reach and edit, and the merge gate can actually run and refuse;
\item \textbf{the verifier works} -- every retained scored case fails on the
untouched starter and passes on the fixed reference, verified case by case;
\item \textbf{the gate says something} -- the agent-visible acceptance gate is red
on the untouched starter, so it can distinguish work from no work.
\end{enumerate}

Additional structural requirements are enforced by
\code{tools/check\_pack\_health.py}: every issue must resolve to an editable
artifact; held-out requirements must be absent from the agent-visible stream; at
least \code{\_MIN\_PUBLIC\_TESTS = 3} public tests must exist; and no editable
file may exceed \code{\_MAX\_EDITABLE\_BYTES = 200{,}000}. Dependencies must be
pinned in-tree, the licence must permit redistribution, and no task may require a
live network. 

Each retained pack includes agent-visible acceptance checks that are red on the starter and green on the reference. The retained public-gate coverage is 12/12 for W06, 8/12 for W07, 6/9 for W08, 7/12 for W09, and 9/12 for W10.

\subsubsection{Exclusion Criteria}
\label{app:exclusion_criteria}

Packs are excluded for unstable evaluation, manual scoring, incompatible licensing, already-satisfied starter behavior, unavoidable future-solution leakage, irreproducible dependencies, or absent organizational demand.

\subsection{Freezing and Manifest Construction}
\label{app:freezing_manifests}

\subsubsection{Starter Artifacts}
\label{app:starter_artifacts}

For W06--W10 the starter is the complete frozen initial repository version,
including its dependency lock files (\code{uv.lock}, \code{pdm.lock},
\code{Cargo.lock} as the project uses); the final version is reserved as the
reference. For W01--W05 the starter is an unimplemented skeleton of the new
program specified by the authoring LLM from a human-selected functional
brief. It supplies package structure and signatures plus public contract
tests, without the original project's implementation or the newly authored
reference code. Both kinds declare their
entry points and their agent-runnable public-test command in the manifest
(\code{entrypoints.cli}, \code{entrypoints.smoke}, \code{public\_tests.command},
\code{public\_tests.dependencies} with exact pins).

The starter tree is identified by \code{starter\_repo\_digest}. Frozen
dependency fixtures are documented in
\code{frozen\_fixture\_exemptions.json}, with schema
\code{frozen\_pack\_fixture\_exemptions\_v1}. Its 20 entries record the
fixture path, pack and tree, scanner rule, SHA-256, byte count, source
reference, and retention reason. The allowlist preserves the byte-identical
fixtures required by the pinned dependencies.

\subsubsection{Evaluator Artifacts}
\label{app:evaluator_artifacts}

Each pack contains \code{tests/hidden/}, including the hidden suite and
\code{specs.json} scoring-unit declaration; \code{issues/public/} for
agent-visible requirements; \code{issues/heldout/} for evaluator-only
requirements; and \code{reference\_repo/} for the reference implementation.
The manifest freezes scoring-unit counts, qualification outcomes, and the
entry points used by the evaluator.

Hidden assets are loaded into an in-process vault keyed by a 32-byte random
token. World serialization retains an
\code{oss\_evaluator\_vault\_binding\_v1} token binding; the vault resolves
the private content at evaluation time using constant-time comparison.
\code{run\_oss\_hidden\_tests} and
\code{materialize\_and\_run\_oss\_hidden\_tests} return aggregate fields
such as \code{passed}, \code{failed}, \code{total}, \code{pass\_rate},
\code{issue\_fix}, and \code{issue\_fix\_rate}.

Qualification and candidate scoring use
\code{build\_time\_machine\_evaluation\_plan} and
\code{evaluate\_time\_machine\_candidate}. Per-contract admission is
resolved through \code{oracle\_blocking\_reasons}. These entry points
connect the frozen cases, baseline status, and evaluated candidate state to
the scoring records described in Appendix~\ref{app:evaluator_outputs}.

\subsubsection{Containers and Dependencies}
\label{app:containers_dependencies}

The container evaluator uses a digest-pinned base image and a per-pack layer
containing pinned test dependencies. The formal container path accepts
\code{docker} or \code{apptainer}, requires
\code{trust\_level = untrusted} and
\code{network\_enabled = False}, and validates an image reference of the
form \code{sha256:[0-9a-f]\{64\}}. The execution platform is
\code{linux/amd64}. The repository build script creates the image, and
\code{ORG\_EVALUATOR\_CONTAINER\_IMAGE} supplies its concrete identifier
at the execution site. A \code{formal\_digest\_pinned\_image\_required}
status records a failed image-policy check.

Network isolation combines process-level and test-wrapper controls.
\code{HTTP\_PROXY}, \code{HTTPS\_PROXY}, and \code{ALL\_PROXY} point
to the closed loopback endpoint \code{http://127.0.0.1:9}; the hidden-test
wrapper restricts \code{socket.socket.connect} to loopback. The search
system operates on its internal and frozen corpora through the separate
interface in Appendix~\ref{app:read_semantics}.

The host development path evaluates an exported candidate tree in a separate
process. Execution-policy and evaluator receipts identify the configuration
used by each run, together with the candidate, test, and evaluator digests.
Container execution and host-process execution therefore have explicit,
recorded configurations for reproducing the corresponding result.

\subsubsection{Versioning and Integrity Receipts}
\label{app:manifest_integrity}

The manifest schema \code{org\_env\_oss\_time\_machine\_pack\_v2}
records \code{starter\_repo\_digest}, \code{reference\_repo\_digest},
\code{hidden\_suite\_hash}, \code{evaluator\_environment\_hash},
\code{qualification\_hash}, \code{qualification\_result\_hash},
\code{result\_hash}, \code{runtime\_test\_status},
\code{agent\_filesystem\_isolation\_status},
\code{oracle\_reachability}, and \code{component\_map}, alongside the
entry-point and test declarations. Run records carry the artifact-identity
fields that associate an evaluated outcome with its frozen pack.

\code{evaluator\_environment\_hash} is computed from the evaluator source.
A scoring-code change produces a new evaluator identity, while the recorded
results retain the identity of the evaluator that produced them. The
verification chain joins the dataset manifest, starter, reference, hidden
suite, evaluator, tool surface, condition, event graph, and run manifest.
Checkpoints additionally carry payload and sidecar digests. The recorded
digest values and their frozen manifests support artifact matching during
qualification, aggregation, and resume.

\subsection{Task and Evaluator Construction}
\label{app:task_evaluator_construction}

\subsubsection{Authoring Provenance and Freeze Order}
\label{app:authoring_provenance}

For W01--W05, authoring proceeds from the human-selected functional brief to
technical documentation and feature specification, a new reference
implementation, and then feature tests and verification logic. The original
authoring trajectories are retained with the constructed artifacts and
endpoint-qualification records. These records document how the specification,
reference, and retained scoring cases were produced.

For W06--W10, the frozen initial/final repository pair anchors the feature and
issue requirements and their acceptance checks. Both construction paths
qualify each retained case on the starter and reference before B0--B3
rollouts. The resulting tests, references, and scoring configuration remain
fixed during the evaluated organizational runs.

\subsubsection{Agent-Visible Task Briefs}
\label{app:visible_briefs}

Briefs are rendered from frozen pack manifests. Repository-based briefs contain the product summary, public bug-fix and feature requirements, acceptance hints, entry points, and public-test command. Construction briefs contain the specification, progressive steps, skeleton layout, and public contract tests. Members also receive the artifacts and results shared during their run. 

\subsubsection{Exposed and Held-Out Behavioral Cases}
\label{app:behavioral_cases}

An \emph{exposed} unit corresponds to a requirement the organization can see: a
seeded public issue, or a step of a published progressive contract. A
\emph{held-out} unit is a requirement in \code{issues/heldout/} that is never
shown to members. Its contracts are scored offline on final mainline states and saved checkpoints across all three models.
In both cases the
\emph{content} of the scoring test is evaluator-only; what an exposed unit gives
the organization is the requirement, plus a public acceptance check derived from
it, not the hidden assertion.

The denominators depend on the scoring granularity declared by each pack,
which is separate from its repair or construction label. In the frozen
manifest records, contract-level scoring gives 16 units for W07, 13 for W08,
4 for W04, and 9 for W03. Other packs decompose hidden contracts into
behavioural cases: \code{W01} has 5 contracts scored in 35 cases and
\code{W02} has 9 contracts scored in 40 cases. Appendix~\ref{app:metrics_statistics} defines the rate aggregation for these scoring units. 

\subsubsection{Progressive and Complete Contracts}
\label{app:contracts}

A complete contract requires every constituent behavioral case to pass;
partial credit remains at case level. In Mini Blobstore (W01), the five
progressive contracts contain seven cases each. Pre-run qualification gives
0/7 passing cases for the starter and 7/7 for the reference within every
contract, yielding 0/5 and 5/5 complete contracts, respectively.

Agent outcomes are evaluated on \emph{mainline}. Workspace results are
reported separately (Appendix~\ref{app:workspace_mainline_results}). 

\subsection{Quality Control and Preflight}
\label{app:benchmark_qc}

\subsubsection{Starter and Reference Sanity}
\label{app:starter_reference_sanity}

The preflight entry points are:
\begingroup\footnotesize
\begin{verse}
\code{PYTHONPATH=. python tools/preflight\_pack\_environment.py --score <pack>}\\
\code{PYTHONPATH=. python tools/check\_pack\_health.py <pack>}\\
\code{PYTHONPATH=. python tools/print\_evaluator\_hashes.py <pack>}
\end{verse}
\endgroup

The first command executes the three qualification gates in
Appendix~\ref{app:inclusion_criteria}, including actual scoring of the
starter and reference. The second checks issue-to-file routing, artifact
resolution, oracle qualification, held-out visibility, public-suite size,
and editable-file size. The third prints the evaluator-environment and
qualification-plan hashes used as required batch-runner arguments.

Qualification receipts are written to \code{manifest.yaml} through
\code{runtime\_test\_status}, \code{qualification\_hash}, and
\code{qualification\_result\_hash}, and to
\code{provenance/authoring\_report.json}. The launch preflight validates
these receipts against the selected frozen workload.

\subsubsection{Evaluator Determinism}
\label{app:evaluator_determinism}

Evaluator-source hashing supplies \code{evaluator\_environment\_hash};
the qualification payload supplies the plan hash. Scoring hashes the
candidate tree before and after execution and raises an error on a detected
mutation. The recorded tree, evaluator, and plan identities link each score
to the exact evaluated state.

Malformed or erroring qualification contracts receive contract-level error
statuses and are excluded by the qualification gate. CI and public-test
infrastructure faults have the explicit statuses
\code{ci\_infrastructure\_error} and
\code{public\_tests\_infrastructure\_error}. Test outcomes and execution
faults consequently retain their own status fields through analysis.

\subsubsection{Lower and Upper Bounds}
\label{app:evaluator_bounds}

Every retained scored case in W01--W10 is qualified against both the
untouched starter and the fixed reference. The starter fails each retained
case and the reference passes it, yielding 0\% and 100\% qualification
scores on each retained acceptance suite. W01--W05 use the newly authored
reference programs; W06--W10 use the frozen final upstream versions.

Qualification failure blocks admission with an explicit reason, including
\code{baseline\_not\_failed:<test\_id>:<status>} and
\code{reference\_not\_passed:<test\_id>:<status>}. Test source, references,
and scoring rules are frozen before organizational rollout and versioned
independently of release packaging.

\subsubsection{Leakage and Identity Checks}
\label{app:leakage_checks}

Four automatic checks validate the member-facing interfaces: artifact
inspection checks hidden and reference content, search tests inspect
retrievable results, perception tests inspect rendered packets, and snapshot
tests verify the aggregate-count representation. The formal world also
validates evaluator-vault bindings at initialization.

A post-run future-recall check compares identifiers written by agents with
identifiers found only in evaluator-side future versions. Together with the
runtime access checks, these tests connect the information invariants in
Table~\ref{tab:invariants} to concrete artifact, search, perception,
snapshot, and output records.

\subsection{Completed Main Study and Evaluation Coverage}
\label{app:pilot_vs_full_suite}

\subsubsection{Main-Study Matrix}
\label{app:current_pilot_manifest}

The main study contains 360 runs: ten workloads, three seeds, and B0--B3 under each of three models. GPT-5.6 Terra, Claude Opus 4.6, and DeepSeek-V4-Flash each contribute 120 runs. Endpoint evaluation and 24-step checkpoint curves cover the three-model study.

\subsubsection{Analysis Populations}
\label{app:coverage_reconciliation}

\begin{table}[H]
\centering
\small
\begin{tabularx}{\linewidth}{@{}L C{0.20\linewidth} L@{}}
\toprule
\rowcolor{PanelFill}
\textbf{Analysis} & \textbf{Runs} & \textbf{Design} \\
\midrule
Main endpoints and checkpoint curves & 360 & Three models; ten workloads; three seeds; B0--B3. \\
Terra+Opus auxiliary analyses & 240 & Two models; ten workloads; three seeds; B0--B3. \\
Autonomous-protocol census & 60 & Terra+Opus B3 runs. \\
\bottomrule
\end{tabularx}
\caption{\textbf{Analysis populations.}}
\label{tab:coverage_reconciliation}
\end{table}

\subsubsection{Workload Manifest Table}
\label{app:workload_manifest_table}

\begingroup
\centering
\scriptsize
\setlength{\tabcolsep}{3.5pt}
\def\RelicCurrentTable{tab:workload_manifest}
\setlength{\LTcapwidth}{\linewidth}
\arrayrulecolor{RuleGray}
\begin{longtable}{P{2.85cm} P{1.45cm} c c c c c P{1.9cm}}
\caption{Scoring-unit inventory in the frozen W01--W10 order.}\label{tab:workload_manifest}\\
\toprule
\relictabletitle{8}{Frozen scoring-unit inventory}
\textbf{Pack} & \textbf{Family} & \textbf{Seeded} & \textbf{Exposed} & \textbf{Held-out} & \textbf{Contracts} & \textbf{Cases} & \textbf{Status} \\
\midrule
\endfirsthead
\multicolumn{8}{@{}l@{}}{\normalfont\footnotesize \textbf{Table~\thetable{} (continued)}}\\[3pt]
\toprule
\relictabletitle{8}{Frozen scoring-unit inventory}
\textbf{Pack} & \textbf{Family} & \textbf{Seeded} & \textbf{Exposed} & \textbf{Held-out} & \textbf{Contracts} & \textbf{Cases} & \textbf{Status} \\
\midrule
\endhead
\midrule
\multicolumn{8}{@{}r@{}}{\normalfont\scriptsize Continued on next page.}\\
\endfoot
\bottomrule
\endlastfoot
\code{W01} & construction & 5 & 35 & 0 & 5 & 35 & main study + case \\
\code{W02} & construction & 7 & 7 & 2 & 9 & 40 & main study \\
\code{W03} & construction & 7 & 7 & 2 & 9 & 9 & main study \\
\code{W04} & construction & 3 & 3 & 1 & 4 & 4 & main study \\
\code{W05} & construction & 3 & 3 & 1 & 4 & 4 & main study \\
\code{W06} & repair & 12 & 12 & 4 & 16 & 16 & main study \\
\code{W07} & repair & 12 & 12 & 4 & 16 & 16 & main study \\
\code{W08} & repair & 9 & 9 & 4 & 13 & 13 & main study \\
\code{W09} & repair & 12 & 12 & 4 & 16 & 16 & main study \\*
\code{W10} & repair & 12 & 12 & 4 & 16 & 16 & main study \\*

\end{longtable}
\arrayrulecolor{black}

\endgroup

\paragraph{Scoring-unit identity check.}
For W03, the manifest declares seven public issue units and two held-out
issue units. Scored B2/B3 endpoint rows carry \code{exposed\_total = 7}
and \code{heldout\_total = 2}, matching the nine contract-level behavioral
cases in Table~\ref{tab:workload_manifest}. This field-level join ties the
exposure split to the frozen task definition.

\section{Experimental Conditions and Run Protocol}
\label{app:conditions}

\subsection{Exact B0--B3 Definitions}
\label{app:condition_definitions}

Conditions are declared as a frozen dataclass with eleven fields and resolved
by condition identifier. The reported configurations are
\code{b0\_single\_agent\_founder}, \code{b1\_persistent\_role\_org},
\code{b2\_policy\_conditioned\_org}, and \code{b3\_full\_relic}.
Each run retains the resolved condition and its configuration fingerprint.

\subsubsection{B0: Reflective Individual}
\label{app:b0_definition}

B0 uses one member with \code{action\_selection\_mode = llm\_direct}.
The member is constructed from the canonical eight-member roster: skill
coverage is the per-domain maximum, while decision priors, communication
style, and work rhythm use roster means. It has a private workspace,
sandbox, and memory with the shared salience threshold and capacity limit,
plus the common registered actions, tools, and task-visible information.

Operational tasks, artifacts, budget records, and repository state persist
throughout the run. \code{institutionalization\_enabled} and
\code{capability\_learning\_enabled} are false. Appraised observations
enter private memory under the shared salience and capacity rules;
inspected or retrieved material enters the bounded context for subsequent
LLM-direct decisions.

\subsubsection{B1: Long-Horizon Role-Based Multi-Agent System}
\label{app:b1_definition}

B1 uses eight persistent members with fixed functional roles, a shared
company workspace, channels, meetings, and private member memory. The roster
and member-local state persist from $t=0$ through $t=336$;
\code{temporary\_team = false} retains them across episode and sprint
boundaries. The model selects from the feasible candidate pool through
\code{action\_selection\_mode = llm\_direct}.

\code{profile\_conditioning\_enabled},
\code{capability\_learning\_enabled}, and
\code{institutionalization\_enabled} are false. Role-scoped affordances,
shared artifacts, communication, and repository delivery remain available
through the common execution substrate. Shared workspace and private
member memory retain their respective visibility and storage interfaces.

\subsubsection{B2: SDL-Based Long-Horizon Multi-Agent System}
\label{app:b2_definition}

B2 uses the same eight-member team and work environment as B1, with SDL action selection (\code{action\_selection\_mode = profile\_policy}), profile conditioning, and member-local skill, reputation, and authority updates.

Protocol proposal, adoption, compilation, and revision are disabled. The
registered institutional actions are \code{propose\_protocol},
\code{amend\_protocol}, \code{support\_protocol}, and
\code{follow\_protocol}. They may enter the candidate menu; with
institutionalization disabled, their handlers return
\code{mechanism\_ablation:institutionalization}. The wish-to-proposal
cadence, adoption, compilation, and registration path are skipped, while
member-local learning, private memory, and shared artifacts continue to
accumulate.

B1--B2 compares the combined addition of SDL, profile conditioning, and member learning. B3--B2 enables the protocol lifecycle on that shared backbone. 

\subsubsection{B3: Persistent Protocol-Governed Organization}
\label{app:b3_definition}

B3 shares the B2 roster, model, tools, candidate construction, features, SDL parameterization, member learning, and execution handlers. Enabling \code{institutionalization\_enabled} activates reflection-to-wish, proposal, validation, governance, and protocol compilation. Adopted protocols enter $R_t$ and $Q_t$, supplying decision features, responsibility signals, and post-handler protocol checks and records. Members can amend and retire adopted rules through the same lifecycle. 

\subsection{Roster, Functions, and Decision Priors}
\label{app:roster_priors}

\subsubsection{Roster Size and Functional Specialties}
\label{app:roster_size}

The canonical roster is eight members, used in full by \Bone{}, \Btwo{} and
\Bthree{}, and aggregated into one member for \Bzero{}. The functional
specialties are a leadership and direction function, two implementation
functions differing in speed-versus-care emphasis, a reliability and verification
function, and four further functions covering review, documentation, customer
signal and coordination.  

\subsubsection{Roles, Responsibilities, and Permissions}
\label{app:role_permissions}

The board starts with unassigned tasks, and members allocate ownership during
execution. Role-scoped affordances determine which functions receive edit
and review candidates and which may approve releases; these affordances are
identical across B1, B2, and B3. Release approval requires a lead role and
the reliability function, capped by roster size for multi-member conditions.
Object permissions follow Appendix~\ref{app:visibility_propagation}, and
escalation combines the registered escalation action with the governance
deadlock rule.

The declared-factor conformance test checks that all multi-member conditions
use the same governance role topology. The comparison therefore retains the
same functional approval structure while varying the declared mechanism
switches.

\subsubsection{Functional Decision Priors}
\label{app:functional_priors}

A prior $p_i(t)$ is a member-specific map of named, real-valued profile and
skill dimensions initialized from the canonical roster. It enters the
32-dimensional SDL representation in Appendix~\ref{app:sdl_details}.
The reported assignment is \code{aligned}, with functional conditioning
fixed before the runs and shared by B2 and B3.

Profile-backed inputs remain fixed, while skill-backed inputs evolve through
the common member-learning path. The
\code{profile\_conditioning\_enabled} switch activates these prior inputs
in B2 and B3. The same roster initialization and profile assignments are
used for the two configurations.

\subsubsection{Member Initialization}
\label{app:member_initialization}

Every member starts with: empty private memory; an empty personal workspace and
an empty sandbox; no owned tasks; default continuous condition
(\code{attention} 1.0, \code{morale} 0.6, \code{trust\_in\_company} 0.7,
\code{perceived\_recognition} 0.5, \code{role\_clarity} 0.4, the remaining
variables 0); the roster's literal prior and skill maps; and eight-domain
reputation and authority at their configured baselines. Task knowledge is
whatever the brief and the starter tree contain
(Appendix~\ref{app:visible_briefs}); nothing about the hidden suite, the
reference or the held-out requirements is present in any member's initial state.

Randomization at initialization is limited to availability registration, which is
seeded. Under fresh-member turnover, the same initialization procedure reconstructs
member-local state for the target organization while preserving the functional
role slots. A fresh member therefore starts with the same kind of local state
as a member at $t=0$ of an ordinary run
(Appendix~\ref{app:fresh_roster}). 

\subsection{Prompts and Information Controls}
\label{app:prompt_controls}

Relic separates selection of \emph{what to do} from model calls used to carry
out or reflect on a selected activity. B0 and B1 ask an LLM to choose one action
from a constrained candidate menu. B2, B3, B3-text, Transfer Text, and Transfer
Exec use SDL for what-action selection and issue no LLM action-selection prompt.
Other model calls remain module-specific, including code/document editing,
reflection, proposal drafting, and proposal evaluation.

\begin{table}[H]
\centering
\scriptsize
\setlength{\tabcolsep}{2.3pt}
\begin{tabularx}{\linewidth}{@{}L *{7}{C{0.082\linewidth}}@{}}
\toprule
\rowcolor{PanelFill}
\textbf{Prompt stage} & \textbf{B0} & \textbf{B1} & \textbf{B2} & \textbf{B3} & \textbf{B3-text} & \textbf{Text} & \textbf{Exec}\\
\midrule
Shared grounding / identity & cond. & cond. & cond. & cond. & cond. & cond. & cond.\\
What-action selection & LLM & LLM & SDL & SDL & SDL & SDL & SDL\\
Code/document editing & cond. & cond. & cond. & cond. & cond. & cond. & cond.\\
Reflection & cond. & cond. & cond. & cond. & cond. & cond. & cond.\\
Separate LLM wish extraction & --- & --- & --- & --- & --- & --- & ---\\
Wish $\rightarrow$ proposal & --- & --- & --- & cond. & cond. & --- & ---\\
Proposal evaluation & --- & --- & --- & cond. & cond. & --- & ---\\
Approval-specific LLM call & --- & --- & --- & --- & --- & --- & ---\\
Readable rules in code-editor context & --- & --- & --- & avail. & avail. & 6 & 6\\
\bottomrule
\end{tabularx}
\caption{Prompt routing across conditions. ``cond.'' denotes a module call when
that activity occurs.}
\label{tab:prompt_routing}
\end{table}

The wish stage maps and filters reflection-generated \code{improvement\_ideas}. Proposal evaluation supplies scores and revisions; governance actions supply approval and adoption.

\subsubsection{Shared System and Grounding Template}
\label{app:shared_prompts}
Cognitive modules compose the system message from a shared workload/company
brief, global grounding rules, the module name, and the module-specific
instruction. The member identity block---role mandate, enabled functional
conditioning, and current memory/context---is prepended to the user message.
The grounding rules require the model to use supplied objects and identifiers,
select only available actions and valid targets where applicable, avoid
inventing outcomes, and return the requested structured output.

\subsubsection{B0/B1 Action-Selection Prompt}
\label{app:condition_prompt_differences}
For B0/B1, the user message presents the member-visible action context and asks
for one exact candidate:
\begin{quote}\ttfamily\small
Choose the next intentional action for \{agent\_id\} at tick \{tick\}.\\[2pt]
\{rendered\_action\_context\}\\[2pt]
Return ONLY JSON matching the action schema; candidate\_action and target must
identify one exact entry from candidate\_options.
\end{quote}
The returned choice is validated against the constrained candidate pool before
execution. B2/B3 and all text/transfer variants do not use this prompt because
SDL performs the corresponding selection.

\subsubsection{Code and Document Editing}
After an edit action has been selected, the code editor receives the target
file, current content, available public evidence, the edit goal, and any
readable organizational rules applicable to the treatment. The editor returns
anchored replacements rather than a whole-file rewrite. In B3/B3-text and both
transfer treatments, readable rules enter the model through a dedicated block:
\begin{quote}\ttfamily\small
HOW THIS ORGANIZATION WORKS:\\
- \{readable rule 1\}\\
- \{readable rule 2\}\\
\ldots
\end{quote}

\subsubsection{Formation Prompts: Reflection, Wishes, and Proposals}
Reflection receives recent work, failures, relevant episodes, product context,
memory, and existing protocol context and asks for structured self/team
assessment, blockers, and concrete \code{improvement\_ideas}. A
\code{protocol\_need} is used when recurring avoidable failure suggests a
standing, checkable rule. The main formation path then maps these ideas into
wishes; no separate LLM wish-extraction call is used.

A qualifying wish is drafted into a proposal with the visible wish record,
available registered actions, and product context. The proposal instruction
requires a concrete checkable requirement, names required approvals, and does
not allow the model to assume adoption. Proposal evaluation separately returns
feasibility, usefulness, risk, adoption score, blocking issues, and suggested
revision. Recurring-pattern synthesis can also produce a candidate protocol;
it remains a proposal until the governance path adopts it.

The exact prompt text, readable transfer rules, and wish-to-protocol measurement
criteria are given in Appendix~\ref{app:exact_prompts}.

\subsubsection{Context Rendering}
Rendering is section-ordered and budget-aware. Candidate menus, task boards,
inboxes, pull requests, test results, and product context retain the information
needed to act; narrative history/reflection sections use bounded rendering.
Inbox triage and private-memory capacity provide additional shared limits across
conditions.

\subsection{Matched Experimental Controls}
\label{app:matched_controls}

\subsubsection{Model and Decoding Configuration}
\label{app:model_configuration}
\label{app:model_config}

The main study uses GPT-5.6 Terra, Claude Opus 4.6, and DeepSeek-V4-Flash,
with 120 runs per model. The DeepSeek runs use the provider model identifier
\code{deepseek-ai/DeepSeek-V4-Flash-0731}. GPT-5.6 Terra response identifiers include
\code{gpt-5.6-terra-2026-07-09} and the unversioned family name. Internal
transfer uses GPT-5.6 Terra; ProgramBench uses GPT-5.6 with \code{xhigh}
reasoning effort. The in-situ binding ablation uses Claude Opus 4.6.
The same-model CooperBench comparison uses Claude Opus 4.6 with high
reasoning effort.

Client transport documentation specifies a \code{chat\_completions}
route, a 180-second request deadline, six retries, and exponential backoff
starting at three seconds. Connect, read, write, and pool timeouts have
separate configuration fields. Each run retains its provider/endpoint
routing, wire API, output limits, decoding fields, and request policy in
the client configuration record. Reasoning effort and temperature are
separately identified fields.

A model-call failure propagates as a failed decision or enters the
module-specific fallback path. \code{llm\_runtime\_identity} records
\code{observed\_response\_models} as a counter of returned model strings,
an endpoint hash, a routing-context fingerprint, and a chained response-ID
digest. \code{llm\_runtime\_fingerprint} hashes that identity block.
Requested model identifiers, observed response identifiers, and routing
metadata thus remain directly inspectable for each recorded configuration.

\subsubsection{Tools and Registered Actions}
\label{app:matched_tools}

One action registry and wiring function install the same candidate mapper,
feature extractor, policy implementation, and execution adapters across the
conditions. Condition switches choose LLM-direct or SDL action selection
and activate member learning and institutionalization. The run record's
\code{tool\_surface\_fingerprint} identifies the offered tool and action
surface used by that configuration.

Protocol and tool creation pass through B3 governance. The four institutional
actions are named in Appendix~\ref{app:b2_definition}; their registry
presence and condition-dependent execution are checked independently of
the common repository-delivery actions.

\subsubsection{Task-Visible Information}
\label{app:matched_information}

The pack and seed fix the starter tree, public requirements or specification,
public-test command, entry points, and issue timing. Public-test outcomes
are computed on each condition's candidate state. The analysis joins paired
conditions through the manifest, starter, hidden-suite, and evaluator
identities retained in the run and evaluator records.

The aggregation pipeline validates starter digests before pooling paired
conditions. A revised public gate or evaluator produces a versioned asset
identity, and the analysis maps each outcome to its designated frozen task
configuration. This join preserves the correspondence among the task,
candidate artifact, scoring procedure, and reported comparison.

\subsubsection{Execution Horizon and Resource Ceilings}
\label{app:matched_budgets}
\label{app:resource_matching}

The main-study matrix uses a 336-step horizon, checkpoints every 24 steps,
168-step sprints, and disabled work rhythm. Token expenditure is measured
per run, with no matched hard token, model-call, or primary-action ceiling.
Company-balance accounting is soft: charging clamps the remaining balance
and records violation flags, while action execution remains governed by the
action and repository validity checks.

The runner also supports a company-wide resource-ledger configuration with
\code{max\_llm\_calls}, \code{max\_llm\_requested\_tokens},
\code{max\_llm\_prompt\_characters}, \code{max\_primary\_actions},
and \code{max\_ticks}. When enabled, its append-only usage and exhaustion
signal operate at organization scope. The run configuration records the
active horizon and resource policy; the reported main-study results use
the scheduling-horizon setting above.

\subsubsection{Seeds and Randomness}
\label{app:seeds_randomness}

Each model--workload--condition combination uses seeds 1401, 2711, and 4013. The mechanism case uses seed 1401. 

Randomness is namespaced from the run seed rather than drawn from one global
stream: policy jitter and softmax sampling use
\code{Random(f"\{seed\}:\{tick\}:\{agent\}")}; availability registration, sandbox
timing, the simulated community and substrate controls each derive their own
stream; and the shuffled-prior transform uses a prefixed string seed. Model calls use the configured provider sampling. Within each step, members are served in roster insertion order.

\subsection{Run Protocol}
\label{app:run_protocol}

\subsubsection{Initialization and Preflight}
\label{app:run_initialization}

Before launch, the pack passes the three qualification gates with
\code{--score}. The evaluator and qualification-plan hashes are printed
and supplied as required runner arguments. The code tree is copied into a
run-specific frozen directory, and the public suite is executed on both
starter and reference to verify red/green acceptance behavior. The runner
then initializes organizational state from the selected manifest.

Evaluator assets are loaded into the token-keyed vault, with the binding
stored in world state. When prompt auditing is enabled, hashing starts at
the first intercepted call and the audit receipt records its call coverage.
The initialized run therefore retains the code, task, evaluator, and
information-interface identities used by its subsequent execution.

\subsubsection{Online Execution}
\label{app:online_execution}

Each step: advance the clock; at a day boundary run the daily passes (growth where
enabled, budget, structural detectors); process meetings and background jobs; then
for each member in roster order, triage the inbox, build perception, construct and
guard candidates, select, execute, appraise into memory, and write graph edges.
Governance passes run on their own cadences (Algorithm~\ref{alg:a2}). State is
written to a checkpoint every 24 steps together with token totals, recorded at the same checkpoint.

\subsubsection{Checkpoint and Endpoint Grading}
\label{app:checkpoint_grading}

Checkpoint scoring evaluates saved states offline with a 120-second
per-contract timeout. Endpoint grading evaluates final mainline state.
The reported endpoint and 24-step checkpoint series cover the three-model,
360-run study, with the tree type, scoring units, and metric eligibility
attached to each evaluated series.

Each checkpoint couples the saved organization and candidate state with
its token totals and evaluated outcomes. Figure~\ref{fig:verified-production-outcomes}
uses these recorded checkpoint steps; the cost sensitivity selects saved
checkpoints according to Appendix~\ref{app:budget_capped_sensitivity}.

\subsubsection{Stopping, Submission, and Resume}
\label{app:run_stopping}

Runs end at the scheduling horizon. Interrupted execution resumes from the last saved checkpoint under the same run identity. Appendix~\ref{app:retry_policy} specifies provider retries and recovery.

\RelicNeedTable{tab:run_matrices}{3}
\subsubsection{Run Matrix and Inventory}
\label{app:run_matrix}

\begingroup
\centering
\footnotesize
\setlength{\tabcolsep}{3.5pt}
\def\RelicCurrentTable{tab:run_matrices}
\setlength{\LTcapwidth}{\linewidth}
\arrayrulecolor{RuleGray}
\begin{xltabular}{\linewidth}{P{3.2cm} P{1.25cm} P{2.4cm} L}
\caption{Experimental accounting.}\label{tab:run_matrices}\\
\toprule
\relictabletitle{4}{Experimental accounting}
\textbf{Evidence line} & \textbf{Runs} & \textbf{Design / coverage} & \textbf{Configuration} \\
\midrule
\endfirsthead
\multicolumn{4}{@{}l@{}}{\normalfont\footnotesize \textbf{Table~\thetable{} (continued)}}\\[3pt]
\toprule
\relictabletitle{4}{Experimental accounting}
\textbf{Evidence line} & \textbf{Runs} & \textbf{Design / coverage} & \textbf{Configuration} \\
\midrule
\endhead
\midrule
\multicolumn{4}{@{}r@{}}{\normalfont\scriptsize Continued on next page.}\\
\endfoot
\bottomrule
\endlastfoot
GPT-5.6 Terra main study & 120 & $10\times3\times4$ & Ten workloads, three seeds, B0--B3. \\
Claude Opus 4.6 main study & 120 & $10\times3\times4$ & Ten workloads, three seeds, B0--B3. \\
DeepSeek-V4-Flash main study & 120 & $10\times3\times4$ & Ten workloads, three seeds, B0--B3; included in the pooled endpoint table. \\
Internal transfer & 60 & 30 Text + 30 Exec & Ten workloads $\times$ three seeds per arm; 30 GPT-5.6 Terra B2 main-study references reused. \\
In-situ binding ablation & 30 new & B3-text; Claude Opus 4.6, seeds 1401/2711/4013 & Thirty matched B3 references reused from the Claude Opus 4.6 main study. \\
Selected case & included & One matched GPT-5.6 Terra main-study unit & W01; GPT-5.6 Terra; seed 1401. \\*
ProgramBench & separate & Same 25 tasks, two systems & External protocol plug-in; no SDL. \\*
CooperBench & separate & Full 652-pair benchmark + 47-pair same-model subset & Two-member Relic; full released-reference comparison plus same-model Solo/Peer control using the same broken-pair exclusion rule. \\*
Human--agent extension & walk\-through study & P2/P3 walkthroughs & Human-facing interface evaluation. \\
\end{xltabular}
\arrayrulecolor{black}

\endgroup

\subsection{Condition-Conformance Tests}
\label{app:condition_conformance}

\subsubsection{Mechanism-Isolation Tests}
\label{app:mechanism_isolation}

The declared-factor test enumerates the fields allowed to change at each
B0--B3 step and compares them with the resolved condition definitions.
For B2--B3, the allowed mechanism switch is
\code{institutionalization\_enabled}. The shared wiring installs the
candidate generator, feature extractor, decision layer, execution operator,
registry, model, and member-information interfaces for each condition.

An action-reachability test checks the offered candidates for the delivery
chain: edit, commit, open, review, CI, merge, and release. It checks the
chain in both B0 and B3. When the candidate-pool capacity limit binds,
delivery actions are retained first. The conformance tests inspect this
pool-priority rule and the shared governance role topology alongside the
condition switches.

\subsubsection{Configuration, Horizon, and Information Matching}
\label{app:resource_information_matching}

The four conditions for a workload and seed are launched from one plan that
fixes workload, horizon, seed, and model. Run records retain
\code{dataset\_manifest\_hash}, \code{starter\_repo\_digest},
\code{hidden\_suite\_hash}, \code{evaluator\_environment\_hash},
\code{tool\_surface\_fingerprint}, \code{ablation\_fingerprint},
\code{llm\_runtime\_fingerprint}, \code{seed}, and
\code{action\_selection\_mode}. These fields bind results to the resolved
configuration and support the workload/version/eligibility joins used by
paired aggregation.

\subsubsection{Member-Information and Access Tests}
\label{app:forbidden_access}

Access tests cover evaluator-private assets, future repository history,
private member workspaces, and live-network access. Artifact, search,
perception, and snapshot checks are specified in
Appendix~\ref{app:leakage_checks}. The optional prompt-visibility
interceptor records the intercepted calls, and the future-recall checker
inspects generated identifiers. The execution-layer proxy and socket
controls in Appendix~\ref{app:containers_dependencies} complement these
member-facing checks.

\section{Internal Protocol Transfer and Executable Binding}
\label{app:transfer}

The internal-transfer experiment uses 30 Text and 30 Exec GPT-5.6 Terra target runs over ten workloads and three seeds. Fresh uses the corresponding 30 B2 main-study runs. Text and Exec share a source-derived frozen rule package and vary its executable binding.

\subsection{Protocol Source and Bundle Freezing}
\label{app:source_selection}

\subsubsection{Source-Run Eligibility}
\label{app:source_run_eligibility}

The reported transfer package is constructed from B3-formed protocols using recurring
capability families observed across source runs as the selection frame rather than target
performance. From these families we selected functionally distinct representative
concerns and normalized them into four source concerns and six closed-set typed runtime
guards. This selection did
not use target-evaluation outcomes. Normalization was limited to making rule semantics
mechanically executable in the target runtime: timed, multi-head, or otherwise
unobservable clauses were removed or rewritten; triggers, required fields/steps,
affected actions, violation conditions, reason codes, and machine bindings were made
explicit. Compatibility checks verified trigger observability and action reachability. Table~\ref{tab:transfer_bundle_mapping} shows the public
functional mapping.

The final package uses bundle schema \code{org\_capability\_bundle\_v2} and binding
schema \code{org\_protocol\_bindings\_v2}. It was frozen before the reported target
evaluation. Text and Exec use exactly the same frozen readable content; the treatment
contrast is the presence or absence of executable bindings.

\begingroup
\centering
\scriptsize
\def\RelicCurrentTable{tab:transfer_bundle_mapping}
\setlength{\LTcapwidth}{\linewidth}
\arrayrulecolor{RuleGray}
\begin{xltabular}{\linewidth}{P{1.0cm} P{3.0cm} P{3.0cm} L}
\caption{Public functional mapping of four source-protocol concerns to six compiled runtime guards.}\label{tab:transfer_bundle_mapping}\\
\toprule
\relictabletitle{4}{Transferred protocol-to-guard mapping}
\textbf{Source concern} & \textbf{Compiled guard} & \textbf{Affected actions} & \textbf{Functional role} \\
\midrule
\endfirsthead
\multicolumn{4}{@{}l@{}}{\normalfont\footnotesize \textbf{Table~\thetable{} (continued)}}\\[3pt]
\toprule
\relictabletitle{4}{Transferred protocol-to-guard mapping}
\textbf{Source concern} & \textbf{Compiled guard} & \textbf{Affected actions} & \textbf{Functional role} \\
\midrule
\endhead
\midrule
\multicolumn{4}{@{}r@{}}{\normalfont\scriptsize Continued on next page.}\\
\endfoot
\bottomrule
\endlastfoot
S1 & issue owner assigned & \code{open\_pr} & customer triage / ownership \\
S1 & branch owned by actor & \code{open\_pr} & ownership map \\
S2 & unchanged failed CI not retried & \code{run\_ci}, \code{ci\_test} & evidence workflow \\
S2 + S3 & current CI attested & \code{merge\_pr} & workflow integration \\*
S3 & independent review & \code{review\_pr}, \code{approve\_pr}, \code{formal\_pr\_review}, \code{merge\_pr} & review gate \\*
S4 & release gate covered & \code{publish\_product\_release} & release governance \\
\end{xltabular}
\arrayrulecolor{black}
\RelicTableNotes{S1--S4 identify four source concerns; the CI-attestation guard combines S2 and S3.}

\endgroup

\subsubsection{Package Provenance and Freeze Record}
\label{app:checkpoint_selection}

The transfer record identifies the package version, content digest, selected
source items, normalization steps, and freeze time. A source-checkpoint
contribution is linked to the corresponding item. The final freeze fixes
the six compiled guards and their readable summaries before target
evaluation, and Text and Exec reference the same package identity.

Source-selection metadata, source IDs, and normalization records are
retained with the frozen package, connecting each compiled guard to the
source concern in Table~\ref{tab:transfer_bundle_mapping}.

\subsubsection{Target-Workload Selection}
\label{app:target_selection}

Targets use the W01--W10 starters, public requirements, and evaluators.
Text and Exec are matched through their target/workload/seed/configuration
mapping, yielding 30 matched units. Fresh uses the corresponding 30-run
GPT-5.6 Terra B2 stratum. Seed rates are averaged within workloads, and
applicable workloads receive equal weight.

Curation includes a reverse-reachability check: required verification steps
are mapped back to agent-reachable artifacts and public issues before task
admission. This check validates that the target action surface supplies
the work needed to satisfy the protocol's verification obligations.
Source-to-target relations and scoring identities remain linked to the
package and the target record.

\subsection{Fresh-Member Initialization}
\label{app:roster_interventions}

\subsubsection{Fresh Roster}
\label{app:fresh_roster}
The target organization is freshly initialized from the canonical eight-role roster.
Member-local state is reset: private memory, personal workspace contents, sandbox cache,
message read/acknowledgement state, commitments, reflections, wishes, and accumulated
member-local experience are not inherited from the source run.  Functional roles are
reinstantiated for the target task, while the transferred organizational package is applied
according to treatment.

\subsection{Organizational-State Arms}
\label{app:transfer_arms}

\subsubsection{Exec: Executable Protocol Binding}
\label{app:f_exec}

Exec initializes fresh target members and supplies the frozen package as
readable rules with six live executable bindings. The target runtime
receives the triggers, scopes, affected actions, required steps and
evidence, responsibility functions, exceptions, and retirement conditions.
Registration and identifier remapping associate subsequent runtime events
with the imported rule and package.

Readable content is identical to Text. During target execution, existing
bindings produce use and consequence records while the package remains
frozen under the treatment in Appendix~\ref{app:transfer_formation_freeze}.

\subsubsection{Text: Content-Matched Protocol Presentation}
\label{app:f_text}

Text initializes fresh target members and presents the same six ordered
rule summaries to model-mediated work, including the code editor, through
\code{HOW THIS ORGANIZATION WORKS}. The prose preserves each rule's
trigger, scope, required steps and evidence, responsible functions,
exceptions, and consequences. The representation uses
\code{capability\_as\_prose} and the \code{TEXT\_ONLY\_DOC\_TYPE}
carrier.

Text retains SDL and member learning with zero compiled bindings for the
transferred package. The readable summaries and their order are checked
against the same frozen package as Exec; the verbatim content is reproduced
in Appendix~\ref{app:exact_prompts}.

\subsubsection{Fresh Reference: Reused B2 Main-Study Runs}
\label{app:fresh_reference}
Fresh uses the 30 GPT-5.6 Terra B2 main-study runs with ten workloads and three seeds, initialized without an inherited protocol package.

\subsubsection{Frozen Formation and the B2 Reference}
\label{app:transfer_formation_freeze}

The treatment freezes the mechanism-producing path for the entire target
window: no new protocol proposals, adoption of additional protocols, or learned
rule revisions. Ordinary code edits, tests, communication, local memory, and
the B2 member-learning paths continue. Imported executable bindings and their use/consequence records remain active while formation is frozen.

\begingroup
\centering\footnotesize
\setlength{\tabcolsep}{3.5pt}
\def\RelicCurrentTable{tab:transfer_freeze_matrix}
\setlength{\LTcapwidth}{\linewidth}
\arrayrulecolor{RuleGray}
\begin{xltabular}{\linewidth}{@{}L c c c@{}}
\caption{Why the main-study GPT-5.6 Terra B2 condition is the fresh-start reference.}\label{tab:transfer_freeze_matrix}\\
\toprule
\relictabletitle{4}{Transfer treatment freeze matrix}
\textbf{Target component} & \textbf{Fresh reference} & \textbf{Text} & \textbf{Exec}\\
\midrule
\endfirsthead
\multicolumn{4}{@{}l@{}}{\normalfont\footnotesize \textbf{Table~\thetable{} (continued)}}\\[3pt]
\toprule
\relictabletitle{4}{Transfer treatment freeze matrix}
\textbf{Target component} & \textbf{Fresh reference} & \textbf{Text} & \textbf{Exec}\\
\midrule
\endhead
\midrule
\multicolumn{4}{@{}r@{}}{\normalfont\scriptsize Continued on next page.}\\
\endfoot
\bottomrule
\endlastfoot
B2 backbone: SDL and member learning & same & same & same\\
New protocol proposal / adoption / revision & off & off & off\\
Fixed transferred protocol content & absent & present & present\\*
Transferred executable bindings & absent & absent & present\\*
Origin of counted runs & main study & target run & target run\\
\end{xltabular}
\arrayrulecolor{black}

\endgroup

\subsection{Transfer Payload}
\label{app:transfer_firewall}

\subsubsection{Permitted Payload}
\label{app:permitted_payload}

Only organizational state may cross: executable rules with their
\code{trigger\_condition}, \code{scope}, \code{affected\_actions},
\code{required\_steps}, \code{required\_fields}, \code{enforcement\_rule},
\code{enforcement\_action}, \code{exception\_rule} and \code{sunset\_rule};
their responsibility mappings over functions; and the carrier documents those
rules reference. The bundle is schema-versioned and validated on both export and
import. 

\subsubsection{Excluded Payload}
\label{app:excluded_payload}

Excluded by construction: source code and patches from the source workload; any
target patch or solution; private member memory;
hidden-test outcomes and any evaluator record; task-specific answers; and future
repository history.  

\subsubsection{Payload and Reset Records}
\label{app:transfer_receipts}
\label{app:payload_receipts}

Each transfer record contains a \code{capability\_transfer} block with
\code{roster\_origin}, \code{capability\_form},
\code{source\_repository\_id}, and \code{protocols\_injected}.
\code{experiment\_phase} is \code{capability\_transfer}, and
\code{pack} identifies the source-to-target relation. Receipt validation
checks the required fields, package contents, and source/target relation
against the selected evaluation plan.

The record stores treatment identity, package-item and carrier-document
counts, package digest, and reset state. A source-to-target role map binds
obligations to target functions. Fresh initialization reconstructs member
memory, workspaces, sandboxes, messages, commitments, reflections, and
wishes while retaining the designated external protocol package. Runtime
events refer to the injected package and rule identifiers, providing the
join from inherited organizational state to later target actions.

\subsection{Content Matching and Binding Control}
\label{app:text_binding_controls}

\subsubsection{Content-Matched Prose Construction}
\label{app:prose_construction}

Text and Exec use the same \code{canonical\_v2} package. The six ordered summaries render to an identical 3,148-character code-editor block in both treatments. Exec activates six compiled bindings; Text activates none. Appendix~\ref{app:exact_prompts} provides the shared readable text and package identifiers.

\subsubsection{Execution-Hook Removal}
\label{app:hook_removal}

Exec registers the transferred package's \code{affected\_actions}
matchers, \code{responsible\_roles} routing hooks, and execution checks.
Text retains the corresponding prose in shared work state with zero
compiled bindings. Both treatments keep the target-stage proposal,
adoption, and revision path disabled.

Conformance checks compare package identities, readable fields, and
rendering order across Text and Exec and inspect the presence of compiled
bindings in Exec and their absence in Text. Existing-binding use,
violation, and execution events are recorded separately from protocol
creation and revision, preserving the frozen target treatment throughout
the observation window.

\RelicNeedTable{tab:transfer_contrasts}{5}
\subsection{Transfer Estimands and Evaluation Scope}
\label{app:transfer_estimands}

\subsubsection{Primary Controlled Comparisons}
\label{app:transfer_contrasts}

\begingroup
\centering
\begingroup
\footnotesize
\arrayrulecolor{RuleGray}
\setlength{\tabcolsep}{3.5pt}
\renewcommand{\arraystretch}{1.23}
\def\RelicCurrentTable{tab:transfer_contrasts}
\setlength{\LTcapwidth}{\linewidth}
\begin{xltabular}{\linewidth}{@{}C{0.145\linewidth} C{0.215\linewidth} C{0.135\linewidth} L@{}}
\caption{\textbf{Internal-transfer comparisons.}}\label{tab:transfer_contrasts}\\
\toprule
\relictabletitle{4}{Internal transfer: controlled contrasts}
\textbf{Contrast} & \textbf{Reference} & \makecell{\textbf{Pass-rate}\\\textbf{effect}} & \textbf{Interpretation} \\
\midrule
\endfirsthead
\multicolumn{4}{@{}l@{}}{\normalfont\footnotesize \textbf{Table~\thetable{} (continued)}}\\[3pt]
\toprule
\relictabletitle{4}{Internal transfer: controlled contrasts}
\textbf{Contrast} & \textbf{Reference} & \makecell{\textbf{Pass-rate}\\\textbf{effect}} & \textbf{Interpretation} \\
\midrule
\endhead
\midrule
\multicolumn{4}{@{}r@{}}{\normalfont\scriptsize Continued on next page.}\\
\endfoot
\bottomrule
\endlastfoot
\textcolor{ExecDark}{\textbf{Exec--Text}} & Actual target treatments & \reliceffectcell{+6.5 pp} & Executable binding of content-matched protocols. \\*
\textcolor{ExecDark}{\textbf{Exec--Fresh}} & Reused GPT-5.6 Terra B2 equivalent & \reliceffectcell{+15.8 pp} & Comparison with the designated no-package reference. \\
\end{xltabular}

\endgroup
\arrayrulecolor{black}
\endgroup

\subsubsection{Interpretation}
\label{app:transfer_pilot_limits}
Exec--Text estimates the gain from executable binding of the content-matched source-derived package. Exec--Fresh measures the gain over fresh B2 initialization.

\section{Metrics and Statistical Analysis}
\label{app:metrics_statistics}

\subsection{Canonical Data Sources}
\label{app:canonical_sources}

\subsubsection{Repository State and Action Log}
\label{app:repo_vs_action_log}

Product and delivery outcomes use the repository objects in
\code{world.repo\_system.repo}, the product artifacts, and evaluator
outputs for the corresponding candidate state. The action log records
explicit action occurrences. Protocol lifecycle measures use the protocol
ledger and proposal histories. These source interfaces retain the object
identifiers needed to join a reported count to its underlying records.

The selected case labels repository and action counters separately and
specifies the runtime paths connecting them in
Table~\ref{tab:case_counters}.

\subsubsection{Evaluator Outputs}
\label{app:evaluator_outputs}

The final-evaluation artifact uses schema
\code{orgenv\_oss\_final\_evaluation\_v1}. Each contract or check
records \code{task\_id}, \code{command}, \code{owner}, \code{kind},
\code{base\_status}, \code{candidate\_status}, \code{required},
\code{baseline\_repo\_digest}, \code{candidate\_repo\_digest},
\code{execution\_policy\_hash}, \code{exit\_code},
\code{stdout\_hash}, and \code{stderr\_hash}.

Passing cases satisfy \code{candidate\_status == passed}; a confirmed fix
additionally satisfies \code{base\_status == failed}. Checkpoint evaluation
uses the same scoring fields at each saved step across the current
360-run study. Infrastructure and malformed-evaluation errors retain
explicit statuses and enter the metric-eligibility rules in
Appendix~\ref{app:missing_metrics}.

\subsubsection{Organizational-State and Protocol Records}
\label{app:protocol_sources}

Protocol measures join proposal-manager objects and their status histories
with the protocol event ledger. Adoption, use, violation, enforcement,
amendment, and retirement each have their own event type. Per-spec scalar
counters provide a cross-check against the corresponding ledger totals.

Distinct affected targets are reconstructed from typed governed-object
references in enforcement events. The references identify PRs, patches,
commits, and other work objects, and link the object-level consequences to
the protocol lineage that generated them.

\subsubsection{Token and Compute Receipts}
\label{app:token_sources}

Provider receipts accumulate per call in \code{usage\_totals} and are
written to the run-level \code{llm\_usage} record, including retry calls.
Input, output, and cached-token counters are retained where supplied by
the provider. Checkpoint token totals are stored with the corresponding
saved state for the matched-spend analysis.

Run duration is recorded in seconds. Provider-token accounting and elapsed
time are separate measurements; tool and container compute follows the
resource-accounting definition in Appendix~\ref{app:token_accounting}.

\RelicNeedTable{tab:metric_dictionary}{3}
\subsection{Metric Dictionary}
\label{app:metric_dictionary}

\begingroup
\centering
\scriptsize
\setlength{\tabcolsep}{3pt}
\def\RelicCurrentTable{tab:metric_dictionary}
\setlength{\LTcapwidth}{\linewidth}
\arrayrulecolor{RuleGray}
\begin{longtable}{P{3.1cm} P{3.05cm} P{2.55cm} c P{2.5cm}}
\caption{Metric dictionary.}\label{tab:metric_dictionary}\\
\toprule
\relictabletitle{5}{Metric dictionary}
\textbf{Metric} & \textbf{Numerator} & \textbf{Denominator} & \textbf{Better} & \textbf{Applicability} \\
\midrule
\endfirsthead
\multicolumn{5}{@{}l@{}}{\normalfont\footnotesize \textbf{Table~\thetable{} (continued)}}\\[3pt]
\toprule
\relictabletitle{5}{Metric dictionary}
\textbf{Metric} & \textbf{Numerator} & \textbf{Denominator} & \textbf{Better} & \textbf{Applicability} \\
\midrule
\endhead
\midrule
\multicolumn{5}{@{}r@{}}{\normalfont\scriptsize Continued on next page.}\\
\endfoot
\bottomrule
\endlastfoot
\multicolumn{5}{l}{\emph{Verified production (evaluator, mainline)}} \\
Evaluator-confirmed seeded issues & issues whose contract failed on starter and passes on candidate & seeded issues in pack & high & all packs \\
Exposed cases on mainline & exposed units passing & exposed units & high & packs separating exposed from held-out \\
Held-out cases on mainline & held-out units passing & held-out units & high & packs with a held-out set \\
Complete contracts on mainline & hidden contracts fully passing & hidden contracts & high & all packs \\
\midrule
\multicolumn{5}{l}{\emph{Workspace and delivery}} \\
Workspace contracts & contracts passing on the workspace tree & hidden contracts & high & packs with a workspace series \\
Workspace exposed cases & exposed units passing on workspace tree & exposed units & high & idem \\
Patch acceptance & accepted patches & generated patches & high & all runs \\
Accepted work reaching mainline & accepted patches present in a merged commit & accepted patches & high & all runs with patches \\
PR merge rate & merged pull requests & opened pull requests & high & runs that opened a PR \\

Releases & release objects published & --- (count) & n/a & all runs \\
\midrule
\multicolumn{5}{l}{\emph{Organizational mechanism}} \\
Currently adopted protocols & registry lineages with terminal \code{adoption\_status=adopted} & --- (count) & n/a & all runs \\
Ever-adopted protocol lineages & distinct run--protocol lineages with an adoption record, including later obsolete rules & --- (count) & n/a & runs with lifecycle records \\
Terminal protocol-registry objects & all registry objects, with adopted, proposed, and obsolete states counted separately & --- (count) & n/a & runs with a terminal registry \\
Mechanism uses & recorded \code{use} events, evidence level labelled & --- (count) & n/a & all runs \\
Runtime protocol enforcements & exported protocol-linked enforcement events; scope stated with the count & --- (count) & n/a & runs with the reported counter \\
Distinct affected targets & distinct governed objects in enforcement events & --- (count) & n/a & bundles with references \\
\midrule
\multicolumn{5}{l}{\emph{Efficiency}} \\
Tokens per task & total tokens summed over tasks & tasks & low & all runs \\
Tokens per confirmed fix & block-level token total & block-level confirmed-fix count & low & positive-fix model--workload blocks; macro averaged \\
Actions per outcome & total actions & confirmed fixes & low & idem \\
\midrule
\multicolumn{5}{l}{\emph{Diagnostic}} \\*
Seeded completion (declared) & tasks the board marks complete & seeded issues & --- & all packs \\*
Completion credibility & declared issues also evaluator-confirmed & declared-complete issues & high & common eligible issue set; positive declarations \\
\end{longtable}
\arrayrulecolor{black}

\endgroup

\subsubsection{Verified Production Metrics}
\label{app:production_metrics}

The four mainline production rates in Table~\ref{tab:metric_dictionary}
are computed from frozen evaluator records. Confirmed seeded issues pair
a baseline-failing contract with a passing candidate status. Exposed and
held-out rates use their own requirement sets and scoring units. A complete
contract requires all constituent behavioral cases to pass, with partial
credit retained at case level.

Each numerator is computed on the evaluated mainline artifact and paired
with the corresponding frozen denominator. This procedure ties the
reported production endpoints to both the starter behavior and the final
integrated code.

\subsubsection{Workspace and Delivery Metrics}
\label{app:delivery_metrics}

Workspace metrics score local candidate trees; mainline metrics score the
integrated shared repository. Delivery accounting follows generated patches,
accepted patches, accepted patches contained in merged commits, opened
pull requests, and merged pull requests. Repository objects supply these
stages, while explicit action occurrences remain separate log counters.

Workspace/mainline comparisons align run identity, checkpoint or endpoint,
and scoring units. Latency measures use recorded event or checkpoint times.
The selected case additionally joins PR identities to their opening,
verification, blocking, and merge records.

\subsubsection{Organizational-Mechanism Metrics}
\label{app:mechanism_metrics}

Protocol measures include proposal, adoption, registration, activation,
action matching, reading, use, enforcement, amendment, and retirement.
Each event retains its protocol identity and time. The generic \code{use}
event credits a successful action matching the protocol's
\code{affected\_actions}; decision and execution effects have their own
linked records.

The census identifies autonomous, seeded, and imported rules by origin.
Revisions are grouped into the parent within-run lineage. Persistence is
measured from a named lifecycle origin, such as adoption or activation,
to the last qualifying use or recorded active step. Formation applies the
unit, step, use-count, and 48-step criterion in
Appendix~\ref{app:formation_criterion}; the strong evidence grade joins
third-party execution, governed-object changes, and evaluator outcomes.

\subsubsection{Efficiency Metrics}
\label{app:efficiency_metrics}

Mean tokens per run averages provider-token totals over contributing runs.
For cost per confirmed issue, tokens and confirmed fixes are pooled across
the contributing seeds within each model--workload block. The block cost
is its token total divided by its confirmed-fix total. Zero-success runs
retain their expenditure within the block, and blocks with positive
confirmed-fix counts supply the eligible block ratios.

Arm estimates macro-average their eligible block ratios: B0--B3 use 18,
21, 21, and 22 blocks, respectively. The paired B3--B2 contrast computes
both conditions on the same 20 common blocks before aggregation.
Table~\ref{tab:token_costs} retains the Terra+Opus auxiliary calculation
on its 14 common blocks. Internal-transfer costs use the separate transfer
analysis associated with Table~\ref{tab:transfer-current}.

\subsubsection{Diagnostic Metrics}
\label{app:diagnostic_metrics}

Completion credibility is the fraction of declared-complete issues confirmed by the evaluator. 

\subsection{Aggregation, Coverage, and Missingness}
\label{app:aggregation_coverage}

\subsubsection{Metric-Specific Pooling}
\label{app:metric_pooling}

For each run $r$ and metric $k$, the analysis records numerator $n_{rk}$,
denominator $d_{rk}$, applicability, and evaluation status. It first
computes the eligible run-level rate $n_{rk}/d_{rk}$, averages the three
seeds within each model--workload--condition block, and then equally
weights applicable model--workload blocks across the three-model study.
The Terra+Opus auxiliary analyses use the same block-macro procedure on
their two-model population.

The numerator and denominator travel together with the metric's scoring
unit and contributing-run set. Measured zero outcomes retain their zero
value; inapplicable or undefined rates retain their NA status. Paired
contrasts apply the same metric-eligibility and workload/version mapping
to both conditions before recomputing the aggregate.

\subsubsection{Checkpoint and Endpoint Coverage}
\label{app:checkpoint_endpoint_coverage}

The reported checkpoint curves and main endpoints cover the 360-run three-model study. Curves use the 24-step grid, with tree type and metric-specific scoring units fixed for each series. 

\subsubsection{Missing and Inapplicable Metrics}
\label{app:missing_metrics}

The analysis retains metric applicability, denominator validity, evaluator
status, run completion, and version eligibility as separate fields.
\NAcell{} denotes an inapplicable or undefined rate, including held-out
metrics for workloads without a held-out set and outcome-normalized ratios
with a zero denominator. A measured failure contributes zero to the
corresponding pass numerator while retaining its valid denominator.

Infrastructure and malformed-evaluation errors retain explicit error
statuses. The recorded eligibility rule determines which metric rows enter
the paired calculation, and both conditions use the same rule. Run and
version records preserve the reason associated with an excluded or
superseded observation.

\subsubsection{Exact Count Records}
\label{app:exact_counts}

Each reported metric is associated with its raw numerator, denominator,
scoring unit, contributing-run count, and analysis-population identifiers.
The aggregation records retain the workload and model weights used to
produce the displayed rates. These fields provide the numerical inputs
for recomputing the table entries and matching an aggregate back to its
constituent run and evaluator records.

\subsection{Estimands}
\label{app:estimands}

\subsubsection{Primary Organization Contrast}
\label{app:primary_estimand}

The primary contrast is B3--B2 within matched model--workload--seed units for verified production and resource outcomes. The design contains 30 matched units per model, or 90 across the three models.

\subsubsection{Secondary Organization Contrasts}
\label{app:secondary_estimands}

\Btwo{} minus \Bone{} estimates the SDL-based configuration bundle,
including selection mode, profile conditioning, and member learning
(Appendix~\ref{app:b2_definition}); \Bone{} minus \Bzero{} estimates
having long-horizon multi-agent collaboration under the documented condition
definitions. All three models cover the same ten-workload, three-seed,
four-condition design. Both are reported as secondary ladder diagnostics. 

\subsubsection{Transfer Estimands}
\label{app:transfer_estimands_stats}

Exec--Text uses 30 matched GPT-5.6 Terra workload--seed units. Exec--Fresh uses the corresponding B2 main-study runs. Seed rates are averaged within workloads, then applicable workloads receive equal weight. Held-out outcomes use W02--W10, or 27 observations per treatment.

\subsubsection{Heterogeneity Estimands}
\label{app:heterogeneity_estimands}

Heterogeneity is summarized by workload, workload family, seed, and model. Model-stratified endpoints cover all three models. Workload/family and leave-one-workload-out summaries use Terra+Opus, with run-level rates macro-averaged over the corresponding blocks.

\subsection{Statistical Analysis and Supplementary Checks}
\label{app:statistical_plan}

\subsubsection{Matched Effects and Clustering}
\label{app:matched_effects}

The paired main-study unit is model--workload--seed. Bootstrap resampling
holds the model--workload blocks fixed and samples seeds within each block.
For a treatment contrast, corresponding seeds are resampled as pairs, and
each draw recomputes the same block-macro estimate used by the endpoint
table. Arm intervals recompute their respective arm estimates.

This procedure preserves the shared starter, requirement set, and evaluator
associated with each workload while retaining the paired treatment
structure. Population-specific resampling settings are given below.

\subsubsection{Confidence Intervals and Bootstrap}
\label{app:confidence_intervals}

The production endpoints in Table~\ref{tab:primary-results} use 50,000
bootstrap draws. Each arm interval recomputes that arm's block-macro
estimate; the B3--B2 interval resamples matched seeds within each fixed
model--workload block.

The three-model outcome-normalized-cost analysis uses 20,000 draws. Arm
intervals use the respective eligible blocks, and the cost contrast
resamples matched seeds within the 20 common B2--B3 blocks. Terra+Opus
workload/family, leave-one-workload-out, and budget-capped analyses use
10,000 draws with random seed 1729.

Internal-transfer intervals come from the transfer analysis associated
with Table~\ref{tab:transfer-current}. The in-situ complete-contract
interval uses 200,000 matched-seed draws within its ten workloads;
Table~\ref{tab:insitu_summary} reports paired 95\% intervals for all four
in-situ endpoints.

\subsubsection{Multiplicity}
\label{app:multiplicity}

The complete-contract B3--B2 contrast is primary; secondary endpoints are reported with pointwise 95\% intervals.

\section{Full Results and Robustness Checks}
\label{app:full_results}

\subsection{Run Inventory}
\label{app:run_inventory}

\subsubsection{Main-Study Coverage}
\label{app:pilot_run_inventory}

The main study contains 120 runs per model and 30 runs per condition within each model. Table~\ref{tab:run_matrices} summarizes the experiment inventory.

\subsubsection{Transfer Runs}
\label{app:transfer_run_inventory}

Table~\ref{tab:transfer_inventory} summarizes the transfer treatments and reused Fresh reference.

\begingroup
\centering\footnotesize
\def\RelicCurrentTable{tab:transfer_inventory}
\setlength{\LTcapwidth}{\linewidth}
\arrayrulecolor{RuleGray}
\begin{xltabular}{\linewidth}{P{2.1cm} P{2.2cm} L}
\caption{Internal-transfer treatment accounting.}\label{tab:transfer_inventory}\\
\toprule
\relictabletitle{3}{Internal-transfer treatment accounting}
\textbf{Condition} & \textbf{Run accounting} & \textbf{State and origin} \\
\midrule
\endfirsthead
\multicolumn{3}{@{}l@{}}{\normalfont\footnotesize \textbf{Table~\thetable{} (continued)}}\\[3pt]
\toprule
\relictabletitle{3}{Internal-transfer treatment accounting}
\textbf{Condition} & \textbf{Run accounting} & \textbf{State and origin} \\
\midrule
\endhead
\midrule
\multicolumn{3}{@{}r@{}}{\normalfont\scriptsize Continued on next page.}\\
\endfoot
\bottomrule
\endlastfoot
Text & 30 new runs & Ten workloads $\times$ three seeds; fresh members and readable protocol content. \\*
Exec & 30 new runs & The same ten-workload, three-seed design; fresh members and content-matched executable package. \\*
\midrule
Fresh reference & 30 reused runs & GPT-5.6 Terra B2 main-study reference. \\
\end{xltabular}
\arrayrulecolor{black}

\endgroup

\subsubsection{Case-Diagnostic Runs}
\label{app:case_run_inventory}

The mechanism case uses W01, GPT-5.6 Terra, seed 1401, and the 336-step horizon. Appendix~\ref{app:case_commit_bound} gives the protocol and delivery timeline.

\subsection{Workload, Seed, and Family Outcomes}
\label{app:per_run_results}

\subsubsection{Per-Workload Outcomes}
\label{app:pack_outcomes}
Table~\ref{tab:pack_outcomes} reports Terra+Opus workload-level B3--B2 effects using the block-macro aggregation in Appendix~\ref{app:metric_pooling}.
\begingroup
\centering
\begingroup
\footnotesize
\arrayrulecolor{RuleGray}
\setlength{\tabcolsep}{2.7pt}
\renewcommand{\arraystretch}{1.23}
\def\RelicCurrentTable{tab:pack_outcomes}
\setlength{\LTcapwidth}{\linewidth}
\begin{xltabular}{\linewidth}{@{}L *{3}{C{0.250\linewidth}}@{}}
\caption{\textbf{Per-workload mainline B3--B2 point estimates.}}\label{tab:pack_outcomes}\\
\toprule
\relictabletitle{4}{Per-workload verified production: B3--B2}
\textbf{Workload} & \makecell{\textbf{Complete}\\\textbf{contracts}} & \makecell{\textbf{Confirmed}\\\textbf{seeded issues}} & \makecell{\textbf{Exposed}\\\textbf{cases}} \\
\midrule
\endfirsthead
\multicolumn{4}{@{}l@{}}{\normalfont\footnotesize \textbf{Table~\thetable{} (continued)}}\\[3pt]
\toprule
\relictabletitle{4}{Per-workload verified production: B3--B2}
\textbf{Workload} & \makecell{\textbf{Complete}\\\textbf{contracts}} & \makecell{\textbf{Confirmed}\\\textbf{seeded issues}} & \makecell{\textbf{Exposed}\\\textbf{cases}} \\
\midrule
\endhead
\midrule
\multicolumn{4}{@{}r@{}}{\normalfont\scriptsize Continued on next page.}\\
\endfoot
\bottomrule
\endlastfoot
W01 & $-6.67$ pp & $-6.67$ pp & $-6.19$ pp \\
W02 & $+5.56$ pp & $+4.76$ pp & $+4.76$ pp \\
W03 & $+14.81$ pp & $+19.05$ pp & $+16.67$ pp \\
W04 & $+16.67$ pp & $+16.67$ pp & $+16.67$ pp \\
W05 & $0.00$ pp & $0.00$ pp & $0.00$ pp \\
W06 & $+6.25$ pp & $+8.33$ pp & $+8.33$ pp \\
W07 & $+13.54$ pp & $+18.06$ pp & $+18.06$ pp \\
W08 & $+7.69$ pp & $+9.26$ pp & $+9.26$ pp \\*
W09 & $+1.04$ pp & $+1.39$ pp & $+1.39$ pp \\*
W10 & $+6.25$ pp & $+8.33$ pp & $+8.33$ pp \\
\end{xltabular}

\endgroup
\arrayrulecolor{black}
\endgroup

\RelicNeedTable{tab:family_outcomes}{3}
\subsubsection{Repair and Construction Split}
\label{app:family_outcomes}
\begingroup
\centering
\begingroup
\footnotesize
\arrayrulecolor{RuleGray}
\setlength{\tabcolsep}{2.8pt}
\renewcommand{\arraystretch}{1.23}
\def\RelicCurrentTable{tab:family_outcomes}
\setlength{\LTcapwidth}{\linewidth}
\begin{xltabular}{\linewidth}{@{}L *{4}{C{0.102\linewidth}} C{0.142\linewidth}@{}}
\caption{\textbf{Construction and repository-based workloads reported separately.}}\label{tab:family_outcomes}\\
\toprule
\relictabletitle{6}{Verified production by workload family}
\textbf{Metric} & \armhead{B0Dark}{B0} & \armhead{B1Dark}{B1} & \armhead{B2Dark}{B2} & \armhead{B3Dark}{B3} & \textcolor{B3Dark}{\textbf{B3--B2}} \\
\midrule
\endfirsthead
\multicolumn{6}{@{}l@{}}{\normalfont\footnotesize \textbf{Table~\thetable{} (continued)}}\\[3pt]
\toprule
\relictabletitle{6}{Verified production by workload family}
\textbf{Metric} & \armhead{B0Dark}{B0} & \armhead{B1Dark}{B1} & \armhead{B2Dark}{B2} & \armhead{B3Dark}{B3} & \textcolor{B3Dark}{\textbf{B3--B2}} \\
\midrule
\endhead
\midrule
\multicolumn{6}{@{}r@{}}{\normalfont\scriptsize Continued on next page.}\\
\endfoot
\bottomrule
\endlastfoot
\relictablepanel{6}{Construction}
Complete contracts on mainline & 6.92\% & 8.10\% & 9.25\% & \bestcell{15.32\%} & \reliceffectcell{+6.07 pp} \\
Evaluator-confirmed seeded issues & 8.37\% & 7.70\% & 10.30\% & \bestcell{17.06\%} & \reliceffectcell{+6.76 pp} \\*
\relictablepanel{6}{Repair}
Complete contracts on mainline & 13.69\% & 28.32\% & 23.37\% & \bestcell{30.32\%} & \reliceffectcell{+6.95 pp} \\*
Evaluator-confirmed seeded issues & 18.33\% & 38.24\% & 31.39\% & \bestcell{40.46\%} & \reliceffectcell{+9.07 pp} \\
\end{xltabular}
\RelicTableNotes{Repository-based workloads W06--W10 include bug fixes and feature development.}

\endgroup
\arrayrulecolor{black}
\endgroup

\subsection{Full Endpoint and Time-Series Results}
\label{app:all_endpoint_results}

\subsubsection{Checkpoint Curves}
\label{app:checkpoint_curves}

Figure~\ref{fig:verified-production-outcomes} pools checkpoint trajectories and endpoints across the three-model, 360-run study. Series are evaluated on the 24-step grid and rendered as step-held curves between checkpoints.

\subsubsection{Workspace and Mainline Outcomes}
\label{app:workspace_mainline_results}

Figure~\ref{fig:verified-production-outcomes} presents workspace and mainline outcomes. Appendix~\ref{app:case_work_delivery} traces repository delivery through seven 48-step windows in the mechanism case.

\subsubsection{Declared Completion and Evaluator Confirmation}
\label{app:declaration_calibration}

Figure~\ref{fig:verified-production-outcomes}a compares declared completion and evaluator confirmation across all three models. Dashed and solid curves use the same checkpoint grid and eligible issue set.

For one eligible run/checkpoint let $D$ be the set of seeded issues declared
complete, $E$ the set confirmed by the evaluator, and $I$ the common eligible
issue set. The two marginal rates are $|D|/|I|$ and $|E|/|I|$. The object-level summary reports declaration confirmation $|D\cap E|/|D|$, unconfirmed declarations
$|D\setminus E|/|D|$, and confirmed-but-undeclared issues $|E\setminus D|$.

\begingroup
\centering
\begingroup
\footnotesize
\arrayrulecolor{RuleGray}
\setlength{\tabcolsep}{3.5pt}
\renewcommand{\arraystretch}{1.23}
\def\RelicCurrentTable{tab:matched_delivery_diagnostics}
\setlength{\LTcapwidth}{\linewidth}
\begin{xltabular}{\linewidth}{@{}L *{4}{C{0.118\linewidth}}@{}}
\caption{\textbf{Object-matched declaration diagnostics.}}\label{tab:matched_delivery_diagnostics}\\
\toprule
\relictabletitle{5}{Matched declaration diagnostics}
\textbf{Matched diagnostic} & \armhead{B0Dark}{B0} & \armhead{B1Dark}{B1} & \armhead{B2Dark}{B2} & \armhead{B3Dark}{B3} \\
\midrule
\endfirsthead
\multicolumn{5}{@{}l@{}}{\normalfont\footnotesize \textbf{Table~\thetable{} (continued)}}\\[3pt]
\toprule
\relictabletitle{5}{Matched declaration diagnostics}
\textbf{Matched diagnostic} & \armhead{B0Dark}{B0} & \armhead{B1Dark}{B1} & \armhead{B2Dark}{B2} & \armhead{B3Dark}{B3} \\
\midrule
\endhead
\midrule
\multicolumn{5}{@{}r@{}}{\normalfont\scriptsize Continued on next page.}\\
\endfoot
\bottomrule
\endlastfoot
Declared issues confirmed, $|D\cap E|/|D|$ & \bestcell{56.68\%} & 56.41\% & 54.01\% & 48.37\% \\*
Declared issues unconfirmed, $|D\setminus E|/|D|$ & \bestcell{43.32\%} & 43.59\% & 45.99\% & 51.63\% \\
\end{xltabular}
\RelicTableNotes{Terra+Opus endpoint declaration records.}

\endgroup
\arrayrulecolor{black}
\endgroup

\subsection{Action Allocation and Delivery Conversion}
\label{app:action_delivery_results}

\subsubsection{Action Totals and Category Coverage}
\label{app:action_family_results}
Table~\ref{tab:action_families} reports mapped and unmapped action totals for the Terra+Opus runs. Action shares use mapped actions within each condition.
\begingroup
\centering
\begingroup
\footnotesize
\arrayrulecolor{RuleGray}
\setlength{\tabcolsep}{3.5pt}
\renewcommand{\arraystretch}{1.23}
\def\RelicCurrentTable{tab:action_families}
\setlength{\LTcapwidth}{\linewidth}
\begin{xltabular}{\linewidth}{@{}L *{4}{C{0.125\linewidth}}@{}}
\caption{\textbf{Terra+Opus action-log totals.}}\label{tab:action_families}\\
\toprule
\relictabletitle{5}{Terra+Opus action-log accounting}
\textbf{Action accounting} & \armhead{B0Dark}{B0} & \armhead{B1Dark}{B1} & \armhead{B2Dark}{B2} & \armhead{B3Dark}{B3} \\
\midrule
\endfirsthead
\multicolumn{5}{@{}l@{}}{\normalfont\footnotesize \textbf{Table~\thetable{} (continued)}}\\[3pt]
\toprule
\relictabletitle{5}{Terra+Opus action-log accounting}
\textbf{Action accounting} & \armhead{B0Dark}{B0} & \armhead{B1Dark}{B1} & \armhead{B2Dark}{B2} & \armhead{B3Dark}{B3} \\
\midrule
\endhead
\midrule
\multicolumn{5}{@{}r@{}}{\normalfont\scriptsize Continued on next page.}\\
\endfoot
\bottomrule
\endlastfoot
Mapped actions & 9,459 & 86,288 & 82,111 & 88,289 \\*
Unmapped actions & 0 & 1 & 0 & 0 \\
\end{xltabular}

\endgroup
\arrayrulecolor{black}
\endgroup

\RelicNeedTable{tab:delivery_funnel}{3}
\subsubsection{Delivery Funnel}
\label{app:delivery_funnel}
\label{app:delivery_funnel_results}

\begingroup
\centering
\begingroup
\footnotesize
\arrayrulecolor{RuleGray}
\setlength{\tabcolsep}{3.5pt}
\renewcommand{\arraystretch}{1.23}
\def\RelicCurrentTable{tab:delivery_funnel}
\setlength{\LTcapwidth}{\linewidth}
\begin{xltabular}{\linewidth}{@{}L *{4}{C{0.125\linewidth}}@{}}
\caption{\textbf{Raw delivery-object counts across 60 runs per condition.}}\label{tab:delivery_funnel}\\
\toprule
\relictabletitle{5}{Delivery funnel: repository-object counts}
\textbf{Stage (60-run raw totals)} & \armhead{B0Dark}{B0} & \armhead{B1Dark}{B1} & \armhead{B2Dark}{B2} & \armhead{B3Dark}{B3} \\
\midrule
\endfirsthead
\multicolumn{5}{@{}l@{}}{\normalfont\footnotesize \textbf{Table~\thetable{} (continued)}}\\[3pt]
\toprule
\relictabletitle{5}{Delivery funnel: repository-object counts}
\textbf{Stage (60-run raw totals)} & \armhead{B0Dark}{B0} & \armhead{B1Dark}{B1} & \armhead{B2Dark}{B2} & \armhead{B3Dark}{B3} \\
\midrule
\endhead
\midrule
\multicolumn{5}{@{}r@{}}{\normalfont\scriptsize Continued on next page.}\\
\endfoot
\bottomrule
\endlastfoot
Generated patches & 1,868 & 9,048 & 8,610 & 8,079 \\
Accepted patches & 1,508 & 6,684 & 6,473 & 6,381 \\
Accepted patches reaching mainline & 510 & 1,435 & 1,971 & \bestcell{2,234} \\
Opened PRs & 498 & 1,695 & 2,361 & 2,908 \\
Merged PRs & 352 & 1,168 & 1,743 & \bestcell{2,090} \\*

\end{xltabular}
\RelicTableNotes{Delivery rates use model--workload macro averaging.}

\endgroup
\arrayrulecolor{black}
\endgroup

\subsubsection{No-Visible-Progress and Rework Diagnostics}
\label{app:progress_diagnostics}

No-visible-progress actions record effort or status without touching an artifact, patch, branch, review, or message. The action taxonomy assigns \code{work\_on\_task}, \code{rest}, \code{defer}, \code{overtime}, and \code{idle} to this category. 

\subsection{Organizational-Mechanism Inventory}
\label{app:mechanism_inventory}

\subsubsection{Proposals and Adoption}
\label{app:proposal_adoption_results}

Table~\ref{tab:mechanism_inventory} summarizes 617 terminal protocol objects across the 60 Terra+Opus B3 runs: 513 adopted, 92 proposed, and 12 obsolete. Adopted objects comprise 393 autonomous and 120 environment-seeded rules.
The 12 obsolete lineages each retain a nonempty \code{adoption\_tick},
which joins their earlier adoption to the terminal lifecycle state.
Runtime \code{protocol\_enforcements} counts accumulate over the
recorded trajectory, including events preceding retirement.

\begingroup
\centering
\begingroup
\footnotesize
\arrayrulecolor{RuleGray}
\setlength{\tabcolsep}{3.5pt}
\renewcommand{\arraystretch}{1.23}
\def\RelicCurrentTable{tab:mechanism_inventory}
\setlength{\LTcapwidth}{\linewidth}
\begin{xltabular}{\linewidth}{@{}L C{0.185\linewidth} C{0.230\linewidth}@{}}
\caption{\textbf{B3 protocol-state inventory and recorded runtime activity.}}\label{tab:mechanism_inventory}\\
\toprule
\relictabletitle{3}{B3 protocol inventory and runtime activity}
\endfirsthead
\multicolumn{3}{@{}l@{}}{\normalfont\footnotesize \textbf{Table~\thetable{} (continued)}}\\[3pt]
\toprule
\relictabletitle{3}{B3 protocol inventory and runtime activity}
\endhead
\midrule
\multicolumn{3}{@{}r@{}}{\normalfont\scriptsize Continued on next page.}\\
\endfoot
\bottomrule
\endlastfoot
\relictablepanel{3}{A. Terminal registry status across the Terra+Opus 60 B3 runs}
\textbf{Terminal status} & \textbf{Autonomous} & \makecell{\textbf{Environment-}\\\textbf{seeded}} \\*
\midrule
Currently adopted & 393 & 120 \\
Proposed, not adopted & 92 & 0 \\
Obsolete after adoption & 12 & 0 \\
\rowcolor{PanelFill}
\textbf{Terminal objects by origin} & \textbf{497} & \textbf{120} \\
\midrule
\multicolumn{2}{@{}l@{}}{All terminal protocol-registry objects} & \textbf{617} \\
\multicolumn{2}{@{}l@{}}{All currently adopted protocols} & \textbf{513} \\
\multicolumn{2}{@{}l@{}}{All ever-adopted lineages, including later obsolete rules} & \textbf{525} \\
\midrule
\relictablepanel{3}{B. Recorded runtime activity by workload}
\textbf{Workload} & \textbf{Recorded uses} & \makecell{\textbf{Runtime}\\\textbf{enforcements}} \\*
\midrule
W01 & 6,450 & 4,199 \\
W02 & 4,453 & 1,270 \\
W03 & 4,848 & 4,297 \\
W04 & 4,648 & 582 \\
W05 & 4,231 & 3,997 \\
W06 & 4,155 & 2,452 \\
W07 & 4,235 & 4,373 \\
W08 & 6,087 & 5,144 \\
W09 & 5,230 & 1,970 \\
W10 & 5,253 & 5,159 \\*
\midrule
\rowcolor{PanelFill}
\textbf{All B3 runs (all protocols)} & \textbf{49,590} & \textbf{33,443} \\*
B0/B1/B2, each condition & 0 & 0 \\
\end{xltabular}
\RelicTableNotes{The environment-seeded rules are \code{proto\_review\_before\_merge} and \code{proto\_experiment\_logging}, each present in all 60 Terra+Opus B3 runs. Runtime activity totals include both autonomous and seeded rules.}

\endgroup
\arrayrulecolor{black}
\endgroup

\subsubsection{Use and Enforcement}
\label{app:use_enforcement_results}

Table~\ref{tab:mechanism_inventory} reports protocol-use and enforcement events. The mechanism case links 211 enforcements to 12 pull requests (Appendix~\ref{app:distinct_prs}). 

\subsubsection{Amendment and Lifecycle}
\label{app:amendment_results}

Of the 280 formed lineages, 182 have revisions, totaling 1,231 revision entries and 1,231 folded or superseded proposal identifiers. Retirement is recorded as \code{obsolete}; repair proposals use \code{policy\_repair\_proposal}. 

\subsubsection{Capability Families}
\label{app:capability_families}

Recurring protocol themes include commit-bound verification, interface-to-contract checking, review assignment, experiment logging, release readiness, and handoff readiness. 

\subsubsection{Corpus, Counting Units, and Formation Coverage}
\label{app:capability_census}

The autonomous-protocol census covers the 60 Terra+Opus B3 runs. Autonomous lineages exclude \code{proto\_experiment\_logging} and \code{proto\_review\_before\_merge}.

The census counts within-run protocol lineages identified by \code{(run, protocol\_id)}, with revisions grouped into their parent lineage. Formation uses the criterion in Appendix~\ref{app:formation_criterion}; the strong grade adds object-state and evaluator-linked evidence.

\paragraph{Coverage.}
The census links terminal protocol states, structured rule content,
formation records, family labels, roles, and revisions across all 60
Terra+Opus B3 runs. Each formed registry ID resolves to its rich
\code{ProtocolSpec}; the \code{(run, protocol\_id)} key joins its
proposal, adoption, use, and revision records while grouping versions into
one lineage.

\begingroup
\centering
\begingroup
\footnotesize
\arrayrulecolor{RuleGray}
\setlength{\tabcolsep}{3pt}
\renewcommand{\arraystretch}{1.20}
\def\RelicCurrentTable{tab:capability_census}
\setlength{\LTcapwidth}{\linewidth}
\begin{xltabular}{\linewidth}{@{}>{\hsize=1.15\hsize\linewidth=\hsize\raggedright\arraybackslash}X C{0.18\linewidth} >{\hsize=.85\hsize\linewidth=\hsize\raggedright\arraybackslash}X@{}}
\caption{\textbf{Autonomous-protocol census over the Terra+Opus B3 population.}}\label{tab:capability_census}\\
\toprule
\relictabletitle{3}{Autonomous formation: Terra+Opus 60-run census}
\textbf{Stage or coverage} & \makecell{\textbf{All 60}\\\textbf{B3 runs}} & \textbf{Interpretation} \\
\midrule
\endfirsthead
\multicolumn{3}{@{}l@{}}{\normalfont\footnotesize \textbf{Table~\thetable{} (continued)}}\\[3pt]
\toprule
\relictabletitle{3}{Autonomous formation: Terra+Opus 60-run census}
\textbf{Stage or coverage} & \makecell{\textbf{All 60}\\\textbf{B3 runs}} & \textbf{Interpretation} \\
\midrule
\endhead
\midrule
\multicolumn{3}{@{}r@{}}{\normalfont\scriptsize Continued on next page.}\\
\endfoot
\bottomrule
\endlastfoot
Autonomous proposal lineages & 497 & Excludes both seeded rules. \\
Currently adopted lineages & 393 & Terminal \code{adoption\_status=adopted}. \\
Unadopted proposal lineages & 92 & Terminal \code{adoption\_status=proposed}. \\
Subsequently obsolete lineages & 12 & Terminal \code{obsolete}; all have an adoption step. \\
Ever-adopted lineages & 405 & 393 currently adopted plus 12 subsequently obsolete. \\
Weak-or-strong formed & 280 & Exact formation count. \\
\quad weak-only & 248 & Strong is excluded from this row. \\*
\quad strong subset & 32 & Object-level and evaluator-linked evidence grade. \\*
\end{xltabular}

\endgroup
\arrayrulecolor{black}
\endgroup

Across the complete 60-run census, 393 of 497 autonomous proposal lineages
remain adopted at the endpoint (79.1\%). Another 12 were adopted and later
became obsolete, giving 405 ever-adopted lineages (81.5\% of proposals);
the remaining 92 are unadopted proposals. The 280 weak-or-strong formed
lineages constitute 69.1\% of the ever-adopted set and 56.3\% of all autonomous
proposal lineages. Of the formed lineages, 248 are weak-only (88.6\%) and
32 strong (11.4\%). Every formed registry ID maps to its rich
\code{ProtocolSpec}, allowing versions to remain in one lineage. 

\subsubsection{Semantic Clustering and Functional Categories}
\label{app:capability_clustering}

We distinguish raw naming from functional grouping. The 280 formed lineages
use 243 distinct raw \code{protocol\_type} strings. These strings capture free naming and task-specific wording. All 280 map to a rich spec with an existing, runtime-recorded
\code{family}. The present grouping uses these recorded labels, with content
inspection organized around triggers, concrete obligations, roles, affected
objects, and intended effects.  Eight of the ten registered
families occur; \code{docs} and \code{other} do not occur as primary family labels.

\begingroup
\centering
\begingroup
\footnotesize
\arrayrulecolor{RuleGray}
\setlength{\tabcolsep}{2.0pt}
\renewcommand{\arraystretch}{1.23}
\def\RelicCurrentTable{tab:capability_types}
\setlength{\LTcapwidth}{\linewidth}
\begin{xltabular}{\linewidth}{@{}L C{0.080\linewidth} C{0.070\linewidth} C{0.060\linewidth} C{0.060\linewidth} C{0.085\linewidth} C{0.090\linewidth} C{0.065\linewidth}@{}}
\caption{\textbf{Recorded functional families of formed autonomous protocols.}}\label{tab:capability_types}\\
\toprule
\relictabletitle{8}{Formed capabilities by functional family}
\textbf{Recorded family} & \makecell{\textbf{Formed}\\\textbf{lineages}} & \textbf{Share} & \makecell{\textbf{Runs}\\\textbf{/60}} & \makecell{\textbf{Tasks}\\\textbf{/10}} & \makecell{\textbf{GPT-5.6}\\\textbf{Terra}} & \makecell{\textbf{Claude}\\\textbf{Opus 4.6}} & \textbf{Strong} \\
\midrule
\endfirsthead
\multicolumn{8}{@{}l@{}}{\normalfont\footnotesize \textbf{Table~\thetable{} (continued)}}\\[3pt]
\toprule
\relictabletitle{8}{Formed capabilities by functional family}
\textbf{Recorded family} & \makecell{\textbf{Formed}\\\textbf{lineages}} & \textbf{Share} & \makecell{\textbf{Runs}\\\textbf{/60}} & \makecell{\textbf{Tasks}\\\textbf{/10}} & \makecell{\textbf{GPT-5.6}\\\textbf{Terra}} & \makecell{\textbf{Claude}\\\textbf{Opus 4.6}} & \textbf{Strong} \\
\midrule
\endhead
\midrule
\multicolumn{8}{@{}r@{}}{\normalfont\scriptsize Continued on next page.}\\
\endfoot
\bottomrule
\endlastfoot
Review and merge & 108 & 38.6\% & 48 & 10 & 38 & 70 & 14 \\
Release engineering & 80 & 28.6\% & 46 & 10 & 39 & 41 & 8 \\
Evidence governance & 56 & 20.0\% & 38 & 10 & 26 & 30 & 6 \\
Debugging & 17 & 6.1\% & 12 & 9 & 3 & 14 & 4 \\
Customer & 10 & 3.6\% & 7 & 4 & 0 & 10 & 0 \\
Ownership & 7 & 2.5\% & 6 & 6 & 2 & 5 & 0 \\
Experiment & 1 & 0.4\% & 1 & 1 & 0 & 1 & 0 \\*
Budget governance & 1 & 0.4\% & 1 & 1 & 0 & 1 & 0 \\*
\midrule
\rowcolor{PanelFill}
\textbf{Total} & \textbf{280} & \textbf{100\%} & --- & \textbf{10} & \textbf{108} & \textbf{172} & \textbf{32} \\
\end{xltabular}
\RelicTableNotes{Family counts use one primary label per lineage. Run and workload coverage may overlap across families.}

\endgroup
\arrayrulecolor{black}
\endgroup

\paragraph{Review and merge.}
The largest family typically responds to an opened or updated PR, or a claim
that work is merge-ready or release-ready. Obligations include mapping issue
acceptance criteria to tests, rerunning applicable CI against current mainline,
recording a non-author review decision, and refusing readiness claims or merges
when evidence, ownership, CI, or review is incomplete. The 108 lineages cover all ten workloads. 

\paragraph{Release engineering.}
These 80 lineages center on release candidates and release gates. A recurring
sequence assigns an owner, locates the failing module, applies a patch, runs
targeted and smoke/readiness tests, obtains reviewer sign-off, and reruns the
release gate before publishing or declaring readiness. This family and
review/merge both use CI and evidence, but differ in their principal object:
a release candidate versus a PR and mainline promotion.

\paragraph{Evidence governance.}
The 56 lineages regulate claims about fixes, compatibility, resolution,
experiment results, or release status. They require applicable acceptance
behavior, preserved compatibility, current verification status, and traceable
evidence; insufficiently verified work remains pending, blocked, or under
review rather than resolved or shipped. Coverage across all ten workloads shows recurrence of this protocol theme.

\paragraph{Debugging and recovery.}
Seventeen lineages join failure signatures, suspected modules, patches,
targeted tests, smoke results, and review sign-off into a closure chain. Four have strong evidence, a larger within-family fraction than the overall strong fraction.

\paragraph{Customer, ownership, experiment, and budget.}
The ten customer lineages, all in the Claude Opus 4.6 records, connect
customer issues to implementation or PRs, triage, owners, and response dates.
Seven ownership lineages specify responsibility and handoff for tasks or
artifacts. One experiment lineage requires shared tracker entries with run
content, reproducibility state, seed/configuration, raw trace, and cost.
The single budget-governance lineage targets \emph{institutional cost}: when a protocol blocks too many requests in a week, its benefit and time cost must be reviewed and revision or retirement may be proposed.

\paragraph{Structural completeness, revisions, and role templates.}
All 280 formed lineages have nonempty triggers, responsibility maps, success
metrics, and enforcement-action fields. Problem evidence is present in 264
(94.3\%), required steps in 274 (97.9\%), and required fields in 270 (96.4\%).
These fields describe the structured content of formed protocols.

\begingroup
\centering
\begingroup
\footnotesize
\arrayrulecolor{RuleGray}
\setlength{\tabcolsep}{3.5pt}
\renewcommand{\arraystretch}{1.23}
\def\RelicCurrentTable{tab:capability_structure}
\setlength{\LTcapwidth}{\linewidth}
\begin{xltabular}{\linewidth}{@{}L C{0.195\linewidth} C{0.165\linewidth}@{}}
\caption{\textbf{Field completeness of the 280 formed rich protocol specifications.}}\label{tab:capability_structure}\\
\toprule
\relictabletitle{3}{Structured-field coverage of formed protocols}
\textbf{Structured field} & \textbf{Nonempty /280} & \textbf{Coverage} \\
\midrule
\endfirsthead
\multicolumn{3}{@{}l@{}}{\normalfont\footnotesize \textbf{Table~\thetable{} (continued)}}\\[3pt]
\toprule
\relictabletitle{3}{Structured-field coverage of formed protocols}
\textbf{Structured field} & \textbf{Nonempty /280} & \textbf{Coverage} \\
\midrule
\endhead
\midrule
\multicolumn{3}{@{}r@{}}{\normalfont\scriptsize Continued on next page.}\\
\endfoot
\bottomrule
\endlastfoot
Trigger condition & 280 & 100\% \\
Responsible roles & 280 & 100\% \\
Success metric & 280 & 100\% \\
Enforcement action & 280 & 100\% \\
Problem evidence & 264 & 94.3\% \\*
Required steps & 274 & 97.9\% \\*
Required fields & 270 & 96.4\% \\
\end{xltabular}

\endgroup
\arrayrulecolor{black}
\endgroup

Executor-role assignments are founder 179, cofounder 27, reliability 18,
editorial 16, fast engineer 13, artifact design 12, community 8, and external
voice 7. Every spec's reviewer and approver fields contain the same
\code{cofounder + founder} pair. The generation template fixes this reviewer/approver pair; executor-role assignments vary by protocol.

\subsubsection{Diversity, Concentration, and Cross-Workload Reuse}
\label{app:capability_diversity}

The three largest families account for 244/280 formed lineages (87.1\%). With
family shares $p_k$, the observed Herfindahl index is
$\sum_k p_k^2=0.276$. Shannon entropy is
$H=-\sum_k p_k\log p_k=1.469$ nats, giving $H/\log 8=0.706$ across the eight
observed families and an effective family count $\exp(H)=4.34$.
The distribution has a long tail but is concentrated around review,
release, and evidence obligations. These statistics summarize lineage-weighted diversity over the recorded functional families.

\begingroup
\centering
\begingroup
\footnotesize
\arrayrulecolor{RuleGray}
\setlength{\tabcolsep}{3.5pt}
\renewcommand{\arraystretch}{1.23}
\def\RelicCurrentTable{tab:capability_model_strata}
\setlength{\LTcapwidth}{\linewidth}
\begin{xltabular}{\linewidth}{@{}L C{0.185\linewidth} C{0.185\linewidth}@{}}
\caption{\textbf{Model strata of the Terra+Opus B3 protocol census.}}\label{tab:capability_model_strata}\\
\toprule
\relictabletitle{3}{Model strata of the capability census}
\textbf{Complete model stratum} & \makecell{\textbf{GPT-5.6}\\\textbf{Terra}} & \makecell{\textbf{Claude}\\\textbf{Opus 4.6}} \\
\midrule
\endfirsthead
\multicolumn{3}{@{}l@{}}{\normalfont\footnotesize \textbf{Table~\thetable{} (continued)}}\\[3pt]
\toprule
\relictabletitle{3}{Model strata of the capability census}
\textbf{Complete model stratum} & \makecell{\textbf{GPT-5.6}\\\textbf{Terra}} & \makecell{\textbf{Claude}\\\textbf{Opus 4.6}} \\
\midrule
\endhead
\midrule
\multicolumn{3}{@{}r@{}}{\normalfont\scriptsize Continued on next page.}\\
\endfoot
\bottomrule
\endlastfoot
B3 runs with content and evidence & 30 & 30 \\
Formed lineages & 108 & 172 \\
Strong lineages & 14 & 18 \\
Mean formed per run & 3.60 & 5.73 \\
\end{xltabular}

\endgroup
\arrayrulecolor{black}
\endgroup

The GPT-5.6 Terra stratum contains 108 formed lineages across five recorded
families, while the Claude Opus 4.6 stratum contains 172 across eight. These
strata characterize how the observed capability census is distributed across
the two model settings.

Review/merge, release engineering, and evidence governance each occur across
10/10 workloads; debugging occurs across nine, ownership six, customer four,
and experiment and budget one each. Within a recurring family, triggers,
fields, and affected artifacts remain workload-specific.

Episode-window reconstruction shows post-adoption use beyond the originating episode window for all 280 formed lineages.  This complements the cross-workload family recurrence and the independent-organizational-unit criterion used in the formation classifier.

\subsubsection{Capability Coverage}
\label{app:capability_blindspots}

The formed-protocol census is concentrated in review/merge, release
engineering, and evidence governance, with smaller families covering debugging,
customer handling, ownership, experiment tracking, and institutional cost.
A complementary formation-funnel view follows visible friction through
recognition, proposal, adoption, and sustained use, while ordinary repairs
remain separate from newly formed organizational mechanisms.

\subsection{Model-Conditioned Friction and Institutional Responses}
\label{app:model_friction_response}

We examine recorded work episodes and episode-to-protocol links under GPT-5.6 Terra, Claude Opus 4.6, and DeepSeek-V4-Flash, abbreviated as Terra, Opus 4.6, and DeepSeek.

\paragraph{Overlapping recorded pressures.}
All three strata encounter debugging, launch pressure, experimentation,
feedback, and claim disputes (Table~\ref{tab:model_friction_episodes}).
Typical debugging and launch-pressure counts are close, although other
categories have different spreads. One Opus 4.6 PDF-workload run records
4,009 experiment episodes; retaining it raises that stratum's experiment
mean to 209.86, while its median is 26.5. We therefore describe typical
activity using medians and interquartile ranges. 

\begingroup
\centering
\footnotesize
\arrayrulecolor{RuleGray}
\setlength{\tabcolsep}{3pt}
\renewcommand{\arraystretch}{1.15}
\def\RelicCurrentTable{tab:model_friction_episodes}
\setlength{\LTcapwidth}{\linewidth}
\begin{xltabular}{\linewidth}{@{}L *{3}{C{0.195\linewidth}}@{}}
\caption{\textbf{Recorded friction and pressure episodes.}
Entries are the reported median [first quartile, third quartile] per run.}
\label{tab:model_friction_episodes}\\
\toprule
\textbf{Episode category} & \textbf{Terra} & \textbf{Opus 4.6} & \textbf{DeepSeek} \\
\midrule
\endfirsthead
\multicolumn{4}{@{}l@{}}{\normalfont\footnotesize \textbf{Table~\thetable{} (continued)}}\\[3pt]
\toprule
\textbf{Episode category} & \textbf{Terra} & \textbf{Opus 4.6} & \textbf{DeepSeek} \\
\midrule
\endhead
\midrule
\multicolumn{4}{@{}r@{}}{\normalfont\scriptsize Continued on next page.}\\
\endfoot
\bottomrule
\endlastfoot
Debugging & \relicestci{33.0}{27.5, 45.8} & \relicestci{31.5}{26.0, 39.8} & \relicestci{29.5}{24.0, 39.3} \\
Launch pressure & \relicestci{34.0}{33.0, 36.8} & \relicestci{35.0}{33.0, 37.8} & \relicestci{33.0}{32.0, 35.0} \\
Experiment activity & \relicestci{27.0}{24.3, 39.0} & \relicestci{26.5}{22.3, 33.0} & \relicestci{21.0}{18.0, 25.0} \\
Feedback ingestion & \relicestci{12.0}{10.3, 73.0} & \relicestci{13.0}{11.0, 37.5} & \relicestci{13.0}{11.0, 14.8} \\
Claim disputes & \relicestci{1.0}{0.0, 1.0} & \relicestci{1.0}{1.0, 1.0} & \relicestci{1.0}{0.0, 1.0} \\
\end{xltabular}

\endgroup
\arrayrulecolor{black}

\paragraph{Episode-to-protocol links.}
Formed protocol specifications link source-episode categories to protocol families through \code{source\_episode\_ids}.
Table~\ref{tab:model_friction_links} reports distinct linked lineages for
selected source--family combinations. For example, Terra contributes 24
launch-pressure-to-release-engineering links; Opus 4.6 contributes 17
debugging-to-release-engineering links and 13 launch-pressure-to-review/merge
links. DeepSeek contributes seven debugging-to-release-engineering links and
five experiment-to-review/merge links. These records connect concrete work
episodes to protocol formation. 

\begingroup
\centering
\footnotesize
\arrayrulecolor{RuleGray}
\setlength{\tabcolsep}{3pt}
\renewcommand{\arraystretch}{1.12}
\def\RelicCurrentTable{tab:model_friction_links}
\setlength{\LTcapwidth}{\linewidth}
\begin{xltabular}{\linewidth}{@{}L L C{0.19\linewidth}@{}}
\caption{\textbf{Recorded links from source episodes to formed protocols.}
Counts pool the three analysis strata.}
\label{tab:model_friction_links}\\
\toprule
\textbf{Source episode} & \textbf{Protocol family} & \makecell{\textbf{Distinct linked}\\\textbf{lineages}} \\
\midrule
\endfirsthead
\multicolumn{3}{@{}l@{}}{\normalfont\footnotesize \textbf{Table~\thetable{} (continued)}}\\[3pt]
\toprule
\textbf{Source episode} & \textbf{Protocol family} & \makecell{\textbf{Distinct linked}\\\textbf{lineages}} \\
\midrule
\endhead
\midrule
\multicolumn{3}{@{}r@{}}{\normalfont\scriptsize Continued on next page.}\\
\endfoot
\bottomrule
\endlastfoot
Launch pressure & Release engineering & 39 \\
Debugging & Release engineering & 34 \\
Launch pressure & Review / merge & 25 \\
Experiment & Review / merge & 18 \\
Feedback ingestion & Review / merge & 14 \\
Debugging & Review / merge & 11 \\
Feedback ingestion & Evidence governance & 5 \\
Experiment & Evidence governance & 4 \\
Launch pressure & Evidence governance & 4 \\
Claim dispute & Evidence governance & 3 \\
Debugging & Debugging & 3 \\
\end{xltabular}
\RelicTableNotes{A lineage may link to multiple source-episode categories.}
\endgroup
\arrayrulecolor{black}

\paragraph{Friction volume and institutional response.}
Workload--seed-block-centered correlations vary by episode category. Debugging with debugging/release institutions
has $r=0.20$ (95\% CI [$-$0.07, 0.46]); claim disputes with evidence
governance, $r=0.19$ [$-$0.15, 0.47]; and launch pressure with review/release,
$r=0.20$ [$-$0.06, 0.46]. Experiment activity with experiment/evidence
institutions is positively associated ($r=0.24$ [0.08, 0.44]), while feedback
with customer/ownership institutions is negatively associated
($r=-0.24$ [$-$0.33, $-$0.14]). 

The episode links connect recorded work experiences to the protocol-formation history.

\RelicNeedTable{tab:token_costs}{5}
\subsection{Cost and Efficiency Results}
\label{app:cost_results}

\subsubsection{240-Run Token Accounting}
\label{app:token_breakdown}
\begingroup
\centering
\begingroup
\footnotesize
\arrayrulecolor{RuleGray}
\setlength{\tabcolsep}{2.0pt}
\renewcommand{\arraystretch}{1.24}
\def\RelicCurrentTable{tab:token_costs}
\setlength{\LTcapwidth}{\linewidth}
\begin{xltabular}{\linewidth}{@{}L *{4}{C{0.125\linewidth}} C{0.17\linewidth}@{}}
\caption{\textbf{Terra+Opus token accounting over 240 runs.}}\label{tab:token_costs}\\
\toprule
\relictabletitle{6}{Main-study token expenditure and outcome-normalized cost}
\textbf{Token metric} & \armhead{B0Dark}{B0} & \armhead{B1Dark}{B1} & \armhead{B2Dark}{B2} & \armhead{B3Dark}{B3} & \makecell{\textcolor{B3Dark}{\textbf{B3--B2}}\\\textbf{paired}} \\
\midrule
\endfirsthead
\multicolumn{6}{@{}l@{}}{\normalfont\footnotesize \textbf{Table~\thetable{} (continued)}}\\[3pt]
\toprule
\relictabletitle{6}{Main-study token expenditure and outcome-normalized cost}
\textbf{Token metric} & \armhead{B0Dark}{B0} & \armhead{B1Dark}{B1} & \armhead{B2Dark}{B2} & \armhead{B3Dark}{B3} & \makecell{\textcolor{B3Dark}{\textbf{B3--B2}}\\\textbf{paired}} \\
\midrule
\endhead
\midrule
\multicolumn{6}{@{}r@{}}{\normalfont\scriptsize Continued on next page.}\\
\endfoot
\bottomrule
\endlastfoot
Terminal total (M) & \bestcell{211.309} & 2,634.198 & 225.058 & 391.372 & -- \\*
Mean / run (M) & \relicbestci{3.522}{3.350,3.695} & \relicestci{43.903}{40.745,47.262} & \relicestci{3.751}{3.439,4.054} & \relicestci{6.523}{5.119,8.379} & \reliceffectci{+2.772}{1.373,4.665} \\*
Per confirmed issue (M) & \relicestci{4.947}{3.396,5.963} & \relicestci{31.760}{25.994,46.592} & \relicbestci{2.907}{2.402,4.632} & \relicestci{3.651}{2.359,6.660} & \reliceffectci{$-$0.174}{$-$1.686,0.633} \\
\end{xltabular}
\RelicTableNotes{Brackets give 95\% intervals. Mean-token/run uses all 60 runs per condition. Per-confirmed-issue arm estimates average each arm's eligible nonzero model--workload blocks; the paired B3--B2 contrast is recomputed on 14 common nonzero blocks (42 paired runs), so it need not equal the arithmetic difference of the two arm estimates.}

\endgroup
\arrayrulecolor{black}
\endgroup

\subsubsection{Outcome-Normalized Cost}
\label{app:normalized_cost}

Table~\ref{tab:token_costs} reports Terra+Opus cost estimates. Table~\ref{tab:primary-results} reports the pooled three-model estimates. Both use the block-macro cost definition in Appendix~\ref{app:efficiency_metrics}.

\subsubsection{Budget-Capped Checkpoint Sensitivity}
\label{app:budget_capped_sensitivity}

For each $(\text{model},\text{workload},\text{seed})$ pair, the
budget-capped sensitivity takes B2's final token expenditure as the cap
and selects the last saved B3 checkpoint at or below that cap. The score
is the evaluator output at the selected checkpoint. All 60 Terra+Opus
matched units have a qualifying checkpoint with an evaluator output.

Token receipts provide recoverable monotone checkpoint curves for the 60
B3 runs, and terminal totals match the corresponding frozen token records.
The analysis joins each checkpoint's recorded cumulative spend, candidate
state, and evaluator output. It equally weights model--workload blocks
and uses 10,000 paired-seed bootstrap draws with seed 1729.

\begingroup
\centering
\begingroup
\footnotesize
\arrayrulecolor{RuleGray}
\setlength{\tabcolsep}{2.25pt}
\renewcommand{\arraystretch}{1.26}
\def\RelicCurrentTable{tab:budget_capped_sensitivity}
\setlength{\LTcapwidth}{\linewidth}
\begin{xltabular}{\linewidth}{@{}L c c C{0.115\linewidth} C{0.135\linewidth} C{0.210\linewidth}@{}}
\caption{\textbf{Budget-capped checkpoint sensitivity.}}\label{tab:budget_capped_sensitivity}\\
\toprule
\relictabletitle{6}{Budget-capped checkpoint sensitivity}
\textbf{Endpoint} & \textbf{Pairs} & \textbf{Blocks} & \armhead{B2Dark}{B2} & \makecell{\armhead{B3Dark}{B3}\\\textbf{under cap}} & \makecell{\textcolor{B3Dark}{\textbf{B3--B2}}\\\textbf{[95\% CI]}} \\
\midrule
\endfirsthead
\multicolumn{6}{@{}l@{}}{\normalfont\footnotesize \textbf{Table~\thetable{} (continued)}}\\[3pt]
\toprule
\relictabletitle{6}{Budget-capped checkpoint sensitivity}
\textbf{Endpoint} & \textbf{Pairs} & \textbf{Blocks} & \armhead{B2Dark}{B2} & \makecell{\armhead{B3Dark}{B3}\\\textbf{under cap}} & \makecell{\textcolor{B3Dark}{\textbf{B3--B2}}\\\textbf{[95\% CI]}} \\
\midrule
\endhead
\midrule
\multicolumn{6}{@{}r@{}}{\normalfont\scriptsize Continued on next page.}\\
\endfoot
\bottomrule
\endlastfoot
Complete contracts & 60 & 20 & 16.310\% & \bestcell{23.760\%} & \reliceffectci{+7.450 pp}{3.491, 10.927} \\
Held-out cases & 54 & 18 & 13.43\% & \bestcell{18.51\%} & \reliceffectci{+5.08 pp}{0.926, 11.111} \\*
Exposed cases & 60 & 20 & 22.570\% & \bestcell{34.290\%} & \reliceffectci{+11.720 pp}{6.781, 19.617} \\*
Evaluator-confirmed seeded issues & 60 & 20 & 20.840\% & \bestcell{27.052\%} & \reliceffectci{+6.212 pp}{4.528, 11.719} \\
\end{xltabular}
\RelicTableNotes{Held-out outcomes use W02--W10; all other rows use W01--W10.}

\endgroup
\arrayrulecolor{black}
\endgroup

All four verified production contrasts remain positive at the selected budget-capped B3 checkpoints.

\RelicNeedTable{tab:transfer_full}{5}
\subsection{Complete Transfer Results}
\label{app:all_transfer_results}

\subsubsection{All Intervention Arms}
\label{app:all_transfer_arms_results}

\begingroup
\centering
\begingroup
\footnotesize
\arrayrulecolor{RuleGray}
\setlength{\tabcolsep}{3pt}
\renewcommand{\arraystretch}{1.20}
\def\RelicCurrentTable{tab:transfer_full}
\setlength{\LTcapwidth}{\linewidth}
\begin{xltabular}{\linewidth}{@{}L *{3}{C{0.177\linewidth}}@{}}
\caption{\textbf{Complete internal-transfer summary.}}\label{tab:transfer_full}\\
\toprule
\relictabletitle{4}{Internal protocol transfer: complete outcomes and costs}
\textbf{Metric} & \armhead{ScratchDark}{Fresh} & \armhead{TextDark}{Text} & \armhead{ExecDark}{Exec} \\
\midrule
\endfirsthead
\multicolumn{4}{@{}l@{}}{\normalfont\footnotesize \textbf{Table~\thetable{} (continued)}}\\[3pt]
\toprule
\relictabletitle{4}{Internal protocol transfer: complete outcomes and costs}
\textbf{Metric} & \armhead{ScratchDark}{Fresh} & \armhead{TextDark}{Text} & \armhead{ExecDark}{Exec} \\
\midrule
\endhead
\midrule
\multicolumn{4}{@{}r@{}}{\normalfont\scriptsize Continued on next page.}\\
\endfoot
\bottomrule
\endlastfoot
Behavioral-case pass rate & 25.4\% & 34.6\% & \bestcell{41.2\%} \\
Exposed-case pass rate & 23.434\% & 39.7\% & \bestcell{44.4\%} \\
Held-out-case pass rate & 11.111\% & 11.1\% & \bestcell{25.9\%} \\
Complete-contract rate & 18.574\% & 31.5\% & \bestcell{32.3\%} \\

Evaluator-confirmed issue rate & 21.720\% & 41.4\% & \bestcell{45.1\%} \\
\midrule
Pull requests opened, mean / run & 47.1 & 52.1 & 48.8 \\
Pull requests merged, mean / run & 38.2 & 42.2 & 39.7 \\
Releases, mean / run & 24.1 & 26.7 & 26.2 \\
Patches generated / accepted, mean / run & 141.8 / 110.0 & 124.7 / 122.4 & 118.0 / 116.7 \\
File-edit actions, mean / run & 141.9 & 121.0 & 118.0 \\
Total actions, mean / run & 1,389.3 & 1,377.8 & 1,385.9 \\
\midrule
Imported mechanisms, mean / run & 0 & 6.0 (text) & 6.0 (executable) \\
New target mechanisms, mean / run & 0 & 0 & 0 \\
Recorded uses, mean / run & 0 & 0 & 588.0 \\
Recorded binding enforcements, mean / run & 0 & 0 & 197.0 \\
Amendments, mean / run & 0 & 0 & 0 \\
\midrule
Mean tokens / run (M) & \bestcell{2.381} & 2.615 & 2.448 \\*
Tokens / evaluated case (k) & 176 & 154 & \bestcell{144} \\*
Tokens / verified pass (k) & 691 & 444 & \bestcell{350} \\
\end{xltabular}
\RelicTableNotes{Count-valued rows are means per run. Confidence intervals for behavioral-case pass rate and cost metrics appear in Table~\ref{tab:transfer-current}.}

\endgroup
\arrayrulecolor{black}
\endgroup

\subsubsection{Primary Controlled Contrasts}
\label{app:transfer_contrast_results}

Exec improves behavioral-case pass rate by 6.5 pp [0.7, 15.6] over Text and 15.8 pp [9.2, 31.8] over Fresh. Exec also exceeds Text on each production endpoint in Table~\ref{tab:transfer_full}.

\subsection{Robustness Checks}
\label{app:robustness}

\subsubsection{Equal-Task and Matched Subsets}
\label{app:equal_task_subset}

Terra+Opus auxiliary analyses use ten workloads, three seeds, and B0--B3. Complete, exposed, and confirmed-issue contrasts have 60 pairs; held-out contrasts have 54 pairs over W02--W10.

\RelicNeedTable{tab:model_stratified_effects}{3}
\subsubsection{Model-Stratified B3--B2 Effects}
\label{app:model_stratified_effects}

\begingroup
\centering
\begingroup
\footnotesize
\arrayrulecolor{RuleGray}
\setlength{\tabcolsep}{3.5pt}
\renewcommand{\arraystretch}{1.23}
\def\RelicCurrentTable{tab:model_stratified_effects}
\setlength{\LTcapwidth}{\linewidth}
\begin{xltabular}{\linewidth}{@{}L C{0.120\linewidth} C{0.120\linewidth} C{0.260\linewidth} C{0.115\linewidth}@{}}
\caption{\textbf{Model-stratified B3--B2 effects.}}\label{tab:model_stratified_effects}\\
\toprule
\relictabletitle{5}{Model-stratified B3--B2 effects}
\textbf{Endpoint} & \armhead{B2Dark}{B2} & \armhead{B3Dark}{B3} & \makecell{\textcolor{B3Dark}{\textbf{B3--B2}}\\\textbf{[95\% CI]}} & \makecell{\textbf{Applicable}\\\textbf{pairs}} \\
\midrule
\endfirsthead
\multicolumn{5}{@{}l@{}}{\normalfont\footnotesize \textbf{Table~\thetable{} (continued)}}\\[3pt]
\toprule
\relictabletitle{5}{Model-stratified B3--B2 effects}
\textbf{Endpoint} & \armhead{B2Dark}{B2} & \armhead{B3Dark}{B3} & \makecell{\textcolor{B3Dark}{\textbf{B3--B2}}\\\textbf{[95\% CI]}} & \makecell{\textbf{Applicable}\\\textbf{pairs}} \\
\midrule
\endhead
\midrule
\multicolumn{5}{@{}r@{}}{\normalfont\scriptsize Continued on next page.}\\
\endfoot
\bottomrule
\endlastfoot
\relictablepanel{5}{GPT-5.6 Terra}
Complete contracts & 18.574\% & \bestcell{23.440\%} & \reliceffectci{+4.866 pp}{$-1.713$, 11.359} & 30/30 \\
Held-out cases & 11.111\% & \bestcell{12.037\%} & \reliceffectci{+0.926 pp}{$-4.630$, 7.407} & 27/30 \\
Exposed cases & 23.434\% & \bestcell{28.082\%} & \reliceffectci{+4.648 pp}{$-0.011$, 12.505} & 30/30 \\
Confirmed seeded issues & 21.720\% & \bestcell{28.082\%} & \reliceffectci{+6.362 pp}{1.296, 14.220} & 30/30 \\
\relictablepanel{5}{Claude Opus 4.6}
Complete contracts & 14.046\% & \bestcell{22.200\%} & \reliceffectci{+8.154 pp}{4.509, 13.271} & 30/30 \\
Held-out cases & 15.741\% & \bestcell{35.185\%} & \reliceffectci{+19.444 pp}{0.167, 49.333} & 27/30 \\*
Exposed cases & 21.706\% & \bestcell{32.518\%} & \reliceffectci{+10.812 pp}{3.911, 17.613} & 30/30 \\*
Confirmed seeded issues & 19.960\% & \bestcell{29.438\%} & \reliceffectci{+9.478 pp}{5.113, 18.018} & 30/30 \\
\relictablepanel{5}{DeepSeek-V4-Flash}
Complete contracts & 9.54\% & \bestcell{13.64\%} & \reliceffectci{+4.10 pp}{1.31, 7.13} & 30/30 \\
Held-out cases & 1.85\% & \bestcell{4.63\%} & \reliceffectci{+2.78 pp}{0.93, 5.56} & 27/30 \\
Exposed cases & 12.79\% & \bestcell{17.08\%} & \reliceffectci{+4.29 pp}{0.77, 7.92} & 30/30 \\
Confirmed seeded issues & 12.05\% & \bestcell{17.08\%} & \reliceffectci{+5.03 pp}{1.73, 8.45} & 30/30 \\
\end{xltabular}
\RelicTableNotes{Each model has 30 matched pairs for complete, exposed, and confirmed-issue outcomes and 27 for held-out outcomes. DeepSeek intervals use 50,000 same-workload, matched-seed bootstrap draws.}

\endgroup
\arrayrulecolor{black}
\endgroup

All four B3--B2 point estimates are positive for GPT-5.6 Terra, Claude
Opus 4.6, and DeepSeek-V4-Flash, with no endpoint direction reversal across
models. For DeepSeek-V4-Flash, the paired 95\% intervals for all four verified endpoints exclude zero. For held-out cases specifically, B0/B1 are 0.00\% [0.00, 0.00], B2 is 1.85\% [0.00, 3.70], and B3 is 4.63\% [0.00, 8.33], with B3--B2 +2.78 pp [0.93, 5.56].
Its additional behavioral diagnostics are positive on mainline (B0/B1/B2/B3:
4.80 [2.78, 7.06] / 11.50 [9.42, 13.59] / 10.28 [7.09, 13.49] / 13.64
[11.55, 15.90]\%; B3--B2 +3.35 pp [0.28, 6.59]) and on workspace (9.88
[6.98, 12.46] / 17.54 [15.20, 20.11] / 15.21 [11.50, 18.63] / 17.57
[14.76, 20.55]\%; +2.37 pp [$-$2.13, 7.15]). Under the same block-macro outcome-normalized cost definition used for the
main-study paired estimand, the DeepSeek B3--B2 effect is $-$1.087M tokens per
confirmed issue. 

\RelicNeedTable{tab:leave_one_out}{3}
\subsubsection{Leave-One-Workload-Out Analysis}
\label{app:leave_one_out}

\begingroup
\centering
\begingroup
\footnotesize
\arrayrulecolor{RuleGray}
\setlength{\tabcolsep}{3pt}
\renewcommand{\arraystretch}{1.23}
\def\RelicCurrentTable{tab:leave_one_out}
\setlength{\LTcapwidth}{\linewidth}
\begin{xltabular}{\linewidth}{@{}L C{0.258\linewidth} C{0.258\linewidth} C{0.142\linewidth}@{}}
\caption{\textbf{Leave-one-workload-out robustness.}}\label{tab:leave_one_out}\\
\toprule
\relictabletitle{4}{Leave-one-workload-out robustness}
\textbf{Endpoint} & \makecell{\textcolor{B3Dark}{\textbf{Full B3--B2}}\\\textbf{[95\% CI]}} & \textbf{LOO $>0$} & \makecell{\textbf{LOO CI}\\\textbf{$>0$}} \\
\midrule
\endfirsthead
\multicolumn{4}{@{}l@{}}{\normalfont\footnotesize \textbf{Table~\thetable{} (continued)}}\\[3pt]
\toprule
\relictabletitle{4}{Leave-one-workload-out robustness}
\endhead
\midrule
\multicolumn{4}{@{}r@{}}{\normalfont\scriptsize Continued on next page.}\\
\endfoot
\bottomrule
\endlastfoot
Complete contracts & \reliceffectci{+6.51 pp}{2.812, 10.517} & 10/10 & 10/10 \\*
Exposed cases & \reliceffectci{+7.73 pp}{2.183, 13.016} & 10/10 & 8/10 \\*
Confirmed seeded issues & \reliceffectci{+7.92 pp}{2.908, 14.409} & 10/10 & 10/10 \\
\midrule
\multicolumn{4}{@{}l@{}}{%
\begin{minipage}[t]{\linewidth}
\begin{tabularx}{\linewidth}{@{}L C{0.258\linewidth} C{0.258\linewidth} C{0.142\linewidth}@{}}
\relictablepanel{4}{Sensitivity range across workload omissions}
\rowcolor{white}
\textbf{Endpoint} & \textbf{Minimum LOO} & \textbf{Maximum LOO} & \makecell{\textbf{Max. abs.}\\\textbf{deviation}} \\*
\midrule
\rowcolor{white}
Complete contracts & \makecell{+5.39 pp\\{\scriptsize omit W04}} & \makecell{+7.98 pp\\{\scriptsize omit W01}} & 1.46 pp \\*
\rowcolor{white}
Exposed cases & \makecell{+6.58 pp\\{\scriptsize omit W07}} & \makecell{+9.27 pp\\{\scriptsize omit W01}} & 1.55 pp \\*
\rowcolor{white}
Confirmed seeded issues & \makecell{+6.68 pp\\{\scriptsize omit W03}} & \makecell{+9.54 pp\\{\scriptsize omit W01}} & 1.62 pp \\
\end{tabularx}
\end{minipage}%
}\\
\end{xltabular}
\RelicTableNotes{Each refit omits one workload from both models and paired conditions. ``LOO CI $>0$'' counts intervals wholly above zero.}

\endgroup
\arrayrulecolor{black}
\endgroup

The three reported endpoints remain positive under every W01--W10 omission, so the
aggregate direction is not attributable to a single workload. Complete-contract
effects range from +5.39 to +7.98 percentage points, exposed effects from +6.58
to +9.27 points, and confirmed-issue effects from +6.68 to +9.54 points. 

\subsection{In-Situ Prose-Only Institutionalization Ablation}
\label{app:insitu_binding_ablation}

\subsubsection{Design and Intervention}
\label{app:insitu_binding_design}

The internal Text/Exec experiment fixes a transferred package and disables
new target-stage formation. Here we test executable binding inside the original
organizational setting while proposal generation, governance, approval,
adoption, and revision remain enabled. The completed ablation uses Claude Opus 4.6,
all ten workloads W01--W10, and three seeds (1401, 2711, and 4013), yielding
30 new B3-text runs matched to the corresponding 30 executable B3 runs from
the main study by workload and seed. 

B3-text retains the B3 roster, member-local learning, reflection, wishes,
protocol proposals, governance, adoption, revision, the core SDL scoring
parameterization, and ordinary task and repository validity checks. The
intervention changes what adoption produces operationally: an adopted
\code{ProtocolSpec} remains a readable shared organizational rule and is rendered
through the text-only decision-context interface, but it does not install an
executable protocol binding. Learned-rule effects on selection and routing,
automatic protocol-use recording, and protocol-specific enforcement are disabled.
Environment-level CI, review, and transaction-validity checks continue to apply.

Both conditions begin at the initial state and independently develop their work, proposals, and adopted rules throughout the run.

\subsubsection{Three-Seed Outcome Comparison}
\label{app:insitu_binding_performance}

The primary in-situ endpoint is complete contracts on mainline. Complete,
exposed, and confirmed-issue outcomes use 30 matched workload--seed units;
held-out outcomes use 27 matched units because W01 has no held-out set. For
each metric, run-level rates are first averaged over the three seeds within
each applicable workload, then equally weighted across applicable workloads:
ten for complete, exposed, and confirmed-issue outcomes, and nine for held-out
outcomes. Table~\ref{tab:insitu_summary} reports the aggregate comparison.

\begingroup
\centering
\begingroup
\footnotesize
\arrayrulecolor{RuleGray}
\setlength{\tabcolsep}{2.8pt}
\renewcommand{\arraystretch}{1.23}
\def\RelicCurrentTable{tab:insitu_summary}
\setlength{\LTcapwidth}{\linewidth}
\begin{xltabular}{\linewidth}{@{}L C{0.155\linewidth} C{0.155\linewidth} C{0.170\linewidth} C{0.205\linewidth}@{}}
\caption{\textbf{Three-seed in-situ B3-text versus executable B3.}}\label{tab:insitu_summary}\\
\toprule
\relictabletitle{5}{In-situ binding ablation: verified mainline outcomes}
\textbf{Endpoint} & \armhead{TextDark}{B3-text} & \armhead{B3Dark}{B3} & \textbf{B3$-$B3-text} & \textbf{Paired 95\% CI} \\
\midrule
\endfirsthead
\multicolumn{5}{@{}l@{}}{\normalfont\footnotesize \textbf{Table~\thetable{} (continued)}}\\[3pt]
\toprule
\relictabletitle{5}{In-situ binding ablation: verified mainline outcomes}
\textbf{Endpoint} & \armhead{TextDark}{B3-text} & \armhead{B3Dark}{B3} & \textbf{B3$-$B3-text} & \textbf{Paired 95\% CI} \\
\midrule
\endhead
\midrule
\multicolumn{5}{@{}r@{}}{\normalfont\scriptsize Continued on next page.}\\
\endfoot
\bottomrule
\endlastfoot
Complete contracts on mainline & 15.03\% & \bestcell{22.20\%} & \reliceffectcell{+7.18 pp} & \reliceffectcell{[+3.54, +10.91] pp} \\
Held-out cases on mainline & 12.96\% & \bestcell{35.19\%} & \reliceffectcell{+22.22 pp} & \reliceffectcell{[+10.85, +29.71] pp} \\
Exposed cases on mainline & 18.47\% & \bestcell{32.52\%} & \reliceffectcell{+14.04 pp} & \reliceffectcell{[+8.99, +20.23] pp} \\
Evaluator-confirmed seeded issues & 17.72\% & \bestcell{29.44\%} & \reliceffectcell{+11.71 pp} & \reliceffectcell{[+6.27, +15.91] pp} \\
\end{xltabular}
\RelicTableNotes{Rates macro-average the three seeds within each workload, then equally weight applicable workloads. Intervals are paired 95\% CIs; held-out outcomes use W02--W10.}
\endgroup
\arrayrulecolor{black}
\endgroup

Executable B3 exceeds B3-text by 7.18 percentage points on complete contracts
(95\% CI [+3.54, +10.91]). The same direction appears on held-out (+22.22 pp),
exposed (+14.04 pp), and evaluator-confirmed seeded-issue (+11.71 pp) endpoints.
All four paired 95\% intervals are above zero.

\subsubsection{Matched Complete-Contract Endpoints}
\label{app:insitu_contract_detail}

Table~\ref{tab:insitu_contracts} reports complete-contract counts for the 30 matched B3-text / B3 pairs.

\begingroup
\centering
\begingroup
\footnotesize
\arrayrulecolor{RuleGray}
\setlength{\tabcolsep}{3.0pt}
\renewcommand{\arraystretch}{1.23}
\def\RelicCurrentTable{tab:insitu_contracts}
\setlength{\LTcapwidth}{\linewidth}
\begin{xltabular}{\linewidth}{@{}L *{3}{C{0.225\linewidth}}@{}}
\caption{\textbf{Matched complete-contract endpoints for B3-text / B3.}}\label{tab:insitu_contracts}\\
\toprule
\relictabletitle{4}{Matched complete-contract endpoints: B3-text / B3}
\textbf{Workload} & \textbf{Seed 1401} & \textbf{Seed 2711} & \textbf{Seed 4013} \\
\midrule
\endfirsthead
\multicolumn{4}{@{}l@{}}{\normalfont\footnotesize \textbf{Table~\thetable{} (continued)}}\\[3pt]
\toprule
\relictabletitle{4}{Matched complete-contract endpoints: B3-text / B3}
\textbf{Workload} & \textbf{Seed 1401} & \textbf{Seed 2711} & \textbf{Seed 4013} \\
\midrule
\endhead
\midrule
\multicolumn{4}{@{}r@{}}{\normalfont\scriptsize Continued on next page.}\\
\endfoot
\bottomrule
\endlastfoot
W01 Mini Blobstore & 0/5 / 0/5 & 0/5 / 0/5 & 0/5 / 0/5 \\
W02 Traffic Watch & 2/9 / 2/9 & 0/9 / 2/9 & 1/9 / 1/9 \\
W03 TG Automation & 1/9 / 5/9 & 1/9 / 1/9 & 1/9 / 3/9 \\
W04 PDF Reformatter & 0/4 / 1/4 & 0/4 / 2/4 & 0/4 / 1/4 \\
W05 FastAPI Dashboard & 0/4 / 0/4 & 0/4 / 0/4 & 0/4 / 0/4 \\
W06 Boltons & 10/16 / 11/16 & 10/16 / 12/16 & 12/16 / 10/16 \\
W07 Celery & 1/16 / 3/16 & 1/16 / 1/16 & 2/16 / 2/16 \\
W08 Soup Sieve & 1/13 / 1/13 & 0/13 / 1/13 & 1/13 / 1/13 \\
W09 cattrs & 4/16 / 5/16 & 2/16 / 4/16 & 4/16 / 2/16 \\
W10 Tenacity & 6/16 / 4/16 & 2/16 / 5/16 & 5/16 / 3/16 \\*
\midrule
\textbf{10-workload macro} & \textbf{17.23\% / 25.42\%} & \textbf{10.49\% / 22.85\%} & \textbf{17.37\% / 18.34\%} \\
\end{xltabular}
\RelicTableNotes{Each cell gives B3-text / executable B3. The three-seed aggregate is 15.03\% / 22.20\%, yielding the +7.18 pp contrast in Table~\ref{tab:insitu_summary}.}
\endgroup
\arrayrulecolor{black}
\endgroup

\subsubsection{Seed-Stratified Endpoints}
\label{app:insitu_seed_endpoints}

Table~\ref{tab:insitu_seed_endpoints} reports seed-stratified endpoint rates.

\begingroup
\centering
\begingroup
\footnotesize
\arrayrulecolor{RuleGray}
\setlength{\tabcolsep}{2.7pt}
\renewcommand{\arraystretch}{1.20}
\def\RelicCurrentTable{tab:insitu_seed_endpoints}
\setlength{\LTcapwidth}{\linewidth}
\begin{xltabular}{\linewidth}{@{}L C{0.110\linewidth} *{3}{C{0.155\linewidth}}@{}}
\caption{\textbf{Seed-stratified endpoint macros for the in-situ ablation.}}\label{tab:insitu_seed_endpoints}\\
\toprule
\relictabletitle{5}{Seed-stratified endpoint macros}
\textbf{Metric} & \textbf{Arm} & \textbf{Seed 1401} & \textbf{Seed 2711} & \textbf{Seed 4013} \\
\midrule
\endfirsthead
\multicolumn{5}{@{}l@{}}{\normalfont\footnotesize \textbf{Table~\thetable{} (continued)}}\\[3pt]
\toprule
\relictabletitle{5}{Seed-stratified endpoint macros}
\textbf{Metric} & \textbf{Arm} & \textbf{Seed 1401} & \textbf{Seed 2711} & \textbf{Seed 4013} \\
\midrule
\endhead
\midrule
\multicolumn{5}{@{}r@{}}{\normalfont\scriptsize Continued on next page.}\\
\endfoot
\bottomrule
\endlastfoot
Complete contracts & B3-text & 17.23\% & 10.49\% & 17.37\% \\*
Complete contracts & B3 & 25.42\% & 22.85\% & 18.34\% \\
Held-out cases & B3-text & 13.89\% & 2.78\% & 22.22\% \\
Exposed cases & B3-text & 20.96\% & 13.10\% & 21.37\% \\
Evaluator-confirmed seeded issues & B3-text & 19.21\% & 13.10\% & 20.87\% \\
\end{xltabular}

\endgroup
\arrayrulecolor{black}
\endgroup

\subsubsection{Binding Manipulation Check}
\label{app:insitu_formation_check}

Table~\ref{tab:insitu_formation} reports proposal, adoption, readable-rule, and executable-binding state across the 30 matched runs.

\begingroup
\centering
\begingroup
\scriptsize
\arrayrulecolor{RuleGray}
\setlength{\tabcolsep}{2.1pt}
\renewcommand{\arraystretch}{1.20}
\def\RelicCurrentTable{tab:insitu_formation}
\setlength{\LTcapwidth}{\linewidth}
\begin{xltabular}{\linewidth}{@{}L C{0.105\linewidth} C{0.105\linewidth} C{0.105\linewidth} C{0.115\linewidth} C{0.130\linewidth}@{}}
\caption{\textbf{Three-seed B3-text manipulation check.}}\label{tab:insitu_formation}\\
\toprule
\relictabletitle{6}{B3-text manipulation check}
\textbf{Check} & \textbf{Seed 1401} & \textbf{Seed 2711} & \textbf{Seed 4013} & \textbf{B3-text total} & \textbf{B3 total} \\
\midrule
\endfirsthead
\multicolumn{6}{@{}l@{}}{\normalfont\footnotesize \textbf{Table~\thetable{} (continued)}}\\[3pt]
\toprule
\relictabletitle{6}{B3-text manipulation check}
\textbf{Check} & \textbf{Seed 1401} & \textbf{Seed 2711} & \textbf{Seed 4013} & \textbf{B3-text total} & \textbf{B3 total} \\
\midrule
\endhead
\midrule
\multicolumn{6}{@{}r@{}}{\normalfont\scriptsize Continued on next page.}\\
\endfoot
\bottomrule
\endlastfoot
Protocol proposals & 106 & 115 & 103 & \textbf{324} & 347 \\
Adopted protocols & 70 & 81 & 76 & \textbf{227} & 321 \\
Endpoint-renderable adopted rules & 66 & 80 & 75 & \textbf{221} & 249 \\
Cells with readable-rules block & 10/10 & 10/10 & 10/10 & \textbf{30/30} & --- \\
Cells with executable binding enabled & 0/10 & 0/10 & 0/10 & \textbf{0/30} & 30/30 \\
Recorded protocol-use events & 0 & 0 & 0 & \textbf{0} & 25,079 \\
Protocol-specific enforcement events & 0 & 0 & 0 & \textbf{0} & 18,594 \\

\end{xltabular}
\RelicTableNotes{Endpoint inventories and whole-run event totals across the matched 30 runs; seed columns each cover ten workloads.}
\endgroup
\arrayrulecolor{black}
\endgroup

\section{Case-Study Evidence: Commit-Bound Verification}
\label{app:case_commit_bound}

This case traces protocol formation, enforcement, revision, and delivery in the W01 Mini Blobstore run with GPT-5.6 Terra and seed 1401. Displayed member, protocol, and event identifiers use consistent aliases.

\subsection{Case Configuration}
\label{app:case_selection}

\subsubsection{Matched Diagnostic Configuration}
\label{app:case_configuration}

Four runs, one per arm, on Mini Blobstore (\code{W01}) at seed 1401 with a 336-step
horizon and model \code{gpt-5.6-terra} with low reasoning effort.  All four share pack,
starter tree, hidden suite, evaluator, seed, horizon, model, tool surface and
information; circadian rhythm is ablated in all four. The primary endpoint:

\begingroup
\centering
\begingroup
\footnotesize
\arrayrulecolor{RuleGray}
\setlength{\tabcolsep}{3.5pt}
\renewcommand{\arraystretch}{1.23}
\def\RelicCurrentTable{tab:case_endpoint}
\setlength{\LTcapwidth}{\linewidth}
\begin{xltabular}{\linewidth}{@{}L *{4}{C{0.127\linewidth}}@{}}
\caption{\textbf{Matched case diagnostic, endpoint on mainline.}}\label{tab:case_endpoint}\\
\toprule
\relictabletitle{5}{Selected case: verified mainline endpoints}
\textbf{Endpoint} & \armhead{B0Dark}{B0} & \armhead{B1Dark}{B1} & \armhead{B2Dark}{B2} & \armhead{B3Dark}{B3} \\
\midrule
\endfirsthead
\multicolumn{5}{@{}l@{}}{\normalfont\footnotesize \textbf{Table~\thetable{} (continued)}}\\[3pt]
\toprule
\relictabletitle{5}{Selected case: verified mainline endpoints}
\textbf{Endpoint} & \armhead{B0Dark}{B0} & \armhead{B1Dark}{B1} & \armhead{B2Dark}{B2} & \armhead{B3Dark}{B3} \\
\midrule
\endhead
\midrule
\multicolumn{5}{@{}r@{}}{\normalfont\scriptsize Continued on next page.}\\
\endfoot
\bottomrule
\endlastfoot
Behavioural cases passing & 0/35 & 0/35 & 0/35 & \bestcell{28/35} \\*
Complete contracts & 0/5 & 0/5 & 0/5 & \bestcell{2/5} \\
\end{xltabular}

\endgroup
\arrayrulecolor{black}
\endgroup

\subsection{Repository and Action Counts}
\label{app:counter_reconciliation}

\subsubsection{Repository-State Counters}
\label{app:repo_state_counters}

\begingroup
\centering
\begingroup
\footnotesize
\arrayrulecolor{RuleGray}
\setlength{\tabcolsep}{3pt}
\renewcommand{\arraystretch}{1.23}
\def\RelicCurrentTable{tab:case_counters}
\setlength{\LTcapwidth}{\linewidth}
\begin{xltabular}{\linewidth}{@{}P{0.125\linewidth} L *{4}{C{0.094\linewidth}}@{}}
\caption{\textbf{Repository objects and action occurrences in the matched case.}}\label{tab:case_counters}\\
\toprule
\relictabletitle{6}{Selected case: repository and action counters}
\textbf{Source} & \textbf{Counter} & \armhead{B0Dark}{B0} & \armhead{B1Dark}{B1} & \armhead{B2Dark}{B2} & \armhead{B3Dark}{B3} \\
\midrule
\endfirsthead
\multicolumn{6}{@{}l@{}}{\normalfont\footnotesize \textbf{Table~\thetable{} (continued)}}\\[3pt]
\toprule
\relictabletitle{6}{Selected case: repository and action counters}
\textbf{Source} & \textbf{Counter} & \armhead{B0Dark}{B0} & \armhead{B1Dark}{B1} & \armhead{B2Dark}{B2} & \armhead{B3Dark}{B3} \\
\midrule
\endhead
\midrule
\multicolumn{6}{@{}r@{}}{\normalfont\scriptsize Continued on next page.}\\
\endfoot
\bottomrule
\endlastfoot
repo state & code patches accepted & 74 & 168 & 113 & 118 \\
repo state & pull requests opened & 1 & 8 & 16 & 80 \\
repo state & pull requests merged & 0 & 2 & 4 & \bestcell{76} \\
\midrule
action log & \code{edit\_repo\_file} & 75 & 169 & 131 & 112 \\
action log & \code{commit\_patch} & 1 & 12 & 50 & 93 \\
action log & \code{open\_pr} & 0 & 0 & 3 & 20 \\
action log & \code{merge\_pr} & 0 & 2 & 1 & 28 \\*
action log & \code{run\_ci} & 33 & 427 & 142 & 126 \\*
CI records & runs recorded & 73 & 651 & 589 & 339 \\
\end{xltabular}
\RelicTableNotes{Some runtime paths create or merge pull requests without emitting a separate \code{open\_pr} or \code{merge\_pr} action-log entry.}

\endgroup
\arrayrulecolor{black}
\endgroup

\subsection{Work Without Delivery}
\label{app:case_work_delivery}

\subsubsection{Local Production}
\label{app:case_local_production}

The baseline arms continued to perform editing and verification actions. Actions taken per
48-step window are flat to the horizon: at $t\,{\in}\,[0,48)$ the four arms take
23, 206, 194 and 208 actions; at $[96,144)$ they take 22, 198, 184 and 231; at
$[192,240)$ they take 18, 204, 183 and 200; at $[288,336)$ they take 17, 185, 174
and 234. \Bone{} is as busy at $t=330$ as at $t=10$. \Bzero{} is a single member, so
its twenty per window is comparable effort per head.

For local editing, \Bthree{} made 112 \code{edit\_repo\_file}
actions and had 118 code patches accepted, against \Bone{}'s 169 and 168.
\textbf{\Bthree{} has the fewest recorded file-edit actions of the three
multi-member arms}.  The selected B1 mainline endpoint
remains zero despite its larger number of these local operations. 

\subsubsection{Shared Delivery}
\label{app:case_shared_delivery}

Where they differ is handover. \Bzero{} committed once. \Bone{} merged two pull
requests, both inside the first 48 steps, opened its last at $t=96$, and spent the
remaining 240 steps writing and checking without handing anything over: its final 96
steps contain 114 CI runs, 63 edits, 62 internal searches, 51 public-test runs, and
no pull request opened or merged. \Btwo{} opened sixteen and merged four, all four
inside the first 48 steps; its final 96 steps went to \code{work\_on\_task} (114)
and \code{update\_task\_status} (77), which is task bookkeeping around work that
never left the desk. \Bthree{}'s final 96 steps are shaped differently:
\code{work\_on\_task} 50, \code{run\_ci} 36, \code{edit\_repo\_file} 27,
\code{commit\_patch} 26, \code{approve\_proposal} 18. 

\RelicNeedTable{tab:case_delivery_windows}{3}
\subsubsection{Delivery Over Time}
\label{app:case_delivery_time}

\begingroup
\centering
\begingroup
\footnotesize
\arrayrulecolor{RuleGray}
\setlength{\tabcolsep}{3.5pt}
\renewcommand{\arraystretch}{1.23}
\def\RelicCurrentTable{tab:case_delivery_windows}
\setlength{\LTcapwidth}{\linewidth}
\begin{xltabular}{\linewidth}{@{}L *{4}{C{0.145\linewidth}}@{}}
\caption{\textbf{Pull requests opened / merged per 48-step window, from repository state
differenced per window.}}\label{tab:case_delivery_windows}\\
\toprule
\relictabletitle{5}{Selected case: shared delivery over time}
\textbf{Window} & \armhead{B0Dark}{B0} & \armhead{B1Dark}{B1} & \armhead{B2Dark}{B2} & \armhead{B3Dark}{B3} \\
\midrule
\endfirsthead
\multicolumn{5}{@{}l@{}}{\normalfont\footnotesize \textbf{Table~\thetable{} (continued)}}\\[3pt]
\toprule
\relictabletitle{5}{Selected case: shared delivery over time}
\textbf{Window} & \armhead{B0Dark}{B0} & \armhead{B1Dark}{B1} & \armhead{B2Dark}{B2} & \armhead{B3Dark}{B3} \\
\midrule
\endhead
\midrule
\multicolumn{5}{@{}r@{}}{\normalfont\scriptsize Continued on next page.}\\
\endfoot
\bottomrule
\endlastfoot
steps 0--48 & 1 / 0 & 6 / 2 & 12 / 4 & 13 / 9 \\
steps 48--96 & 0 / 0 & 2 / 0 & 3 / 0 & 13 / 12 \\
steps 96--144 & 0 / 0 & 0 / 0 & 1 / 0 & 13 / 14 \\
steps 144--192 & 0 / 0 & 0 / 0 & 0 / 0 & 7 / 9 \\
steps 192--240 & 0 / 0 & 0 / 0 & 0 / 0 & 13 / 9 \\
steps 240--288 & 0 / 0 & 0 / 0 & 0 / 0 & 10 / 15 \\*
steps 288--336 & 0 / 0 & 0 / 0 & 0 / 0 & 11 / 8 \\*
\midrule
\textbf{Total} & 1 / 0 & 8 / 2 & 16 / 4 & \bestcell{80 / 76} \\
\end{xltabular}

\endgroup
\arrayrulecolor{black}
\endgroup

Early \Bthree{} adoptions occur at $t=21$, $39$, $58$, $60$, and $65$.
All recorded baseline merges occur in the first 48-step window; baseline PR
openings extend into later windows, through $t=96$--144 for \Btwo{}.
Adoption timing, PR creation, and merged delivery are shown as distinct series. 

\RelicNeedTable{tab:case_proposals}{5}
\subsection{Capability Inventory and Governance}
\label{app:case_capability_inventory}

\subsubsection{All Proposed Mechanisms}
\label{app:case_all_proposals}

\begingroup
\centering
\begingroup
\footnotesize
\arrayrulecolor{RuleGray}
\setlength{\tabcolsep}{3.5pt}
\renewcommand{\arraystretch}{1.23}
\def\RelicCurrentTable{tab:case_proposals}
\setlength{\LTcapwidth}{\linewidth}
\begin{xltabular}{\linewidth}{@{}L C{0.133\linewidth} C{0.133\linewidth} C{0.10\linewidth} C{0.10\linewidth} C{0.12\linewidth}@{}}
\caption{\textbf{Every mechanism this run proposed, adopted or failed to adopt.}}\label{tab:case_proposals}\\
\toprule
\relictabletitle{6}{Selected case: complete proposal and adoption inventory}
\textbf{Proposal} & \textbf{Proposed} & \textbf{Adopted} & \textbf{Uses} & \textbf{Enf.} & \textbf{Amend.} \\
\midrule
\endfirsthead
\multicolumn{6}{@{}l@{}}{\normalfont\footnotesize \textbf{Table~\thetable{} (continued)}}\\[3pt]
\toprule
\relictabletitle{6}{Selected case: complete proposal and adoption inventory}
\textbf{Proposal} & \textbf{Proposed} & \textbf{Adopted} & \textbf{Uses} & \textbf{Enf.} & \textbf{Amend.} \\
\midrule
\endhead
\midrule
\multicolumn{6}{@{}r@{}}{\normalfont\scriptsize Continued on next page.}\\
\endfoot
\bottomrule
\endlastfoot
\code{P01} & step~12 & step~21 & 419 & 211 & 2 \\
\code{P02} & step~13 & step~39 & 58 & 0 & 0 \\
\code{P03} & step~7 & step~58 & 18 & 0 & 0 \\
\code{P04} & step~48 & step~60 & 247 & 0 & 4 \\
\code{P05} & step~48 & step~65 & 242 & 0 & 7 \\
\code{P06} & step~264 & step~276 & 39 & 0 & 0 \\
\code{P07} & step~44 & never & --- & --- & --- \\
\code{P08} (1st) & step~267 & never & --- & --- & --- \\*
\code{P09} (2nd) & step~278 & step~291 & --- & --- & --- \\*
\code{P10} (3rd) & step~321 & never & --- & --- & --- \\
\end{xltabular}
\RelicTableNotes{Event-ledger proposal and adoption records. A dash denotes an unavailable count.}

\endgroup
\arrayrulecolor{black}
\endgroup

\subsubsection{Adopted and Rejected Mechanisms}
\label{app:case_adoption_results}

The run proposes ten mechanisms, adopts seven, and leaves three unadopted. A merge-readiness handoff is adopted on its second attempt. P02 uses a generic review-before-merge template; P03 uses a generic experiment-logging template. 

\subsubsection{Uses, Enforcements, and Amendments}
\label{app:case_mechanism_counts}

Counts are in Table~\ref{tab:case_proposals}. At the $t=336$ endpoint,
elapsed time since adoption is $336-21=315$ steps for the principal rule and
$336-60=276$ steps for the current-commit integration-evidence rule. Lifecycle records separately track activation, revision, retirement, and last-active state. Amendments
total 13 across three protocols, the last at $t=317$, so revision continued to
within twenty steps of the horizon. The case's 211 reported enforcements are
linked to \code{P01} and 12 distinct pull requests
(Appendix~\ref{app:distinct_prs}). \code{P02}--\code{P06} have positive
reported use counts and zero reported enforcements.   

\subsection{Related Rules for Verification and Evidence Freshness}
\label{app:semantic_convergence}

\subsubsection{Five Convergent Rules}
\label{app:five_rules}

Five related mechanisms address evidence completeness, current-commit
verification, and readiness as repository state changes. Four members
proposed them (M02, M04, M01, M05). The following clauses retain the
source requirements, with task-identifying names and object identifiers
anonymized; the clauses are:

\begin{itemize}
\item \code{P01} (M02, proposed step~12, adopted
step~21, 315 steps from adoption to the endpoint): a pull request affecting public callables or
cross-module behaviour is merge-eligible only when its review record identifies
every changed public callable signature, identifies every added or modified
cross-module invocation, maps each identified item to the applicable written
contract, and \emph{records successful execution of the applicable public smoke
tests through the highest stacking step affected by the change}; anything
unmappable, or any required smoke failure, must be resolved or explicitly rejected
before merge approval.
\item \code{P04} (M04, proposed step~48, adopted
step~60, 276 steps from adoption to the endpoint, revised four times): a pull request changing a
stacking-step implementation or a shared-state boundary must not be merged unless
its recorded CI result and relevant public smoke result \emph{both identify the
exact current PR head commit as the tested commit} and both pass; \emph{if mainline
changes after either result is recorded, the PR must be synchronized and both rerun
on the resulting current head}; and a reviewer or merger \emph{must be able to
compare the recorded tested commit with the current head and determine pass or fail
without relying on an author's statement}.
\item \code{P05} (M01, adopted step~65, revised
seven times): no release, launch announcement or readiness claim may proceed until
the checklist is complete, \emph{CI passes on current mainline}, required reviews
are complete, and the external claim is limited to the verified dependency chain.
\item \code{P06} (adopted step~276): not
merge-ready until the review record contains a complete trace for each changed
completion entry point and \emph{a linked passing affected integration or
public-contract smoke result for the PR head}.
\item \code{P09} (adopted step~291 on the second
attempt): the owner \emph{records current mainline revision and passing CI
immediately before merge; any mainline movement invalidates that}.
\end{itemize}

\subsubsection{Shared Principle}
\label{app:shared_principle}

The following formulation summarizes their shared verification concern:

\begin{quote}
\emph{A verification result belongs to a commit, not to a pull request --- and when
mainline moves, the result expires.}
\end{quote}

The integration-evidence rule explicitly binds both CI and public smoke
results to the current PR head and requires reruns after mainline movement.
The readiness rule refers to current-mainline evidence, while the
contract-boundary rule emphasizes checkable interface-to-contract mappings.
The trace and handoff rules add related obligations. 

The learned rules organize responsibility, review evidence, and readiness obligations around the shared commit-bound CI workflow.

\subsubsection{Relation to the Observed Failure Mode}
\label{app:principle_failure_relation}

The CI ledger measures the absence of that discipline in the other arms
(Table~\ref{tab:ci_ledger}): \Bzero{} made one commit and ran CI against it 73
times, with none passing; \Bone{} made twelve commits, recorded 651 CI runs, one
single commit accounting for 162 of them, and passed two. The baseline arms repeatedly tested a small set of commits while delivering few changes.
This repeated-verification pattern coexists with the environment-level
commit-binding checks shared across conditions and motivates the additional
organizational evidence discipline learned in Relic.

\RelicNeedTable{tab:ci_ledger}{5}
\subsection{CI Ledger}
\label{app:ci_ledger}

\subsubsection{Runs, Commits, Passes, and Merges}
\label{app:ci_summary}

\begingroup
\centering
\begingroup
\footnotesize
\arrayrulecolor{RuleGray}
\setlength{\tabcolsep}{3.5pt}
\renewcommand{\arraystretch}{1.23}
\def\RelicCurrentTable{tab:ci_ledger}
\setlength{\LTcapwidth}{\linewidth}
\begin{xltabular}{\linewidth}{@{}L *{5}{C{0.141\linewidth}}@{}}
\caption{\textbf{CI ledger for the matched diagnostic.}}\label{tab:ci_ledger}\\
\toprule
\relictabletitle{6}{Selected case: commit-bound verification and delivery}
\textbf{Condition} & \textbf{CI runs} & \textbf{Commits} & \makecell{\textbf{Runs/}\\\textbf{commit}} & \textbf{CI passes} & \textbf{Merged} \\
\midrule
\endfirsthead
\multicolumn{6}{@{}l@{}}{\normalfont\footnotesize \textbf{Table~\thetable{} (continued)}}\\[3pt]
\toprule
\relictabletitle{6}{Selected case: commit-bound verification and delivery}
\textbf{Condition} & \textbf{CI runs} & \textbf{Commits} & \makecell{\textbf{Runs/}\\\textbf{commit}} & \textbf{CI passes} & \textbf{Merged} \\
\midrule
\endhead
\midrule
\multicolumn{6}{@{}r@{}}{\normalfont\scriptsize Continued on next page.}\\
\endfoot
\bottomrule
\endlastfoot
\armhead{B0Dark}{B0} & 73 & 1 & 73.0 & 0 & 0 \\
\armhead{B1Dark}{B1} & 651 & 12 & 54.2 & 2 & 2 \\*
\armhead{B2Dark}{B2} & 589 & 50 & 11.8 & 5 & 4 \\*
\armhead{B3Dark}{B3} & 339 & 93 & 3.6 & \bestcell{126} & \bestcell{76} \\
\end{xltabular}

\endgroup
\arrayrulecolor{black}
\endgroup

\subsubsection{Repeated-Verification Pathology}
\label{app:reverification_pathology}

\Bzero{} made one commit in 336 steps and ran CI against it 73 times; none passed.
\Bone{} made twelve commits and recorded 651 CI runs, of which its most-tested
object, \code{C-A}, accounts for 162 by itself; two passed. \Btwo{} sits
between, at 11.8 runs per commit and five passes. \Bthree{} tested each of 93
commits about three and a half times.

The distinction that matters is between testing a lot and testing something that has
moved. A rule requiring current-head evidence and freshness after mainline
movement expresses a discipline relevant to this pathology.  

\subsection{One Rule's Governed Lifecycle}
\label{app:gate_lifecycle}

\subsubsection{Originating Friction and Challenge}
\label{app:gate_friction}

The initiating episode is \code{EP-A}, recorded as a speed-versus-quality
tension in which the timeline did not establish that the implementation work had
been validated end to end. The first challenge is a formal review in which M01
refuses pull request PR-C and states the requirement in the refusal: add the missing
tests, provide the relevant test file and evidence, and request a review before it
can proceed. Recurrence follows as repeated \code{ci\_contract\_break} events
against different pull requests inside the same window. All three are events with
identifiers, actors and steps, and all three were visible to the proposer at
proposal time. 

\subsubsection{Proposal, Support, and Adoption}
\label{app:gate_adoption}

The proposal is made at $t=12$, naming signature drift across a module boundary as
the friction it is written against. It is supported and adopted at $t=21$, with
M02 as supporter. On adoption the compiled spec is registered and enters the action
context. 

\subsubsection{First Use and First Enforcement}
\label{app:gate_first_effect}

The first use event occurs at step 22, one step after adoption. At step
26, the ledger records a protocol-linked refusal of the requested merge
and retains paired before/after snapshots of the governed PR. The event
identity, step, protocol reference, and object snapshots connect the
adopted rule to this concrete execution consequence. Subsequent PR
records link the same object to repair, verification, and its final
delivery state.

\subsubsection{Amendments While in Force}
\label{app:gate_amendments}

The rule is amended twice while in force, at $t=33$ by M01 and at $t=59$ by M02.
Governance applies to the amendments as well as to the rule: the source of the second,
\code{AM-A}, was itself edited by M02 at $t=39$ and by M01 at $t=43$
before it was adopted. Enforcement per step rose from 0.50 to 0.68 across the first
amendment and from 0.55 to 0.69 across the second. These rates summarize the rule's recorded enforcement activity across the two amendment intervals. 

\subsubsection{Later Reuse}
\label{app:gate_reuse}

The principal rule records 419 uses and 211 enforcements through $t=336$, spanning 315 steps from adoption to the endpoint. It satisfies the formation criterion in Appendix~\ref{app:formation_criterion}. P09 is adopted at $t=291$ and has a 45-step post-adoption observation window. 

\subsection{Pull-Request-Level Consequences}
\label{app:pr_level_effects}

\subsubsection{Distinct Pull Requests Affected}
\label{app:distinct_prs}

The 211 enforcement events involve \textbf{12 distinct pull requests}. 

\subsubsection{Delayed and Later-Merged Work}
\label{app:delayed_merges}

Eleven of the twelve blocked requests subsequently merged, 15 to 112 steps
after their first recorded block, usually with additional patches. These block-to-merge intervals include the subsequent implementation, verification,
and coordination needed before delivery. 

\subsubsection{Work Still Blocked at the Observation Horizon}
\label{app:permanent_block}

One request, \code{PR-B}, remained blocked and unmerged at the recorded
horizon, without an observed override, despite approval by an agent
occupying the reviewer role.  Its head, \code{C-B}, was
verified 87 times. The organization responded by building a tool named after the
repair rather than by overriding the gate, preserving the rule while pursuing a repair. 

\subsubsection{Unaffected Merges}
\label{app:unaffected_merges}

Of the 76 merges, \textbf{65 were never blocked by this rule}. Eleven
previously blocked requests later merged and one remained unmerged at the
horizon.

\subsection{Case Interpretation}
\label{app:case_claim_boundaries}
\label{app:case_supported_claims}
\label{app:case_unsupported_claims}

The case traces recurring integration friction into a proposed and adopted rule, followed by repeated use, pull-request enforcement, amendments, and sustained shared delivery.

\section{Integrity and Reproducibility}
\label{app:reproducibility}

This section connects the execution interfaces, frozen artifacts, run
records, and recovery procedures used to reproduce and inspect the
reported experiments.

\subsection{Evaluator Isolation}
\label{app:evaluator_isolation}

\subsubsection{Filesystem and Process Interfaces}
\label{app:evaluator_filesystem}

The member-facing substrate exposes the project identifier, product name,
dataset directory, and starter directory. Evaluator assets occupy the
private hidden-test, held-out-requirement, and reference directories.
Scoring creates a one-shot candidate workspace outside the simulation and
hashes its tree before and after execution. The recorded hashes bind the
score to the exported candidate state.

In the container configuration, evaluation runs in a separate untrusted,
network-disabled container built from a digest-pinned image. In the host
configuration, it runs in a separate process against the exported tree.
The execution-policy receipt identifies the mode, dependencies, and
artifact identities used for the corresponding score
(Appendix~\ref{app:containers_dependencies}).

\subsubsection{Hidden-Test and Reference Isolation}
\label{app:hidden_reference_isolation}

The evaluator resolves hidden suites and reference content through the
32-byte token-keyed vault described in
Appendix~\ref{app:evaluator_artifacts}. World state and checkpoints retain
the versioned binding. Constant-time token comparison controls vault
lookup. Artifact, search, perception, and snapshot tests inspect the
member-facing representations, and initialization validates that private
evaluator assets use the vault interface.

\subsubsection{Output Interface}
\label{app:evaluator_output_interface}

Endpoint evaluation runs after rollout and writes evaluator-side
contract/check records. During rollout, members receive the outputs of
public tests, CI, and ordinary readiness/release gates, including their
associated failure reasons. Hidden-test source, assertions, held-out
requirements, and reference implementation remain in the private evaluator
interface. The run records retain both the member-visible execution
feedback and the offline evaluation needed to reconstruct the sequence.

\subsection{Temporal and Network Controls}
\label{app:temporal_firewall}

\subsubsection{Network Restrictions}
\label{app:network_restrictions}

Repository operations use the in-process repository model, and the search
system's five domains use internal or frozen corpora. The frozen-web
domain is a seeded document snapshot. Replay resolves searches through
the saved corpus and cache; an unresolved lookup raises a cache-miss
error. Product and evaluator execution use the proxy and loopback-only
socket controls in Appendix~\ref{app:containers_dependencies}.
Model-provider calls are the designated network egress. When enabled,
the prompt-visibility interceptor records call hashes and coverage.

\subsubsection{Future-History Restrictions}
\label{app:future_history_restrictions}

The member-visible starter is the designated frozen task snapshot.
Public issue text supplies the problem and acceptance requirements.
Later repository versions, hidden requirements, reference code, and
source-identifying metadata are retained in evaluator-side provenance.
Workload names and upstream version pairs in the paper link the results
to those frozen source records.

\subsubsection{Access Validation}
\label{app:firewall_violations}

Runtime access tests, enabled prompt audits, and post-run future-recall
checks record information-interface validity. A flagged run is handled
under the recorded exclusion policy. The flag, run identity, and validity
reason remain associated with its execution and analysis record.

\subsection{Benchmark Identifiers and Artifact Joins}
\label{app:benchmark_identity}
\label{app:anonymization}

W01--W10 consistently join workload manifests, result tables, transfer
records, and case-study references. Table~\ref{tab:pack_inventory} gives
their names, original pack IDs, and upstream version pairs in the frozen
analysis order. Member, protocol, and event aliases retain consistent
mappings within the case. These identifiers connect the paper's
summaries to the corresponding task definitions and event records.

\subsection{Run-Bundle Schema}
\label{app:run_bundle}

\subsubsection{Configuration and Identity}
\label{app:bundle_config}

The versioned record \code{experiment\_run\_record.json} uses schema
\code{orgenv\_experiment\_run\_v2}. Its identity fields include
\code{run\_id}, \code{pack}, \code{condition}, \code{arm\_id},
\code{seed}, \code{provider}, \code{model}, and
\code{reasoning\_effort}. Configuration fields retain
\code{action\_selection\_mode}, \code{profile\_assignment},
\code{experiment\_phase}, \code{oss\_control},
\code{evaluation\_perturbation}, and \code{mechanism\_ablations}.

The record schema associates these values with the
\code{ablation\_fingerprint}, \code{llm\_runtime\_identity},
\code{llm\_runtime\_fingerprint}, \code{resource\_budget\_fingerprint},
\code{tool\_surface\_fingerprint}, and
\code{information\_budget\_fingerprint}. Artifact fields include
\code{dataset\_manifest\_hash}, \code{starter\_repo\_digest},
\code{reference\_repo\_digest}, \code{candidate\_repo\_digest},
\code{hidden\_suite\_hash}, \code{evaluator\_environment\_hash},
\code{event\_graph\_hash}, and \code{run\_manifest\_hash}.
\code{started\_at}, \code{ended\_at}, and \code{duration\_seconds}
retain execution timing.

\subsubsection{Execution Records}
\label{app:bundle_execution}

The bundle contains the action log, typed world-event log, model-call
records, and perception-derived decision context. \code{ActionDecision}
rows retain validation status and rejection reasons. Checkpoints retain
member state and private memory, repository and product state, the
protocol ledger and proposal objects, scheduler and background-job state,
and the deep per-step replay frames and final snapshot.

Typed identifiers join actions to the work objects they changed and to
the governance or protocol records relevant to the transition. The same
joins support event-ledger counts, distinct-object counts, and case-study
reconstruction.

\subsubsection{Evaluation and Integrity Records}
\label{app:bundle_evaluation}

The final-evaluation artifact retains per-contract status and digest
fields, and the reported checkpoint series covers the three-model study.
\code{llm\_usage} supplies provider-token receipts.
\code{organizational\_capability\_evidence} retains formation evidence
and its hash, while transfer runs carry a \code{capability\_transfer}
receipt. Enabled prompt audits use \code{prompt\_visibility\_audit}
with a chained hash.

A \code{completeness} block checks the record against the expected
field profile for its run and phase, and \code{record\_notes} preserves
validation details. Required-field counts, availability statuses, and
artifact identities remain attached to the record. These checks provide
an explicit input-validation step for replay and aggregation.

\subsection{Checkpointing and Replay}
\label{app:checkpoint_replay}

\subsubsection{Serialization}
\label{app:serialization}

Whole-world serialization stores a versioned payload with a SHA-256,
a separately hashed metadata sidecar, and the case-plan, source-provenance,
model-binding, and resource-budget fingerprints. The live model client
is detached before serialization and reattached after loading. The
payload preserves the organization, member, scheduler, and background-job
state needed to continue from the saved step.

\subsubsection{Resume Equivalence}
\label{app:resume_equivalence}

A zero-step resume loads the checkpoint and checks state equivalence
before advancing the simulation. The resume path validates the case-plan,
source-provenance, model-binding, and resource-budget fingerprints and
rejects a mismatch. The supervised launcher uses this path for recovery
after machine interruption.

\subsubsection{Replay and Continued Execution}
\label{app:replay_semantics_integrity}

Recorded-event replay reconstructs the checkpoint world state, seeded
environment randomness, saved search corpus and cache, and evaluator
identity. Case-study timelines use the recorded events and checkpoints.
Continued execution restores this state and then issues new provider
calls under the saved configuration. The run records retain the restored
checkpoint and subsequent calls as one execution lineage.

\subsection{Run Validity and Recovery}
\label{app:infrastructure_failures}

\subsubsection{Failure Taxonomy}
\label{app:failure_taxonomy}

Timeout, transport, empty/malformed response, and schema failures enter
the provider-error handling path with bounded within-call retries.
Returned-model counters preserve the observed model identity. Container
and dependency faults are recorded as CI or public-test infrastructure
errors. Qualification and evaluator failures retain contract-level
statuses, and checkpoint scoring uses its per-contract timeout.
Checkpoint hashes validate serialized artifacts; machine termination
recovers from the last saved checkpoint.

\subsubsection{Retry and Rerun Policy}
\label{app:retry_policy}

Provider errors are retried within the call according to its deadline
and retry policy. An interrupted run resumes from the last checkpoint
under the same scenario--condition--seed identity, retaining earlier
partial artifacts. A new statistical draw uses a new seed. A provider
blackout activates the batch circuit breaker and pauses remaining
launches. Superseded records retain their lineage and status while the
designated completed record supplies the aggregate.

\subsubsection{Run Inclusion and Version Mapping}
\label{app:run_exclusions}

The main endpoint matrix contains 360 condition runs. Each paired
analysis applies its declared population, metric eligibility, and frozen
asset/version mapping. Run identity, completion status, scoring
eligibility, selected checkpoint lineage, and replacement reason are
retained as separate analysis fields.

Starter or evaluator changes produce a new frozen asset identity. The
aggregation record joins each retained outcome to its designated task,
source package when applicable, manifest, evaluator, and completion
status. Superseded or invalidated records remain linked through the
provenance record, supporting reconstruction of the final included set.

\subsection{Prompts, Schemas, and Configuration Artifacts}
\label{app:prompts_schemas}
\label{app:artifact_release}

\subsubsection{Exact Condition Prompts}
\label{app:exact_prompts}

Relic separates what-action selection from the model calls used to perform,
reflect on, or formalize a selected activity. B0 and B1 use an LLM to choose
from a constrained candidate menu. B2, B3, B3-text, Transfer Text, and Transfer
Exec use SDL for what-action selection; their model calls are module-specific.

\paragraph{Shared cognitive-module guard.}
For reflection, proposal drafting, proposal evaluation, and protocol synthesis,
the system message combines the rendered workload/company brief, global
grounding rules, the module name, and the module instruction. The common guard
is:
\begin{verbatim}
You are the cognition of one agent in a simulated startup. You PROPOSE, REASON,
COMPOSE, EVALUATE and VERBALIZE. You do NOT change the world: the system
validates and executes. Never invent facts, objects, or actions that are not in
the provided context. Return ONLY valid JSON matching the schema.
\end{verbatim}

The composition pattern is:
\begin{verbatim}
SYSTEM =
  COMPANY_BRIEF_TEMPLATE.format(company_name, product_name,
                                product_stage, product_purpose)
  + "\n\nGlobal rules:\n"
  + GLOBAL_GROUNDING_RULES
  + "\n\nCurrent module: " + module_name
  + "\n\n" + module_instruction

USER =
  agent_identity_for(agent, world)
  + "\n\n"
  + module_user(rendered_context)
\end{verbatim}

\paragraph{B0/B1 action choice.}
\begin{verbatim}
Choose the next intentional action for {agent_id} at tick {tick}.

{rendered_action_context}

Return ONLY JSON matching the action schema
(candidate_action and target_object_id MUST identify
one exact entry from candidate_options).
\end{verbatim}
The model receives the available action menu, candidate options, valid targets,
work state, recent events, task board, inbox, pull requests, test results, and
other member-visible context. The returned action is validated against the
candidate pool before execution.

\paragraph{Code editing.}
After an edit action has been selected, the code editor receives the complete
target file, public evidence available to the member, the edit goal, and
readable organizational rules when present:
\begin{verbatim}
Target file:
{target_object_id} -- {title}
{summary}

CURRENT FILE CONTENT:
{complete_file_content}

{current_build_error, if present}
{published_contract, if present}
{imported_public_API, if present}
{readable_protocol_rules, if present}

Edit goal:
{edit_goal}

Reason:
{rationale}

Known gaps:
{known_gaps}

Return the patch JSON with `edits` = the list of anchored
replacements that make this change. Quote each `search`
exactly from the file above; do not return the whole file.

{revision_brief, if an earlier patch was refused}
{anchor_correction, if a replacement did not apply}
\end{verbatim}

\paragraph{Reflection and wish creation.}
The reflection user message is:
\begin{verbatim}
Reflect on the situation.
CONTEXT:
{JSON(reflection_context)}
\end{verbatim}
Its full formation instruction is:
\begin{verbatim}
Reflect on the agent's own and the team's situation after recent events/episodes.
Output an honest self_assessment, team_assessment, blockers, and concrete
improvement_ideas (each with need_type/description/urgency/risk_if_unaddressed).
`need_type` says what kind of thing is missing, and the schema lists the kinds.
Choose protocol_need when the same avoidable mistake keeps happening and what is
missing is a standing rule everyone follows -- a checkable requirement on work
before it lands. Say in `description` what the rule should require and which
recurring failure it prevents; the organization writes and votes on the rule
from that. Choose policy_repair_need when an existing rule is the problem and
should be relaxed or repealed. The rules you are shown carry what each has been
holding up lately: a rule turning away most of the work while nothing ships is a
candidate however sound it reads, and one turning away nothing is not. Then say
in `description` how it should read instead -- write the replacement requirement
in full, keeping the part that was earning its keep and dropping the part
nothing can satisfy. A repair that only says the rule is too strict cannot be
applied: the wording you give is the wording work will be judged against
afterwards. When `what_has_been_failing` is present it is what the gates
actually said, in their words, with how the recent failures divide by kind.
Read it before you name a need: a failure that keeps arriving in the same words
is the clearest thing you have to reason from, and the need you name should
answer what those verdicts say, not what the situation feels like.
\end{verbatim}
The main-study path maps and filters the structured \code{improvement\_ideas}
from this reflection output into wishes; there is no separate LLM wish-extraction
call.

\paragraph{Wish $\rightarrow$ proposal.}
The module instruction is:
\begin{verbatim}
Turn a wish into a concrete, feasible proposal composed from EXISTING
actions/capabilities/artifacts. Do not assume adoption -- name
approval_required_from. A wish is worded as a suggestion -- "propose a
checklist", "create a matrix". proposed_solution is not: it becomes the rule
the organization is held to, so write what must be true of a piece of work,
checkable by someone who was not there. Never write it as a plan to make a rule.
\end{verbatim}
The user message is:
\begin{verbatim}
Draft a proposal for this wish.
CONTEXT:
{
  "wish": {visible wish record},
  "available_actions": [registered action identifiers],
  "product_context": {visible product context}
}
\end{verbatim}

\paragraph{Recurrent-wish clustering.}
The clustering system prompt is:
\begin{verbatim}
You read what the members of one organization have been asking for and name the
few themes that recur. A theme is worth a standing rule when the same avoidable
failure keeps costing the organization and a checkable requirement on work would
have caught it. Group the wishes: cite in `wish_numbers` the ones each theme
rests on, and name no theme that rests on fewer than two. Do not restate a rule
the organization has already adopted, and do not return two themes that a single
rule would cover -- each must stand on its own. `enforcement_rule` must be
checkable against a piece of work by someone who was not there: name what must be
true, not what people should care about. Return nothing rather than something
vague. Reply JSON only.
\end{verbatim}
The user message provides already adopted rules and numbered wishes and asks for
at most the configured number of themes worth a standing rule.

\paragraph{Pattern $\rightarrow$ candidate protocol.}
\begin{verbatim}
Synthesize a candidate protocol from a repeated problem pattern. It is only a
proposal -- never adopt it yourself.
\end{verbatim}
The corresponding user message provides the repeated pattern, occurrence count,
candidate seed, and available actions.

\paragraph{Proposal evaluation.}
\begin{verbatim}
Evaluate the proposal's feasibility/usefulness/risk/adoption. You only score and
suggest revisions; you cannot approve or reject.
\end{verbatim}
with user message:
\begin{verbatim}
Evaluate this proposal.
PROPOSAL:
{JSON(proposal)}
\end{verbatim}
The evaluator returns feasibility, usefulness, risk, adoption score, blocking
issues, and suggested revision. Approval/adoption itself follows the governed
review and approval path.

\paragraph{Wish-to-protocol formation criteria.}
\begin{table}[H]
\centering
\footnotesize
\begin{tabularx}{\linewidth}{@{}P{0.24\linewidth}L@{}}
\toprule
\rowcolor{PanelFill}\textbf{Stage} & \textbf{Recorded criterion}\\
\midrule
Reflection $\rightarrow$ wish & Each idea records need type, description, urgency, and risk; wishes retain source reflection, related objects/episodes, supporters, urgency, and status. Similar open wishes are merged.\\
Wish $\rightarrow$ proposal & Eligible when urgency is at least 0.85, it has at least two supporters, spans at least two episodes, or has founder endorsement; proposal provenance retains source wish/reflection/episode links.\\
Proposal $\rightarrow$ adoption & A protocol proposal with available provenance represents recurrence across at least two episodes or two wishes. Adoption requires at least three steps of review and two distinct approvers in the reported semi-automatic governance configuration.\\
Weak formation & Valid proposal/adoption, at least two post-adoption uses on distinct use units and steps, and at least 48 steps from adoption to the last qualifying use.\\
Strong evidence & Weak formation plus validated third-party enforcement, governed-state change, and independently evaluated outcome evidence.\\
\bottomrule
\end{tabularx}
\caption{Wish-to-protocol formation and measurement criteria.}
\label{tab:wish_protocol_criteria}
\end{table}

\paragraph{Formation-record fields.}
The per-run formation interface includes \code{protocol\_count},
\code{weak\_protocol\_count}, \code{strong\_protocol\_count},
\code{time\_to\_emergence}, \code{repeated\_protocol\_use\_rate},
\code{cross\_context\_protocol\_reuse\_rate},
\code{protocol\_persistence\_rate},
\code{valid\_third\_party\_enforcement\_rate},
\code{measurable\_impact}, and \code{protocol\_proposal\_rate}.
Post-adoption persistence, amendment/repair, and episode-transfer records
provide lifecycle detail. The autonomous census joins these records by
within-run lineage and applies the formation criteria in
Table~\ref{tab:wish_protocol_criteria}.

\paragraph{Readable protocol exposure in Transfer Text and Exec.}
Both treatments use the frozen \code{canonical\_v2} package. Its semantic
SHA-256 is
\code{a627adfcf6c299d0fabb185345bbc01106ea6901380aa0fbfee37c830ad8cdb6}.
The shared 3,148-character code-editor rule block has UTF-8 SHA-256
\code{9c505e567afab8853436c1e7cbc609e51928de2e29e64be299d4721bef09ab49}.

\begin{table}[H]
\centering
\small
\begin{tabularx}{0.88\linewidth}{@{}L C{0.25\linewidth} C{0.25\linewidth}@{}}
\toprule
\rowcolor{PanelFill}\textbf{Transfer component} & \textbf{Text} & \textbf{Exec}\\
\midrule
Frozen package & \code{canonical\_v2} & \code{canonical\_v2}\\
Six readable summaries and code-editor rendering order & identical & identical\\
Compiled machine bindings & 0 & 6\\
New target protocol formation/revision & disabled & disabled\\
\bottomrule
\end{tabularx}
\caption{Content matching for the Text/Exec transfer comparison.}
\label{tab:transfer_prompt_content_match}
\end{table}

The six verbatim readable summaries are:
\begin{enumerate}\footnotesize
\item \textbf{Issue ownership before PR opening.} ``Before \code{open\_pr}, resolve the source branch exactly as the action handler does. If its commits link work, every explicit linked task must exist and have a non-empty \code{owner\_id}. Every native Issue must have \code{owner\_id} set, and every Task associated with that issue must also exist and be owned. A ProductArtifact issue has no inferred owner: it must resolve to at least one associated Task, and every associated Task must exist and be owned. A branch with no issue or task linkage is outside this guard. Assign every missing owner and retry; editing and committing remain allowed.''
\item \textbf{Branch ownership.} ``Before \code{open\_pr}, resolve the source branch exactly as the action handler does and require \code{actor.id} to equal \code{branch.owner\_id}. A non-owner is blocked without changing the branch; the recorded owner can open the request, so editing, committing, review, and merge repair lanes remain available.''
\item \textbf{CI retry on changed state.} ``Before \code{run\_ci} or \code{ci\_test}, resolve the pull request exactly as the action handler does. Block only when its latest CI result is failed and already attests the current source-branch head against the current mainline commit list. A new branch head or changed mainline permits a retry; a \code{not\_run} infrastructure result follows the native retry path.''
\item \textbf{Current CI attestation at merge.} ``Before \code{merge\_pr} for a pull request carrying commits or patches, require approved status and no merge conflict; require \code{pr.commit\_ids} to exactly equal the source branch commit list and \code{pr.patch\_ids} to exactly equal the ordered flattening of those commits' \code{patch\_ids}; require both \code{pr.ci\_passed} and a latest passed CI result for the current branch head and require \code{ci\_base\_main\_commit\_ids} to equal the current \code{repo.main\_commit\_ids}. When the active profile exposes a deterministic candidate-tree digest, \code{ci\_tree\_hash} must exactly equal that digest; ordinary RepoLite uses the commit head and mainline base as its exact identity and does not require an unavailable digest. Missing, failed, conflicted, unsynchronized, or stale evidence blocks only merge; commit and \code{run\_ci} remain available to repair it.''
\item \textbf{Independent review.} ``When the roster contains an agent other than the pull-request author, \code{review\_pr}, \code{approve\_pr}, and \code{formal\_pr\_review} by that author are blocked, and \code{merge\_pr} requires at least one \code{approved\_by} identifier that is both a current roster member and different from \code{author\_id}. Unknown actor identifiers cannot review or approve. A one-agent roster may self-review so the workflow does not deadlock; \code{request\_changes} and all repair actions remain allowed.''
\item \textbf{Release-gate coverage.} ``Before \code{publish\_product\_release}, resolve the release candidate exactly as the action handler does. Require approved status, the roster-aware \code{release\_approval\_met} check, no blockers, a non-empty \code{included\_pr\_ids} list whose requests all exist, are merged, and have \code{ci\_passed}. Each candidate must carry the exact non-empty, duplicate-free gate profile selected by \code{release\_gates\_for(world)}. Each profile gate must have exactly one recorded passed result or its exact identifier must already be listed in \code{waived\_gates}; unknown gates, waivers, and results are rejected. Run readiness, collect the required approvals, and repair blockers before retrying publish.''
\end{enumerate}

\subsubsection{Configuration Artifacts}
\label{app:brief_templates}

Frozen workload manifests and brief templates define the public task inputs,
entry points, progressive contract steps, and public-test commands.
Machine-readable schemas define the member and organizational state,
proposal and protocol fields, and runtime bindings. The exact model
prompts and transferred rule summaries above identify the text consumed
by the corresponding cognitive modules.

Transfer metadata links the source package, content digest, freeze record,
carrier documents, role mapping, and target reset state to the treatment
configuration. Appendix~\ref{app:transfer_receipts} specifies the
\code{capability\_transfer} fields used to inspect the injected package
and its later execution records.

\section{Human--Organization Interfaces}
\label{app:hci}

Relic primarily studies whether autonomous agent organizations can accumulate
persistent organizational capabilities. Once such an organization can maintain
its own workstreams, responsibilities, review processes, protocols, evidence,
and institutional state, however, a second problem appears: \emph{how should a
human participate in an organization whose internal operating representation
and execution tempo are increasingly machine-native?}

We explore three interface designs for human participation in a continuously operating agent organization.

\subsection{Why Direct Human Participation Becomes Difficult}
\label{app:hci_motivation}

Our initial design treated the human as another organizational member. This
preserves an appealing symmetry: human and agent members communicate through
the same channels, observe organizational objects available to their roles, and
participate in the same workflows. In practice, this design exposes a
representation mismatch.

Relic maintains structured organizational state including task ownership,
review and CI status, active protocols, approvals, episodes, evidence,
commitments, requests, and institutional memory. These representations are
useful for agent decision making and organizational governance, but they are
not necessarily the representation in which a human wants to understand
ongoing work. A person entering the organization may first have to determine
which workstreams matter, what is blocked, which evidence is current, who owns
a decision, and whether an apparent problem has already been handled elsewhere.

Conversely, a natural-language message from a human is only one event in a
continuously operating organization. Asking ``Why is this blocked?'' does not
itself guarantee that the organization will stop, that the relevant agent will
immediately answer, or that no other authorized member will act before the
human finishes considering the issue.

A second prototype introduced a secretary agent that could help the human
delegate implementation and verification work. This reduced some execution
burden, but much of the underlying organizational state remained directly
exposed. The human still had to interpret the machine-native organization and
decide which internal objects deserved attention.

\begin{figure}[t]
    \centering
    \includegraphics[width=\linewidth]{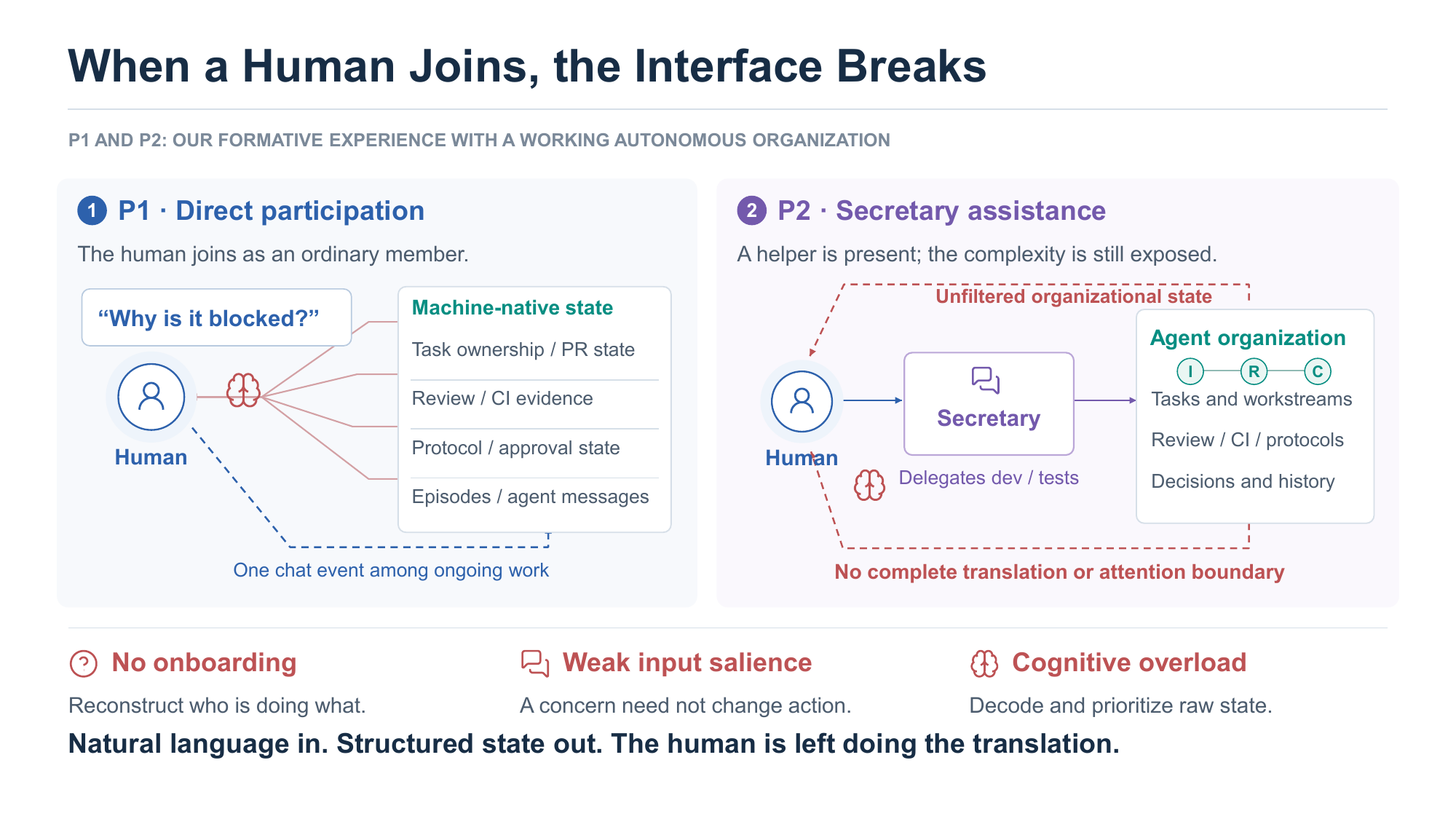}
    \caption{
    \textbf{Interface progression before P3.}
    In P1, the human joins the autonomous organization directly and must
    interpret machine-native organizational state. In P2, a secretary assists
    with delegated work, but substantial organizational complexity remains
    directly exposed. These experiences motivate a liaison-first abstraction
    boundary in P3.
    }
    \label{fig:hci_teaser}
\end{figure}

Figure~\ref{fig:hci_teaser} summarizes the interface progression that motivated
a stronger human--organization abstraction boundary.

\subsection{Moving the Human--Organization Abstraction Boundary}
\label{app:hci_abstraction}

We organize the design space into three interface levels. The progression changes how organizational complexity is presented to the human.

\paragraph{P1: Direct organization.}
The human directly navigates organizational state and communicates with
organizational members. This maximizes direct visibility into the organization
but places the burden of orientation, interpretation, and attention allocation
on the human.

\paragraph{P2: Transparent liaison.}
A secretary assists with communication and delegated work while the underlying
agents, workstreams, and organizational structures remain directly visible.
The secretary reduces some execution burden, but the human still operates close
to the machine-native representation and remains responsible for understanding
a substantial fraction of organizational state.

\paragraph{P3: Liaison-first interaction.}
The secretary becomes the primary human-facing interface. The organization is
not simplified internally. Instead, its state is summarized, translated, and
selectively surfaced to the human. Detailed organizational evidence and
provenance remain available on demand.

\begin{table}[t]
    \centering
    \small
    \arrayrulecolor{RuleGray}
    \begin{tabular}{p{1.2cm}p{2.8cm}p{4.8cm}}
        \toprule
        \relictabletitle{3}{Hybrid human--organization interface levels}
        & \textbf{Primary interface} & \textbf{Main interaction property} \\
        \midrule
        P1 &
        Organization directly &
        Human interprets organizational state and communicates with members
        directly. \\
        P2 &
        Secretary + visible organization &
        Secretary assists, but substantial organizational complexity remains
        exposed. \\
        P3 &
        Secretary first; organization on demand &
        Secretary translates state and intent while preserving access to
        evidence and provenance. \\
        \bottomrule
    \end{tabular}
    \arrayrulecolor{black}
    \caption{
    Three human--organization interface levels.
    }
    \label{tab:hci_levels}
\end{table}

\subsection{P3: A Liaison-First Hybrid Human--Agent Organization}
\label{app:hci_p3}

P3 places the secretary at the human-facing boundary while preserving the
autonomous organization. It combines three concurrently operating components:
a full autonomous Relic organization, a human member with a human-facing
office, and the shared organizational workflow.

\begin{figure}[t]
    \centering
    \includegraphics[width=\linewidth]{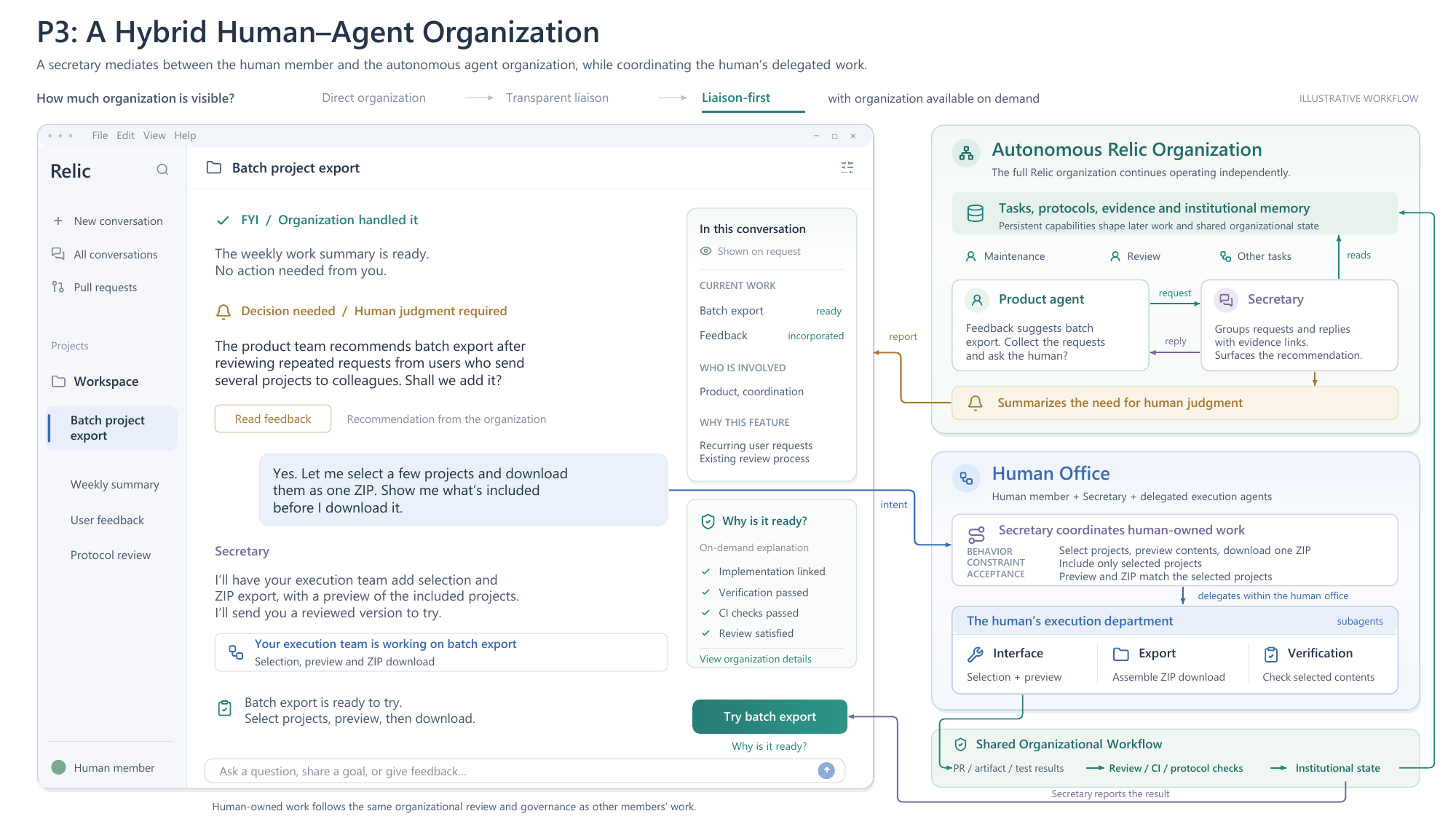}
    \caption{
    \textbf{P3: a liaison-first hybrid human--agent organization.}
    The autonomous Relic organization continues operating independently
    (upper right). The human interacts primarily through a secretary, which
    summarizes organizational state, surfaces items requiring human judgment,
    and translates human intent into organizational work. The human also has
    delegated execution agents inside a human office. Human-owned artifacts
    subsequently enter the same shared review, CI, protocol, and institutional
    workflow as other organizational work.
    }
    \label{fig:hci_p3}
\end{figure}

\paragraph{Autonomous Relic organization.}
The upper-right component of Figure~\ref{fig:hci_p3} is the autonomous Relic organization. It continues to maintain its own
workstreams, tasks, responsibilities, review processes, protocols, evidence,
and institutional memory. Other organizational members continue working
independently of the human-facing interaction.

The organization may discover problems, aggregate evidence, form
recommendations, execute unrelated work, and update its institutional state
without waiting for continuous human direction. 

\paragraph{Human office.}
The human is represented as a member of the larger organization and retains
their own execution capacity. The human office contains the human, the
secretary, and delegated execution agents that can carry out work belonging to
the human member's responsibilities. For example, a human may specify a desired
behavior and constraints, after which the secretary can coordinate
implementation or verification work on the human's behalf.

The secretary consequently has two distinct responsibilities:

\begin{enumerate}
    \item \textbf{organizational liaison:} translate organizational state into
    human-readable summaries and route human intent back into organizational
    objects; and

    \item \textbf{human-work coordinator:} decompose and delegate work that
    belongs to the human member's own organizational responsibilities.
\end{enumerate}

The secretary coordinates the human-facing interface and delegated work while
shared governance and the organization's persistent decision processes
continue to govern organization-wide work.

\paragraph{Shared organizational workflow.}
Artifacts produced through the human office re-enter the same organizational
workflow as work produced by other members. Code, documents, tests, and other
artifacts remain subject to applicable review, CI, evidence requirements, and
active protocols. 

Together, these components form a \emph{hybrid human--agent organization}: the
agent organization continues operating autonomously, while the human
participates through a human-facing abstraction layer and retains delegated
execution capacity.

\subsection{Bidirectional Translation}
\label{app:hci_translation}

The secretary acts as a bidirectional abstraction layer between two different
representations of work.

\paragraph{Organization to human.}
The autonomous organization may contain many simultaneously active tasks,
reviews, failures, protocols, and commitments. The secretary compresses this
state into human-facing updates. We distinguish three forms of communication:

\begin{itemize}
    \item \textbf{informational updates}, for work that the organization has
    handled without requiring human action;

    \item \textbf{decision requests}, for cases in which human judgment,
    preference, authorization, or unavailable information is required; and

    \item \textbf{on-demand explanation}, through which the human can request
    the evidence, provenance, review history, or organizational trace behind a
    summary.
\end{itemize}

P3 therefore uses progressive disclosure. The default representation is
concise, while the underlying organizational state remains inspectable when the
human needs additional evidence.

\paragraph{Human to organization.}
Human input is expressed in ordinary language but may imply different
organizational actions. For example, ``I do not trust this result'' may
indicate a request for additional evidence, whereas ``always review these
changes'' may express a persistent policy-level intention rather than a
one-time task.

The secretary can translate such input into structured organizational objects
such as tasks, concerns, review requests, constraints, acceptance criteria, or
protocol proposals. The resulting object then follows the organization's
ordinary validation, approval, and execution path.

\subsection{Human Attention as an Organizational Resource}
\label{app:hci_attention}

A highly autonomous organization should not require the human to inspect every
internal action. At the same time, aggressive filtering can conceal important
disagreement or make consequential decisions difficult to inspect.

This creates an organizational problem of deciding \emph{what} deserves human
attention. Routine and reversible issues may be handled internally; recurring
but non-urgent issues may be summarized periodically; high-impact,
irreversible, or unresolved decisions may require direct escalation. Escalation
policy therefore becomes part of the human--organization interface rather than
only a notification preference.

The completed walkthroughs expose a second dimension of the same problem:
\emph{when} human attention can arrive. An organization must decide not only
which events deserve human judgment, but also whether its own execution should
wait, slow down, continue autonomously, or defer irreversible actions while
that judgment is pending.

\subsection{Formative Paired Walkthrough Protocol}
\label{app:hci_walkthrough}

Four members of the author team completed formative paired walkthroughs of P2 and P3 after implementation. None had participated in the design or implementation of Relic or the two interfaces. Each explored ongoing work, inspected blockers and evidence, communicated a requested change, and responded to decisions requiring human judgment. 

\subsection{Paired Internal Walkthrough Results}
\label{app:hci_observations}

\paragraph{P3 reduced organizational interpretation burden.}
Across the four paired walkthroughs, the liaison-first P3 interface was
consistently easier to follow than P2. In P2, the secretary could assist with
delegated work, but the evaluator still had to inspect and interpret substantial
machine-native organizational state before deciding how to intervene. Active
workstreams, agent activity, pending reviews, protocol state, evidence, and
other internal objects remained directly salient to the human interaction.

In P3, the secretary absorbed much of this interpretation burden. It summarized
ongoing organizational state, surfaced the subset of issues that required
attention, and allowed the evaluator to request underlying evidence or
provenance when needed. The organization itself remained unchanged internally;
what changed was the human-facing representation. The interaction therefore
shifted from navigating the organization toward judging the decisions and
exceptions that the organization surfaced.

\paragraph{Human deliberation became the remaining bottleneck.}
A recurring observation was that improved legibility did not remove the
difference in operating speed between humans and autonomous agents. While an
evaluator was reading a summary, inspecting evidence, or considering whether
to approve, reject, or redirect work, other agents could continue operating
elsewhere in the organization.

In some interactions, useful work advanced while the human was still
deliberating. When another agent had sufficient authority over a pending item,
that item could also be resolved before the human responded. The resulting
friction was therefore no longer primarily about understanding the state of
the organization. It was about whether human judgment could arrive on the same
timescale as the organization that was waiting for, or acting around, that
judgment.

This makes the timing of intervention a separate interface variable. A human
may understand the issue perfectly and still be too slow to participate in
every decision at machine execution speed.

\paragraph{Pausing provides control but changes the workflow.}
The interface can pause the organization while the human deliberates. This is
a practical control mechanism: it prevents the relevant organizational state
from continuing to evolve while the human reads evidence and forms a judgment.

However, pausing also changes the interaction regime. A continuously operating
autonomous organization becomes temporarily synchronized to human decision
speed. If every consequential decision requires such a pause, the resulting
system behaves less like an autonomous organization with intermittent human
oversight and more like a workflow that repeatedly waits for a human operator.

\paragraph{Representation mismatch and temporal mismatch are distinct.}
The P2/P3 comparison separates two problems that can otherwise be conflated.

The first is \emph{representation mismatch}: directly exposing machine-native
organizational state makes it costly for a human to understand what is
happening. P3 substantially reduces this burden through liaison-first
summarization and progressive disclosure.

The second is \emph{temporal mismatch}: even when the relevant state is
presented clearly, human deliberation may remain slower than the organization
that is acting around that decision. Better summarization reduces the time
needed to understand a situation, but it does not make human reasoning operate
at agent execution speed.

A human--organization interface must therefore decide both \emph{what} should
be escalated and \emph{how the organization should behave while judgment is
pending}. Possible policies include waiting only for irreversible actions,
continuing reversible work, slowing selected workstreams, assigning decision
deadlines, or allowing another authorized member to proceed after a timeout.

\subsection{Authority and Interaction Records}
\label{app:hci_boundary}

Visibility, permissions, and decision rights follow the human and
secretary roles. Human-owned artifacts re-enter the shared review, CI,
and protocol workflow. Interaction records connect the human's original
instruction, the secretary's structured interpretation, the resulting
work or governance object, and subsequent execution.

The human can request the organizational evidence and provenance behind a
summary. This record preserves the path from a human-facing decision to
its source evidence and organizational consequences while the shared
governance process remains active.

\section{External Software Benchmark: CooperBench}
\label{app:cooperbench}

CooperBench evaluates joint implementation of interacting features in existing
software repositories~\citep{cooperbench2026}. We evaluate the complete
two-member Relic architecture on the full benchmark against released Solo,
Peer/coop+git, and hierarchical Team trajectories. We additionally evaluate a fixed 48-pair same-model subset as a controlled
test of coordination loss and recovery.

\subsection{Full-Benchmark Scope and Success Criterion}
\label{app:cooperbench_scope}

The full CooperBench evaluation space contains 652 exact interacting-feature
pairs. A pair succeeds when the delivered candidate passes the official tests
for both features (\code{both\_passed=true}); each exact pair is one scoring
unit.

Benchmark-validity auditing ran alongside Relic execution. Once an issue
family was established, all affected exact pairs were entered into the
validity ledger. The completed audit excludes 183 exact pairs and leaves
469 valid pairs. Relic passes 371 of those 469 pairs; the Raw and
definite-defect matched analyses use their corresponding exact-pair identities.

We report three validity policies:

\begin{itemize}
    \item \textbf{Raw.}
    No benchmark-validity exclusion is applied. The matched Relic analysis uses
    the 616 exact pairs with formal official verdicts.

    \item \textbf{Definite-defect exclusion.}
    We remove only pairs affected by a directly established
    specification--verification mismatch, unpublished mandatory interface or
    representation, explicit behavioral conflict, or established cross-feature
    incompatibility. This policy excludes
    \textbf{111 unique exact pairs from 21 issue families}, leaving 541 pairs.

    \item \textbf{Final validity exclusion.}
    We remove every exact pair admitted by the completed benchmark-validity
    audit. The final ledger excludes \textbf{183 unique exact pairs}, leaving a
    \textbf{469-pair} validity-filtered benchmark.
\end{itemize}

For released systems, trajectories for all 652 pair identities are available.
We therefore recompute their results on the same fixed 652-, 541-, and
469-pair universes. Only an official PASS contributes to the numerator; a
non-passing or missing formal verdict does not further shrink the
policy-defined denominator.

\subsection{Two-Member Relic Adaptation}
\label{app:cooperbench_adaptation}

The adapter places two feature owners in one shared organization. Each starts
from its assigned feature brief and works through a private workspace and
branch. Requirements and local findings reach the other member through explicit
messages or shared artifacts. Pair-specific run identities keep workspaces,
patches, messages, and organizational state separate across concurrent pairs.

The B3-2 configuration retains SDL, governed protocol formation, and executable
bindings. SDL selects feasible work, communication, review, and governance
actions; the model produces code and messages. Members can propose rules from
visible coordination failures, submit them for validation and approval, and
apply adopted rules to subsequent work.

Joint delivery follows owner-local editing and public testing, commits and pull
requests bound to a code revision, peer review, integration replay, and
synchronization to shared mainline. Both members attest to the same mainline
digest and export the same joint patch to their submission slots. The official
evaluator then applies the two native feature test suites to that delivered
candidate.

Public-probe receipts retain the probe, associated public requirements,
candidate and relevant baseline statuses, and bounded execution errors and
output. Fixed patch guards and artifact-identity checks implement the adapter,
while learned protocols retain their own proposal, adoption, and execution
records. The final official evaluation is linked to the exact pair identity,
model, execution settings, and adapter revision.

Initial adapter development used a separate task to check coordination,
integration, and submission before the reported evaluations. The full-benchmark
Relic runs use Claude Opus 4.6 with high reasoning effort.

\subsection{Baseline Configurations}
\label{app:cooperbench_baselines}

The full-benchmark comparison uses three released GPT-5.5-hao references
~\citep{cooperbench_team_trajectories_2026,cooperbench_code_2026}.
\textbf{Solo} assigns both interacting features to one coding agent.
\textbf{Peer / coop+git} assigns one feature to each of two peers without a
permanent lead. \textbf{Team no-proto} uses the released lead--member hierarchy
with shared task state and scratchpad.

Across the released runs, these structures exhibit a stable coordination
hierarchy:
\[
\text{Peer} < \text{Solo} < \text{Team}.
\]
Splitting interacting features across plain peers reduces success relative to
Solo, whereas the structured lead--member Team exceeds Solo. Relic retains the
peer topology with no permanent lead and adds persistent organizational state
and executable governance to that structure.

The separate 48-pair controlled comparison uses Claude Opus 4.6 with high
reasoning effort for Relic, Official Peer, and Official Solo. Official Peer
assigns one interacting feature to each peer; Official Solo assigns both
features to one agent. Relic retains the same model and two-peer feature split.

\subsection{Benchmark-Validity Audit}
\label{app:cooperbench_audit}

\subsubsection{Hidden Tests Versus Hidden Requirements}

We distinguish withheld \emph{test instances} from withheld
\emph{requirements}. A benchmark may hide concrete test code, input instances,
edge cases, reference implementations, and evaluator data. However, for a
hidden evaluation to measure satisfaction of the stated task, the acceptance
semantics exercised by those tests must be supported by the public task
contract available to the evaluated system. In short,
\emph{test instances may be hidden, but task requirements should not be}.

For example, if a public specification requires correct handling of all valid
JSON inputs, a previously unseen valid JSON instance is an appropriate hidden
test. By contrast, if the public specification requires JSON and CSV support
while the evaluator additionally requires YAML, the evaluator has introduced
an unpublished requirement rather than merely an unseen test instance.
Likewise, a requirement to produce an error log does not by itself justify an
evaluator that accepts only one unpublished exact log string.

Our audit therefore asks whether behavior enforced by the official evaluator is
supported by the public specification.

\subsubsection{Audit Scope and Outcome Distribution}

The audit identifies \textbf{38 specification--verification issue families}
affecting \textbf{183 distinct exact feature pairs}. The sum of family-level
pair incidences is 194 because 11 exact pairs instantiate two issue families;
deduplication by exact pair identity yields 183 affected scoring units.

Once an issue family is admitted, it applies to every affected exact pair.
In the complete 183-pair final exclusion set, the raw Relic states comprise
\textbf{21 PASS, 123 FAIL, and 39 pairs without an official verdict}.
The 111-pair definite-defect subset contains
\textbf{5 PASS, 76 FAIL, and 30 pairs without an official verdict}.
The validity audit asks whether a scoring unit can distinguish failure to
implement the public requirement from failure to satisfy an unpublished or
materially under-specified evaluator assumption.

All original evaluator outcomes and experimental artifacts are retained.
Validity exclusion changes only whether an exact pair contributes to the
corresponding aggregate score; it does not rewrite its raw outcome.

\subsubsection{Validity Evidence Levels}
\label{app:cooperbench_validity_levels}

The six failure classes below describe \emph{what kind} of contract problem is
present. Independently, each issue family is assigned one of two evidence
levels: a \emph{definite defect} or a \emph{specification ambiguity}.

\paragraph{Definite defects.}
A definite defect is directly established from the public specification and
evaluator behavior: an explicit specification--verification mismatch, an
unpublished mandatory interface or internal structure, an unpublished exact
representation, an explicit behavioral conflict, or an established
cross-feature incompatibility. The audit contains
\textbf{21 definite-defect families covering 111 unique exact pairs}.

\paragraph{Specification ambiguities.}
A specification ambiguity occurs when more than one behavior remains
consistent with the public contract, while the evaluator accepts only one such
interpretation. The D/A-tiered audit contains \textbf{12 ambiguity families
covering 71 unique exact pairs}; seven of these pairs also appear in the
definite-defect set. The completed audit contains five additional issue
families covering eight further exact pairs, for \textbf{38 issue families and
183 excluded exact pairs} in total.

Thus, the definite-defect sensitivity policy excludes 21 families / 111 pairs,
while the completed final validity policy excludes 38 families / 183 pairs.

\begin{table}[H]
\centering
\begingroup
\footnotesize
\arrayrulecolor{RuleGray}
\setlength{\tabcolsep}{2.5pt}
\renewcommand{\arraystretch}{1.08}

\begin{tabularx}{\linewidth}{
    @{}
    C{0.055\linewidth}
    L
    C{0.12\linewidth}
    C{0.12\linewidth}
    C{0.12\linewidth}
    C{0.12\linewidth}
    @{}
}
\toprule
\rowcolor{PanelFill}
\textbf{Class} &
\textbf{Specification--verification failure} &
\textbf{Issue families} &
\textbf{Primary pairs} &
\textbf{D coverage} &
\textbf{A coverage}\\
\midrule

C1 &
Public contract and evaluator acceptance are inconsistent &
9 & 48 & 47 & 0\\

C2 &
Hidden interface, internal structure, or input-domain requirement &
7 & 34 & 30 & 5\\

C3 &
Callback protocol is not defined by the public contract &
3 & 18 & 0 & 20\\

C4 &
Behavioral scope or boundary semantics are not determined &
11 & 49 & 3 & 48\\

C5 &
Exact output representation is not publicly specified &
3 & 20 & 21 & 0\\

C6 &
Cross-feature contract or retained-test incompatibility &
5 & 14 & 12 & 0\\

\midrule
\textbf{Across-class total / union} &
&
\textbf{38} &
\textbf{183} &
\textbf{111} &
\textbf{71}\\
\bottomrule
\end{tabularx}

\caption{\textbf{CooperBench validity taxonomy and evidence levels.}
``Issue families'' counts all 38 root issue families in the completed audit.
``Primary pairs'' assigns each affected pair to one class in a fixed C1--C6
order so that the column sums to 183. D/A coverage reports the explicitly
tiered definite-defect and ambiguity subsets and is not additive across
classes.}
\label{tab:cooperbench_validity_taxonomy}

\endgroup
\arrayrulecolor{black}
\end{table}

The completed audit adds one C1 issue, one C2 issue, two C4 issues, and one
C6 issue beyond the D/A-tiered family set; these five families account for the
eight additional excluded exact pairs.

\subsubsection{Complete Issue-Family Inventory}
\label{app:cooperbench_issue_families}

Table~\ref{tab:cooperbench_issue_family_inventory} lists every admitted issue
family. Because CooperBench scores exact feature pairs rather than isolated
features, a defect associated with one repeatedly paired feature can propagate
to several scoring units.

\begingroup
\scriptsize
\arrayrulecolor{RuleGray}
\setlength{\tabcolsep}{2.0pt}
\renewcommand{\arraystretch}{1.06}

\begin{longtable}{
    @{}
    C{0.05\linewidth}
    C{0.075\linewidth}
    P{0.245\linewidth}
    P{0.52\linewidth}
    C{0.065\linewidth}
    @{}
}
\caption{\textbf{Complete CooperBench specification--verification issue-family
audit.} D denotes a definite defect; A denotes specification ambiguity; ``--''
marks a final exclusion family without a separate D/A sensitivity-tier
assignment. ``Pairs'' is the number of exact feature pairs affected by the
family and may overlap across rows.}
\label{tab:cooperbench_issue_family_inventory}\\

\toprule
\rowcolor{PanelFill}
\textbf{ID} &
\textbf{Class / tier} &
\textbf{Repository / task / feature} &
\textbf{Observed contract problem} &
\textbf{Pairs}\\
\midrule
\endfirsthead

\toprule
\rowcolor{PanelFill}
\textbf{ID} &
\textbf{Class / tier} &
\textbf{Repository / task / feature} &
\textbf{Observed contract problem} &
\textbf{Pairs}\\
\midrule
\endhead

\midrule
\multicolumn{5}{r}{\footnotesize Continued on next page}\\
\endfoot

\bottomrule
\endlastfoot

R01 &
C1 / D &
\texttt{go\_chi / 56 / f1} &
The public example permits ``GET, HEAD'' as one Allow-header value, whereas
the evaluator indexes two separate values from \code{Header.Values("Allow")}. &
4\\

R02 &
C2 / D &
\texttt{go\_chi / 27 / f3} &
The evaluator directly requires internal logging symbols that are not part of
the publicly specified interface. &
3\\

R03 &
C4 / A &
\texttt{go\_chi / 26 / f3} &
The public contract does not fully determine the version-selection entry point
or fallback behavior for an unknown selector or role; the evaluator selects one
specific behavior. &
3\\

R04 &
C4 / A &
\texttt{outlines / 1371 / f2} &
The public contract does not specify the behavior for unregistering a
nonexistent filter; the evaluator requires
\code{unregister\_filter("non\_existent")} to succeed without raising. &
3\\

R05 &
C4 / D &
\texttt{outlines / 1371 / f4} &
The public task specifies JSON/CSV behavior, while the evaluator additionally
requires YAML support. &
3\\

R06 &
C4 / A &
\texttt{outlines / 1655 / f4} &
The public contract permits colon or hyphen separators but does not determine
whether they may be mixed; the evaluator rejects mixed forms. &
9\\

R07 &
C6 / D &
\texttt{outlines / 1706 / f1} &
The new feature requires wording that includes both Transformers and MLXLM,
while retained tests from interacting features still match the older
Transformers-only wording. &
7\\

R08 &
C1 / D &
\texttt{tiktoken / 0 / f9} &
The public contract names the parameter \code{compress}, whereas the evaluator
calls \code{compression}. &
9\\

R09 &
C2 / D &
\texttt{tiktoken / 0 / f10} &
The public task specifies LRU/\code{use\_cache} behavior, while the evaluator
additionally requires an assignable \code{enc.cache} attribute that is not
published in the contract. &
9\\

R10 &
C3 / A &
\texttt{tiktoken / 0 / f7} &
The validator protocol is underspecified: the contract does not determine
whether validation returns a Boolean or raises directly, while the evaluator
fixes a False-returning predicate protocol. &
9\\

R11 &
C5 / D &
\texttt{click / 2068 / f10} &
The public task requires error handling, but the evaluator additionally
requires the unpublished exact text ``Editing failed''. &
11\\

R12 &
C1 / D &
\texttt{click / 2956 / f7} &
The public example/model input uses an underscore-form parameter name, whereas
the hidden acceptance condition requires the hyphenated CLI spelling. &
7\\

R13 &
C4 / A &
\texttt{click / 2956 / f3} &
The public contract specifies \code{ValueError} for negative batch size but
does not determine the exception class for non-multiple input; the evaluator
requires \code{TypeError}. &
7\\

R14 &
C6 / D &
\texttt{click / 2068 / paired features} &
One feature explicitly requires \code{wait(timeout=timeout)}, while retained
mocks in interacting-feature tests expose a narrower \code{wait} signature. &
3\\

R15 &
C6 / D &
\texttt{click / 2068 / f1--f8} &
The new feature moves to \code{edit\_files}, while the interacting feature's
retained tests still depend on the older internal \code{edit\_file} API. &
1\\

R16 &
C1 / D &
\texttt{jinja / 1559 / f3} &
The public task requires retries beginning at priority 0, while the evaluator
also requires the priority-0 case to invoke the operation only once. &
9\\

R17 &
C4 / A &
\texttt{jinja / 1465 / f5} &
The public task discusses \code{None}, empty-string, and falsy values but does
not define grouping semantics for a completely missing attribute; the
evaluator requires one particular treatment. &
9\\

R18 &
C3 / A &
\texttt{HF datasets / 3997 / f2,f3} &
The \code{custom\_criteria} callback arguments are not specified. The evaluator
accepts a single feature argument but rejects the alternative
\code{(name, feature)} protocol. &
7\\

R19 &
C4 / A &
\texttt{HF datasets / 3997 / f4} &
The public contract does not specify whether a JSON-formatted string supplied
as an update value should be automatically deserialized into a dictionary; the
evaluator requires that behavior. &
4\\

R20 &
C3 / A &
\texttt{pillow / 68 / f4} &
The public task specifies an \code{error\_handler} callable and high-level
strategy but not its exact arguments or return protocol; the evaluator fixes
\code{(error,key,value)} and specific fallback/skip semantics. &
4\\

R21 &
C5 / D &
\texttt{pillow / 68 / f5} &
The public task requires relevant logging, whereas the evaluator accepts a
particular unpublished sentence form such as ``Orientation value: 3''. &
4\\

R22 &
C1 / D &
\texttt{pillow / 290 / f4} &
The public contract permits fallback to the maximum color count when the MSE
threshold cannot be met, whereas the evaluator imposes strict error ordering.
The reference image remains above the stated threshold even at 256 colors. &
4\\

R23 &
C5 / D &
\texttt{llama\_index / 17244 / f4} &
The public \code{force\_mimetype} contract does not require the exact
\code{data:image/png;base64,} representation, while the evaluator does. &
6\\

R24 &
C4 / A &
\texttt{llama\_index / 18813 / f5} &
The public task mentions trimming but does not define it as removal of zero
bytes from both ends of the raw binary payload; the evaluator fixes that
interpretation. &
5\\

R25 &
C2 / D &
\texttt{llama\_index / 18813 / f6} &
For \code{format=mp3} without an explicit mimetype, the evaluator requires
unpublished \code{BytesIO.mimetype} and \code{BytesIO.as\_base64} attributes. &
5\\

R26 &
C2 / D &
\texttt{dirty\_equals / 43 / f6} &
The public task defines an issuer parameter, while test collection directly
requires an unpublished class-subscript API such as
\code{IsCreditCard["Visa"]}. &
8\\

R27 &
C4 / A &
\texttt{dirty\_equals / 43 / f7} &
The public contract does not determine whether an invalid hash algorithm should
fail at construction or evaluate to false during comparison; the evaluator
requires a constructor-time \code{ValueError}. &
8\\

R28 &
C2 / D &
\texttt{react\_hook\_form / 153 / f6} &
The public contract does not make the otherwise required \code{onValid}
argument optional, while the evaluator supplies \code{undefined} and requires
successful handling. &
5\\

R29 &
C1 / D &
\texttt{react\_hook\_form / 153 / f5} &
The public task explicitly permits either a configuration object or an
additional parameter, while the evaluator accepts only a third-argument
\code{options} object. &
5\\

R30 &
C1 / D &
\texttt{react\_hook\_form / 85 / f2} &
The public contract permits either \code{formState.isLoading} or a dedicated
state representation, while the evaluator accepts only the former. &
4\\

R31 &
C2 / A &
\texttt{dspy / 8635 / f4} &
The public task permits \code{None} to be intercepted at the call site, while
the evaluator directly invokes a private helper with \code{None}; the
private-helper input contract is not specified. &
5\\

R32 &
C1 / D &
\texttt{dspy / 8587 / f3} &
The public task describes timestamps as optional, while the evaluator requires
specific private \code{\_t0}/\code{\_t\_last} mappings and fixed output keys. &
5\\

R33 &
C6 / D &
\texttt{pillow / 25 / f1--f2} &
Feature~1 requires readonly state to be preserved when saving to a different
file, whereas Feature~2 requires the default
\code{preserve\_readonly=False} behavior to ensure that the destination is
writable before saving. The same default save-as operation therefore cannot
simultaneously satisfy both public feature contracts. &
1\\

R34 &
C2 / -- &
\texttt{click / 2068 / f5--f8} &
The public feature permits lock-conflict failure but does not specify the
exception class; the evaluator recognizes only a specific exception type. &
1\\

R35 &
C4 / -- &
\texttt{pillow / 25 / f1--f5, f2--f5, f3--f5} &
The public corner-marking contract does not specify source-image immutability
or restoration after a marked save, while the evaluator enforces that boundary. &
3\\

R36 &
C4 / -- &
\texttt{dirty\_equals / 43 / f1--f3} &
The public email contract does not specify a minimum final-label length, while
the evaluator rejects the one-character final label used by this exact pair. &
1\\

R37 &
C1 / -- &
\texttt{dspy / 8394 / f2--f3} &
The public namespace contract requires the same value to be identical to
\code{None} after assignment and remain callable as a context manager. &
1\\

R38 &
C6 / -- &
\texttt{dspy / 8635 / f1--f6, f5--f6} &
The paired public contracts impose incompatible adapter/probe revision and
composition requirements across the interacting features. &
2\\

\end{longtable}
\arrayrulecolor{black}
\endgroup

\paragraph{Why 38 issues affect 183 scoring units.}
CooperBench scores exact feature pairs rather than isolated features.
Consequently, a contract defect associated with one repeatedly paired feature
propagates to every scoring unit containing that feature. For example, the
unpublished exact-output requirement in \texttt{click/2068/f10} affects
11 pairs; the \texttt{tiktoken} parameter-name and hidden-cache issues each
affect nine; the \texttt{jinja} priority conflict affects nine; and the
\texttt{outlines} separator ambiguity affects nine.

The 183 excluded pairs arise from \textbf{38 shared feature- or pair-level
contract issues propagated through CooperBench's pairwise composition}.

\subsection{Released-Baseline Sensitivity}
\label{app:cooperbench_baseline_sensitivity}

The definite-defect policy leaves $652-111=541$ exact pairs, while the final
validity policy leaves $652-183=469$. We apply the identical pair sets to all
released trajectories.

\begin{table}[H]
\centering
\begingroup
\small
\arrayrulecolor{RuleGray}
\setlength{\tabcolsep}{3.4pt}
\renewcommand{\arraystretch}{1.10}

\begin{tabularx}{\linewidth}{
    @{}
    L
    C{0.10\linewidth}
    C{0.22\linewidth}
    C{0.22\linewidth}
    C{0.22\linewidth}
    @{}
}
\toprule
\rowcolor{PanelFill}
\textbf{Validity policy} &
\textbf{$N$} &
\textbf{GPT-5.5 Solo} &
\textbf{Peer / coop+git} &
\textbf{Team no-proto}\\
\midrule

Raw &
652 &
362/652 (55.5\%) &
329/652 (50.5\%) &
\bestcell{403/652 (61.8\%)}\\

Definite-defect &
541 &
339/541 (62.7\%) &
305/541 (56.4\%) &
\bestcell{380/541 (70.2\%)}\\

Final validity &
469 &
311/469 (66.3\%) &
277/469 (59.1\%) &
\bestcell{349/469 (74.4\%)}\\

\bottomrule
\end{tabularx}

\caption{\textbf{Released GPT-5.5-hao CooperBench trajectories under three
benchmark-validity policies.}
Only official PASS outcomes enter the numerator; the denominator is fixed by
the validity policy.}
\label{tab:cooperbench_released_sensitivity}

\endgroup
\arrayrulecolor{black}
\end{table}

The validity policy changes absolute pass rates but not the structural ordering:
\[
\text{Peer / coop+git} < \text{Solo} < \text{Team no-proto}.
\]
In the released CooperBench settings, plain peer cooperation is therefore the
weakest of the three structures: splitting the interacting features across
peers reduces performance relative to Solo, whereas the structured
lead--member Team exceeds Solo.

\subsection{Relic Results and Matched-Pair Sensitivity}
\label{app:cooperbench_full_results}

Within each validity policy, Relic, Solo, Peer, and Team are compared on
exactly the same exact-pair identities with formal Relic verdicts.

The final validity comparison contains all 469 valid exact-pair identities;
the Raw and definite-defect sensitivity rows use their corresponding matched
identity sets.

\begin{table}[H]
\centering
\begingroup
\small
\arrayrulecolor{RuleGray}
\setlength{\tabcolsep}{3.0pt}
\renewcommand{\arraystretch}{1.10}

\begin{tabularx}{\linewidth}{
    @{}
    L
    C{0.19\linewidth}
    C{0.19\linewidth}
    C{0.19\linewidth}
    C{0.19\linewidth}
    @{}
}
\toprule
\rowcolor{PanelFill}
\textbf{Validity policy} &
\textbf{Relic} &
\textbf{Peer} &
\textbf{Solo} &
\textbf{Team no-proto}\\
\midrule

Raw ($n=616$) &
388/616 (63.0\%) &
317/616 (51.5\%) &
350/616 (56.8\%) &
\bestcell{391/616 (63.5\%)}\\

Definite-defect ($n=535$) &
\bestcell{383/535 (71.6\%)} &
303/535 (56.6\%) &
337/535 (63.0\%) &
378/535 (70.7\%)\\

Final validity ($n=469$) &
\bestcell{371/469 (79.1\%)} &
277/469 (59.1\%) &
311/469 (66.3\%) &
349/469 (74.4\%)\\

\bottomrule
\end{tabularx}

\caption{\textbf{Matched-pair CooperBench sensitivity.}
Within each row, all systems are scored on exactly the same pair identities
with formal Relic verdicts under that policy.}
\label{tab:cooperbench_relic_sensitivity}

\endgroup
\arrayrulecolor{black}
\end{table}

\paragraph{Raw comparison: lifting peer coordination to the Team regime.}
On the unfiltered matched set, Relic reaches 63.0\% on 616 identities,
compared with 51.5\% for the released Peer reference and 56.8\% for Solo on
the same identities. This is a +11.5 percentage-point gain over Peer and a
+6.2-point gain over Solo, with Relic 0.5 percentage points below the
hierarchical Team reference on the same pairs (63.5\%).

The released CooperBench baseline ordering is
\[
\text{Peer} < \text{Solo} < \text{Team}.
\]
Peer/coop+git is the weakest released coordination setting, whereas the
lead--member Team is the strongest. Relic retains two peer feature owners with
no permanent lead and reaches essentially the same performance regime as the
specialized hierarchical Team on the unfiltered matched set.

\paragraph{Sensitivity to benchmark validity.}
Under the definite-defect policy, Relic reaches 71.6\%, compared with 56.6\%
for Peer, 63.0\% for Solo, and 70.7\% for Team on the same 535 pair identities.
The corresponding differences are +15.0 points over Peer, +8.6 points over
Solo, and +0.9 points over Team.

Under the final validity policy, Relic reaches 79.1\% on the complete
469-pair validity set, compared with 59.1\% for Peer, 66.3\% for Solo, and
74.4\% for Team. The corresponding differences are +20.0 points over Peer,
+12.8 points over Solo, and +4.7 points over the hierarchical Team reference.

Relic's advantage increases under stricter validity policies. On the raw
matched set it exceeds Peer by 11.5 points and Solo by 6.2 points while
remaining within 0.5 points of Team; under the definite-defect and
final-validity policies it exceeds Team by 0.9 and 4.7 points, respectively.

\paragraph{From the weakest released topology to Team-level performance.}
The released CooperBench hierarchy shows a coordination penalty for plain
peers: splitting interacting features across two peers performs worse than
assigning both to one agent, while a specialized lead--member hierarchy is
required to exceed Solo.

Relic keeps the peer feature-owner structure and adds persistent organizational
state, governed protocol formation, and executable coordination rules. This
moves the peer topology from the weakest released CooperBench setting to
Team-level performance on the unfiltered matched set, above Team under the
definite-defect policy, and further above the hierarchical reference on the
final validity set.

\subsection{Frozen 48-Pair Same-Model Coordination Check}
\label{app:cooperbench_pair_list}

The same-model coordination check uses a fixed 48-pair subset spanning
30 task instances and 12 repositories, selected independently of Relic
outcomes. The final audit excludes one of those identities, leaving 47 valid
pairs scored for Relic, Official Peer, and Official Solo.

\begingroup
\scriptsize
\arrayrulecolor{RuleGray}
\setlength{\tabcolsep}{4.0pt}
\renewcommand{\arraystretch}{1.08}

\begin{longtable}{
    @{}
    C{0.045\linewidth}
    P{0.48\linewidth}
    C{0.15\linewidth}
    C{0.15\linewidth}
    @{}
}
\caption{\textbf{Frozen CooperBench same-model evaluation pairs.}}
\label{tab:cooperbench_48_pair_outcomes}\\
\toprule
\rowcolor{PanelFill}
\textbf{\#} & \textbf{Repository} & \textbf{Task} & \textbf{Pair}\\
\midrule
\endfirsthead

\toprule
\rowcolor{PanelFill}
\textbf{\#} & \textbf{Repository} & \textbf{Task} & \textbf{Pair}\\
\midrule
\endhead

\midrule
\multicolumn{4}{r}{\footnotesize Continued on next page}\\
\endfoot

\bottomrule
\endlastfoot

1 & \texttt{dottxt\_ai\_outlines} & 1371 & F1/F2\\
2 & \texttt{dottxt\_ai\_outlines} & 1655 & F1/F3\\
3 & \texttt{dottxt\_ai\_outlines} & 1655 & F6/F7\\
4 & \texttt{dottxt\_ai\_outlines} & 1655 & F7/F10\\
5 & \texttt{dspy} & 8563 & F1/F4\\
6 & \texttt{go\_chi} & 27 & F3/F4\\
7 & \texttt{go\_chi} & 27 & F2/F4\\
8 & \texttt{huggingface\_datasets} & 7309 & F1/F2\\
9 & \texttt{llama\_index} & 17070 & F1/F2\\
10 & \texttt{llama\_index} & 17244 & F5/F6\\
11 & \texttt{llama\_index} & 17244 & F2/F6\\
12 & \texttt{openai\_tiktoken} & 0 & F4/F8\\
13 & \texttt{openai\_tiktoken} & 0 & F1/F5\\
14 & \texttt{pallets\_click} & 2800 & F1/F4\\
15 & \texttt{pallets\_click} & 2800 & F1/F2\\
16 & \texttt{pallets\_jinja} & 1621 & F6/F10\\
17 & \texttt{pallets\_jinja} & 1621 & F1/F6\\
18 & \texttt{pillow} & 25 & F1/F5$^{\dagger}$\\
19 & \texttt{pillow} & 25 & F1/F4\\
20 & \texttt{pillow} & 68 & F1/F5\\
21 & \texttt{pillow} & 290 & F3/F5\\
22 & \texttt{pillow} & 290 & F2/F3\\
23 & \texttt{react\_hook\_form} & 153 & F2/F6\\
24 & \texttt{react\_hook\_form} & 153 & F1/F3\\
25 & \texttt{samuelcolvin\_dirty\_equals} & 43 & F2/F3\\
26 & \texttt{samuelcolvin\_dirty\_equals} & 43 & F2/F4\\
27 & \texttt{typst} & 6554 & F2/F6\\
28 & \texttt{typst} & 6554 & F1/F3\\
29 & \texttt{dottxt\_ai\_outlines} & 1706 & F4/F6\\
30 & \texttt{dottxt\_ai\_outlines} & 1706 & F5/F6\\
31 & \texttt{dspy} & 8394 & F3/F4\\
32 & \texttt{dspy} & 8394 & F3/F5\\
33 & \texttt{dspy} & 8587 & F1/F4\\
34 & \texttt{dspy} & 8587 & F2/F3\\
35 & \texttt{dspy} & 8635 & F1/F4\\
36 & \texttt{dspy} & 8635 & F4/F6\\
37 & \texttt{go\_chi} & 26 & F1/F2\\
38 & \texttt{go\_chi} & 26 & F2/F4\\
39 & \texttt{go\_chi} & 56 & F1/F5\\
40 & \texttt{go\_chi} & 56 & F2/F3\\
41 & \texttt{huggingface\_datasets} & 3997 & F1/F2\\
42 & \texttt{huggingface\_datasets} & 6252 & F4/F6\\
43 & \texttt{llama\_index} & 18813 & F1/F5\\
44 & \texttt{pallets\_click} & 2068 & F1/F4\\
45 & \texttt{pallets\_click} & 2956 & F1/F8\\
46 & \texttt{pallets\_jinja} & 1465 & F1/F7\\
47 & \texttt{pallets\_jinja} & 1559 & F1/F8\\
48 & \texttt{react\_hook\_form} & 85 & F3/F4\\

\end{longtable}
\arrayrulecolor{black}
\endgroup

\noindent$^{\dagger}$\texttt{pillow / 25 / F1/F5} was part of the original frozen 48-pair selection but is classified as BROKEN by the final benchmark-validity audit. We therefore exclude it from the same-model aggregate, exactly as broken pairs are excluded from the full-benchmark comparison. The reported controlled comparison consequently uses 47 final-valid pairs; this excluded identity was a Relic PASS and an Official Solo/Peer FAIL.

\subsubsection{Same-Model Results}
\label{app:cooperbench_results}

\begin{table}[H]
\centering
\begingroup
\small
\arrayrulecolor{RuleGray}
\setlength{\tabcolsep}{3.5pt}
\renewcommand{\arraystretch}{1.08}

\begin{tabularx}{\linewidth}{
    @{}
    L
    C{0.22\linewidth}
    L
    C{0.12\linewidth}
    @{}
}
\toprule
\rowcolor{PanelFill}
\textbf{System} &
\textbf{Model} &
\textbf{Structure} &
\textbf{Pass / 47}\\
\midrule

\armhead{ExecDark}{Relic (B3-2)}
& Claude Opus 4.6 (high)
& Two peer feature owners; no fixed lead
& \bestcell{\RelicCooperSolved{}/47}\\

Official Solo
& Claude Opus 4.6 (high)
& One agent receives both interacting features
& 26/47\\

Official Peer
& Claude Opus 4.6 (high)
& Two peers split the interacting features
& 13/47\\

\bottomrule
\end{tabularx}

\caption{\textbf{Same-model CooperBench coordination check on 47 final-valid pairs from the frozen 48-pair subset.}}
\label{tab:cooperbench_48_results}

\endgroup
\arrayrulecolor{black}
\end{table}

Official Peer drops from Solo's 26/47 to 13/47 when the interacting features
are divided across peers. Relic retains the same model, reasoning setting,
two-peer feature split, and absence of a permanent lead, yet reaches 28/47.
Thus Relic not only recovers the coordination loss observed in Official Peer
but exceeds the same-model Solo reference on this controlled subset.

Together with the full-benchmark comparison, the frozen-subset experiment provides
a direct same-model control over the peer coordination setting, while the full
benchmark evaluates the same peer-organizational design at benchmark scale
against the released Solo, Peer, and hierarchical Team references.

\section{External Software Benchmark: ProgramBench}
\label{app:programbench}
\label{app:programbench_external}

ProgramBench~\citep{yang2026programbench} evaluates a frozen executable protocol layer attached to the native Bash-action interface of a single mini-SWE-agent coding agent.

\subsection{System Configuration}
\label{app:programbench_system}

The baseline is official mini-SWE-agent~\citep{yang2024sweagent}. The treatment adds a stateful
protocol sidecar around the same single-agent Bash interface. The model
chooses actions, edits the candidate code, maintains the conversation
history, and decides when to submit. Before dispatch, the sidecar checks
the proposed action and accumulated execution evidence. Its response
allows the action, requests current evidence, or refuses a transition
that violates an executable rule.

The treatment uses mini-SWE-agent 2.4.6, ProgramBench 1.2.4, and
GPT-5.6 with \code{xhigh} reasoning effort and all-turn context.
Protocol content remains frozen throughout the target runs. Binding rules
and advisory rules are identified separately in
Table~\ref{tab:programbench-protocol-package}.

\subsection{Task Selection and Fixed Evaluation Set}
\label{app:programbench_tasks}

Adapter integration was developed on a separate demonstration task. We randomly sampled 25 ProgramBench tasks, fixed their identities, and evaluated both systems on this paired set.

\begin{table}[H]
\centering
\begingroup
\footnotesize
\arrayrulecolor{RuleGray}
\setlength{\tabcolsep}{4pt}
\renewcommand{\arraystretch}{1.08}
\begin{tabularx}{\linewidth}{@{}X X X@{}}
\toprule
\rowcolor{PanelSubFill}
\multicolumn{3}{@{}l}{\textbf{Fixed random sample of 25 ProgramBench tasks}}\\
\midrule
1.\ \code{zip-password-finder}
& 10.\ \code{tig}
& 19.\ \code{muffet} \\
2.\ \code{tui-journal}
& 11.\ \code{parqeye}
& 20.\ \code{elfcat} \\
3.\ \code{oranda}
& 12.\ \code{igrep}
& 21.\ \code{svd2rust} \\
4.\ \code{scc}
& 13.\ \code{loop}
& 22.\ \code{argc} \\
5.\ \code{rustowl}
& 14.\ \code{dropbear}
& 23.\ \code{code-minimap} \\
6.\ \code{datasurgeon}
& 15.\ \code{bartib}
& 24.\ \code{tty-clock} \\
7.\ \code{xh}
& 16.\ \code{gdal}
& 25.\ \code{nomino} \\
8.\ \code{rust-sloth}
& 17.\ \code{proj}
& \\
9.\ \code{7zip}
& 18.\ \code{quinn}
& \\
\bottomrule
\end{tabularx}
\caption{\textbf{ProgramBench evaluation set.} The task identifiers are
randomly sampled once and then held fixed for both systems.}
\label{tab:programbench-task-set}
\endgroup
\arrayrulecolor{black}
\end{table}

The sample contains 17 Rust tasks, three C++ tasks, three C tasks, and two Go
tasks. 

\subsection{Frozen Protocol Package and Runtime Binding}
\label{app:programbench_protocols}

The treatment uses the fixed \code{programbench\_pack\_v0} package.  It contains six executable binding rules and three advisory rules.
Harness adaptation maps their triggers, evidence fields, and actions to the
native mini-SWE-agent interface; the package remains fixed during evaluation.

\begin{table}[H]
\centering
\begingroup
\footnotesize
\arrayrulecolor{RuleGray}
\setlength{\tabcolsep}{4pt}
\renewcommand{\arraystretch}{1.08}
\begin{tabularx}{\linewidth}{
    @{}
    >{\raggedright\arraybackslash}p{0.09\linewidth}
    >{\raggedright\arraybackslash}p{0.14\linewidth}
    X
    @{}
}
\toprule
\rowcolor{PanelSubFill}
\textbf{Rule} & \textbf{Mode} & \textbf{Runtime intent} \\
\midrule
\code{PB0} & Binding &
Require an original candidate implementation rather than copying, linking, or
delegating execution to the reference executable. \\

\code{P0} & Binding &
Protect public probes and other harness-scored surfaces from modification by
the candidate. \\

\code{P3} & Binding &
Invalidate stale verification after source changes; modified code must be
rebuilt or rechecked in its current state. \\

\code{P4} & Binding &
Preserve build/check failure semantics; an unsuccessful or unavailable check
cannot be represented as successful evidence. \\

\code{P7} & Binding &
Require \code{patch.txt}, when present, to correspond to the current candidate
code state. \\

\code{P8} & Binding &
Gate final submission on originality, non-empty source changes, current build
evidence, absence of known failing current checks, and inspection of the
current diff. \\

\addlinespace[2pt]

\code{P1} & Advisory &
Encourage probing the available reference behavior on real inputs before
repeated implementation changes. \\

\code{P5} & Advisory &
Encourage obtaining new reference evidence when the same failure recurs
without new information. \\

\code{P6} & Advisory &
Encourage periodic re-inspection of the current source and diff during long
trajectories. \\
\bottomrule
\end{tabularx}
\caption{\textbf{ProgramBench protocol package.} Binding rules can affect
execution; advisory rules are recorded as shadow guidance and do not themselves
block an action.}
\label{tab:programbench-protocol-package}
\endgroup
\arrayrulecolor{black}
\end{table}

\subsection{Runtime Footprint}
\label{app:programbench_runtime}

Across the 25 treatment trajectories, the coding agent executes 2,215 Bash
actions. The protocol layer records 933 validation receipts and 815 inspections
of source, diffs, or reference behavior, while issuing 40 hard refusals.
Twenty-six hard refusals arise from \code{P3}, requiring current verification
after the code state changes, and fourteen arise from \code{P4}, preserving a
failed or unavailable build/check as a failure rather than accepting it as
successful evidence.

The three advisory rules additionally record 293 shadow cases in which their
recommended condition would have requested a different next step:
239 under \code{P6}, 37 under \code{P1}, and 17 under \code{P5}. These shadow
events do not block execution.

The 933 validation receipts comprise 370 \code{PASS}, 48 \code{FAIL}, and 515 \code{INCONCLUSIVE} outcomes.

\subsection{Scoring and Outcomes}
\label{app:programbench_scoring}

We compute each task's behavioral pass percentage using ProgramBench's native evaluator, excluding tests marked ignored by that evaluator, and macro-average over the fixed 25 tasks. Both systems use the same scoring procedure.

Official mini-SWE-agent reaches 64.164\%; the protocol treatment reaches 70.916\%, an absolute gain of 6.752 percentage points (10.5\% relative). All 25 treatment trajectories produce submissions and completed evaluations.

\stopcontents[appendices]
\end{document}